\documentclass[10pt,twocolumn,letterpaper]{article}

\usepackage{cvpr}              
\usepackage{comment}

\usepackage[dvipsnames]{xcolor}

\definecolor{cvprblue}{rgb}{0.21,0.49,0.74}
\usepackage[pagebackref,breaklinks,colorlinks,citecolor=cvprblue]
{hyperref}
\usepackage{multirow}
\usepackage{diagbox} 
\usepackage{bm}
\usepackage{adjustbox}
\usepackage{float}
\usepackage{graphicx}
\usepackage[capitalize]{cleveref}
\crefdefaultlabelformat{#2\textcolor{black}{#1}#3} 
\crefname{section}{Sec.}{Secs.}
\Crefname{section}{Section}{Sections}
\crefname{tab}{Tab.}{Tabs.}
\crefname{fig}{Figure.}{Figures.}

\def\paperID{10046} 
\def\confName{CVPR}
\def\confYear{2025}

\title{360Recon: An Accurate Reconstruction Method Based on \\ Depth Fusion from 360 Images}

\author{Zhongmiao Yan\and Qi Wu \and Songpengcheng Xia\and Junyuan Deng\and Xiang Mu\and Renbiao Jin\and Ling Pei\thanks{Corresponding authors}
}

\begin{document}
\newcommand{\warning}[1]{\textcolor[RGB]{255, 0, 0}{#1}}
\newcommand{\Qi}[1]{\textcolor[RGB]{0, 0, 255}{#1}}
\newcommand{\yzm}[1]{\textcolor[RGB]{160,82,45}{#1}}
\newcommand{\djy}[1]{\textcolor[RGB]{108,203,95}{#1}}
\maketitle
\begin{abstract}
360-degree images offer a significantly wider field of view compared to traditional pinhole cameras, enabling sparse sampling and dense 3D reconstruction in low-texture environments. This makes them crucial for applications in VR, AR, and related fields. However, the inherent distortion caused by the wide field of view affects feature extraction and matching, leading to geometric consistency issues in subsequent multi-view reconstruction. In this work, we propose 360Recon, an innovative MVS algorithm for ERP images. The proposed spherical feature extraction module effectively mitigates distortion effects, and by combining the constructed 3D cost volume with multi-scale enhanced features from ERP images, our approach achieves high-precision scene reconstruction while preserving local geometric consistency. Experimental results demonstrate that 360Recon achieves state-of-the-art performance and high efficiency in depth estimation and 3D reconstruction on existing public panoramic reconstruction datasets. \footnote{Code will be released upon paper acceptance.}
\vspace{-0.6cm}

\end{abstract}    
\section{Introduction}
\label{sec:intro}

\begin{figure}[t]
  \centering
   \includegraphics[width=\linewidth]{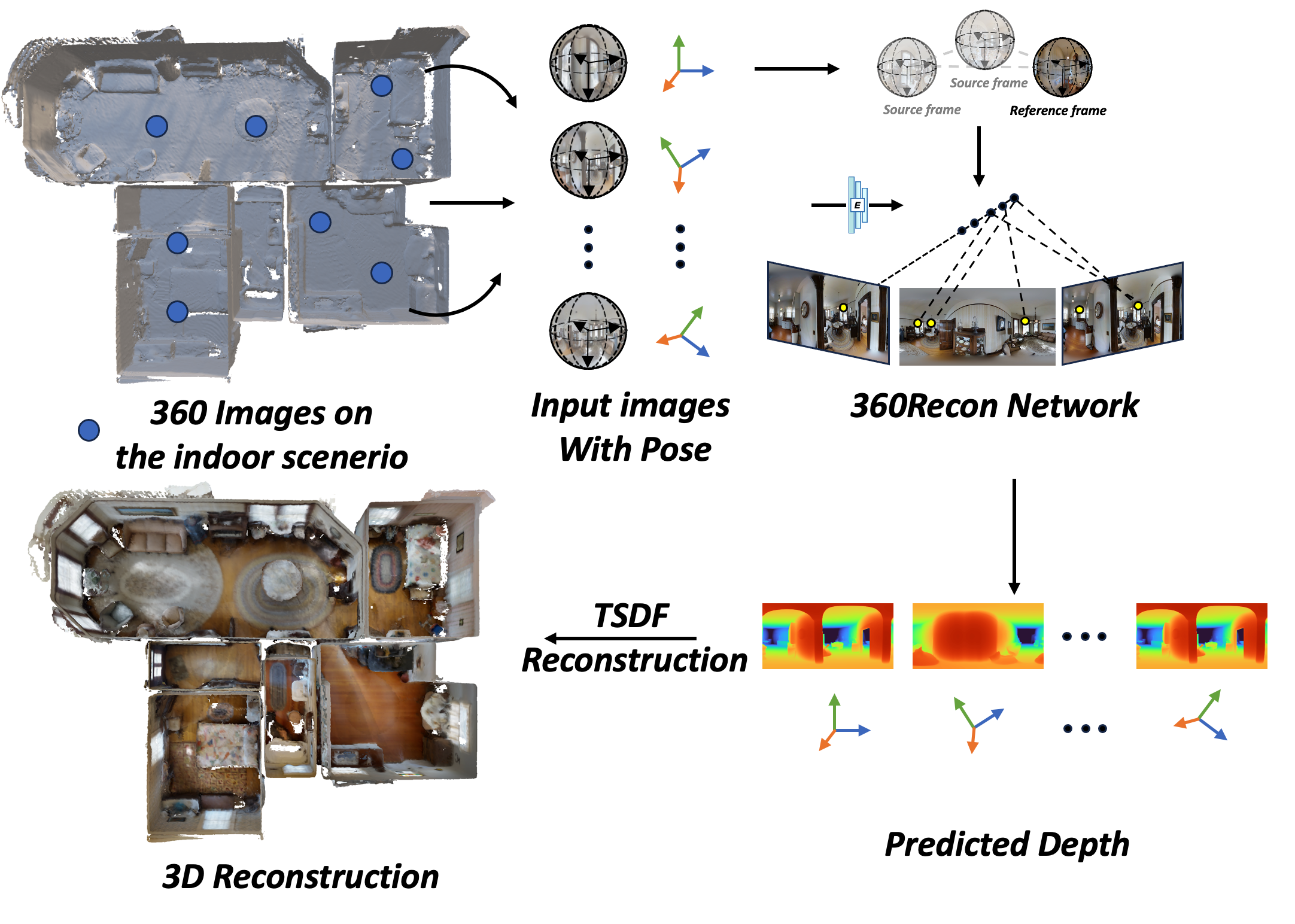}

   \caption{Illustration of the proposed 360Recon algorithm, enabling precise 3D scene reconstruction through depth prediction from 360° images.}
   \label{fig:intro}
   \vspace{-0.7cm}
\end{figure}

3D scene reconstruction is a fundamental requirement for various applications, including autonomous driving~\cite{yao2018mvsnet}, augmented/virtual reality (AR/VR)~\cite{gao2023visfusion}, and scene inspection~\cite{360VIO}. Vision-based 3D scene reconstruction, in particular, has been a significant focus of research. Despite numerous studies successfully achieving 3D scene reconstruction using conventional cameras, such as pinhole and fisheye cameras~\cite{simplerecon, yao2018mvsnet}, these algorithms still encounter challenges when applied to environments with sparse sampling density, weak textures, and transparent media, limiting their effectiveness.

With advancements in hardware, 360° cameras have increasingly captured the attention of the research community. Whether configured in a stitched or back-to-back setup, these cameras offer wide-angle views and rich environmental data. Compared to traditional methods that require dense image captures, 360° cameras typically need only a few images to describe the environment~\cite{360mvsnet} adequately. However, the distortion caused by their wide field of view presents challenges in feature extraction and description during 3D reconstruction.

Current research on 360° camera-based scene reconstruction primarily focuses on depth estimation from single images, enabling the recovery of structural information from a single frame~\cite{Bifusev2, panoformer}. However, monocular depth estimation cannot estimate absolute scale, and its geometric inconsistency issues are further amplified in reconstruction tasks. MVS-based methods for 360° images have so far shown limited success. The only existing method, 360MVSNet~\cite{360mvsnet,panogrf}, applies 3D CNNs to smooth and fuse the learned scene features. However, due to its neglect of distortion effects in the ERP (Equirectangular Projection) image representation and the texture information inherent in the images, it performs poorly when handling real 360-degree images with varying camera baselines.

In this study, we present 360Recon, an MVS reconstruction algorithm designed for ERP representations. By integrating a 360-depth-based TSDF reconstruction method, our approach achieves highly accurate 3D reconstructions by improving the quality of multi-view depth predictions. Unlike 360-MVSNet, our method addresses the distortion inherent in panoramic images through the proposed Spherical Feature Extraction, significantly enhancing the geometric accuracy of extracted features. Furthermore, instead of relying on a time-consuming and computationally intensive 3D CNN in the spatial domain, our algorithm employs a 2D CNN on each ERP image to refine image priors. This is combined with a 3D feature volume constructed via Spherical Sweeping and optimized using geometric loss functions, resulting in improved accuracy and efficiency for local depth estimation.

In summary, our key contributions are threefold:
\begin{itemize}

\item We introduce \textbf{360Recon}, a novel MVS reconstruction algorithm designed for ERP image formats. By incorporating a 360-depth-based fusion algorithm, our approach enables high-precision 3D scene reconstruction using only a small number of images.

\item We propose a spherical feature extractor and 2D image prior enhancement techniques, which significantly improve reconstruction performance together. The spherical feature extractor mitigates the distortion caused by wide fields of view, while the 2D image prior enhancement boosts the accuracy of depth estimation in local regions.

\item Extensive quantitative and qualitative evaluations demonstrate that our method achieves state-of-the-art performance on three public datasets.
\end{itemize}

\section{Related Work}

\begin{figure*}[t]  
    \centering
    \includegraphics[width=0.95\textwidth]{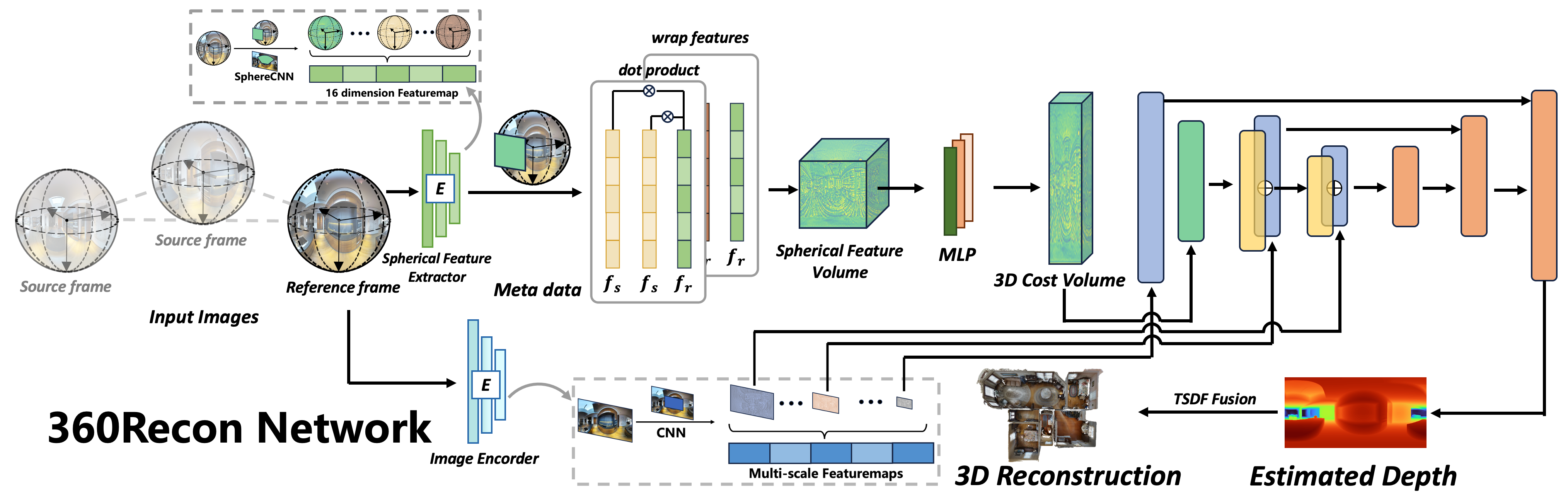} 
        \caption{\textbf{Pipeline of our method:} First, the spherical feature extractor mitigates distortion effects, followed by the enhancement of local geometric consistency using features extracted from ERP images. Finally, a self-developed 360° depth TSDF fusion approach is applied for the final 3D scene reconstruction.}
        \vspace{-0.5cm}
    \label{fig:pipeline}
\end{figure*}

This paper focuses on the  ERP representation of panoramic images, excluding methods that use multiple fisheye cameras, such as OmniMVS~\cite{OmniMVS}.

Our approach is closely related to previous work in 360-degree depth estimation, multi-view depth estimation, and 3D reconstruction. These areas are discussed below.

\subsection{360° Monocular Depth Estimation}
Monocular depth estimation from panoramic images does not require pose information, offering broader applicability. However, it faces challenges in maintaining the physical scale of depth values and frame-to-frame consistency. 

One approach for depth estimation of panoramic images is to divide the panoramic image into undistorted pin-hole camera model images. Then, depth estimation is performed either using the pin-hole camera model images alone or by combining both the pin-hole images and the ERP images. BiFuse~\cite{BiFuse} uses dual-branch networks to process ERP and CubeMap projections, mitigating distortion. OmniFusion~\cite{OmniFusion} and 360MonoDepth~\cite{360MonoDepth} divide the ERP representation of panoramic images into a series of spherical sections and attempt to perform depth estimation using pin-hole camera depth prediction for each section. HRDFuse~\cite{HRDFuse} processes both the ERP representation of the panoramic image and the spherical section images separately, then combines them to ultimately generate the depth map.

Methods such as Spherical View Synthesis~\cite{sphericalviewsynthesis}, BiFuse++~\cite{Bifusev2}, and SPDET~\cite{SPDET}, leverage multi-frame consistency for self-supervised training but still rely on single-frame information for depth estimation, which prevents them from predicting the true physical scale in new scenes. 

PanoFormer~\cite{panoformer} and S2Net~\cite{S2Net} introduce spherical sampling methods into the Transformer architecture to mitigate the distortion effects. Panoramic Depth Calibration~\cite{CalibratingPanoDepth} aligns predicted depth scales with real-world measurements but requires fine-tuning when testing on new scenes.

\subsection{Multi-View 360° Depth Estimation}
Pose information of images is generally required for multi-view depth estimation, ensuring spatial consistency and real-world scale across frames, which is essential for 3D reconstruction. Some multi-view depth estimation algorithms use a fixed camera baseline for multi-view constraints, which limits the algorithm's applicability to a broader range of panoramic camera types. 360SD-Net~\cite{360sdnet} uses two panoramic cameras, one on top and one on the bottom, for stereo depth estimation, with the distance between the camera centers serving as the fixed baseline for the algorithm. MODE~\cite{MODE} uses multiple frames at the same time but also requires baseline parameters for depth estimation. 

360MVSNet~\cite{360mvsnet} utilizes a cost volume for depth prediction, but it does so without incorporating prior knowledge, relying instead on computationally expensive 3D CNNs. PanoGRF~\cite{panogrf} incorporates monocular depth estimates into the cost volume, yet it still suffers from a lack of sufficient prior information.

IndoorPanoDepth~\cite{indoorPanoDepth} leverages the NeRF~\cite{nerf} method to achieve high-quality depth estimation, but its computational complexity makes it impractical for large-scale scenes. FoVA-Depth~\cite{fova} converts panoramic images into pinhole camera models using the nvTorchCam library~\cite{nvtorchcam} for depth estimation. However, this approach sacrifices the wide field of view, resulting in reduced depth consistency.

\subsection{3D Reconstruction}
We categorize related work in 3D reconstruction based on the use of 360-degree and pinhole camera images.

Classic methods for dense 3D reconstruction from images typically generate depth maps for each view~\cite{Colmap} and subsequently apply techniques such as Poisson surface reconstruction~\cite{poisson} to generate the surface. Kinect Fusion~\cite{KinectFusion} introduced real-time 3D reconstruction by employing TSDF~\cite{TSDF}, which allows for mesh generation via the marching cubes algorithm~\cite{marchingcubes}. However, in the field of 360-degree images, there is a lack of depth fusion methods. Existing reconstruction algorithms~\cite{Bifusev2,fova,360mvsnet} rely on depth values to back-project pixel points into 3D space for point cloud reconstruction. OmniSDF~\cite{OmniSDF}  uses an omnidirectional signed distance function but is limited by small-scale scene sweeps.

For pinhole-based 3D reconstruction, methods such as SimpleRecon~\cite{simplerecon}, TransformerFusion~\cite{transformerfusion}, and NeuralRecon~\cite{NeuralRecon} face challenges when image sampling is sparse. The use of pinhole models for multi-view reconstruction compromises the panoramic camera's wide field of view and makes it more difficult to identify co-visible frames.

\section{Method}
\label{method}
\subsection{Overall Architecture}

An overview of our 360Recon pipeline is shown in ~\cref{fig:pipeline}. Given a reference 360 frame $\bm{I}^r$, a set of source 360 frames $\bm{I}^{s}_{i\in(1,\cdots, n)}$ along with their intrinsics $\bm{K}$ and camera poses $\bm{P}$, our target is to obtain the depth estimation $\bm{D}^r_{pred}$ for each reference frame and incrementally fuse them to reconstruct the 3D scene. To achieve this goal, we first extract image features from images using our proposed spherical feature extractor (see ~\cref{subsec:spherical}). These features are used to construct the 4D Cost Volume through Spherical Sweeping, which is then reduced to a 3D probability volume through a multi-level perceptron (MLP) network (see ~\cref{subsec:sweeping}). We use an Encoder-Decoder structure to integrate multi-view information from the 3D cost volume and enhanced features from the images, generating the final depth prediction values across multiple scales. Our 360-degree depth fusion algorithm is then applied for 3D reconstruction (see ~\cref{sec:tsdf-reconstruction}). The network is supervised with a combination of loss functions  (see ~\cref{subsec:Loss}).

\subsection{Spherical Feature Extractor}
\label{subsec:spherical}

Due to the distortion caused by the horizontal stretching of the ERP representation of panoramic images, directly using standard square convolution kernels alone will result in significant differences for features at the same spatial location but at different vertical positions. This can affect the feature-matching process during subsequent spherical sweeping (\cref{subsec:sweeping}). To mitigate the impact of the uncertain distortion, we adopt the spherical convolution approach proposed in SphereNet~\cite{SphereNet}, where the convolution kernel is made variable by changing the pixel sampling method of the convolution.

For a given pixel point $p$ and its spherical coordinates $(\theta ,\phi)$, we use a fixed sampling pattern on the tangent plane at that point. An orthogonal coordinate system is established with the tangent point as the center of the plane. For a $3\times3$ convolution kernel, the sample pattern is $\mathbf{x}_{(0,0)}=(0,0)$, $\mathbf{x}_{(\pm1,0)}=(\pm\tan\Delta_\theta,0)$, $\mathbf{x}_{(0,\pm1)} =(0,\pm\tan\Delta_{\phi})$, $\mathbf{x}_{(\pm1,\pm1)} =(\pm\tan\Delta_\theta,\pm\sec\Delta_\theta\tan\Delta_\phi)$, which represent the coordinate of the sample points on the tangent plane.
The reason we choose this sampling pattern on the is that the sampled points in the spherical coordinate system with the tangent point as the origin are expressed as $\mathbf{s}_{(0,0)} =\begin{pmatrix}0,0\end{pmatrix}$, $\mathbf{s}_{(\pm1,0)} =(\pm\Delta_{\phi},0)$, $\mathbf{s}_{(0,\pm1)} =(0,\pm\Delta_\theta)$, $\mathbf{s}_{(\pm1,\pm1)}  = (\pm\Delta_\phi,\pm\Delta_\theta)$. This ensures uniform sampling of longitude and latitude in a fixed pattern at different positions on the sphere, rather than the traditional uniform sampling of the convolution kernel on a plane.

We introduce the spherical convolution method to mitigate the distortion of ERP images while maintaining the feature extraction capabilities of the network. We insert the spherical feature extraction layer into the ResNet network~\cite{Resnet}. The original network layers only contain regular convolution blocks. We parallelly add the Spherical CNN block with the original network layers, applying both regular CNN and Spherical CNN to the same input. The results generated by both are then summed to produce the output of this layer. By applying the same feature extractor to $\bm{I}^r$ and source frames $\bm{I}^s_i$, we obtain the feature maps $\bm{F}^r$, $\bm{F}^s_i$ used for Spherical Sweeping and Cost Volume construction.

\begin{figure}[h]  
    \centering
    \includegraphics[width=1\linewidth]{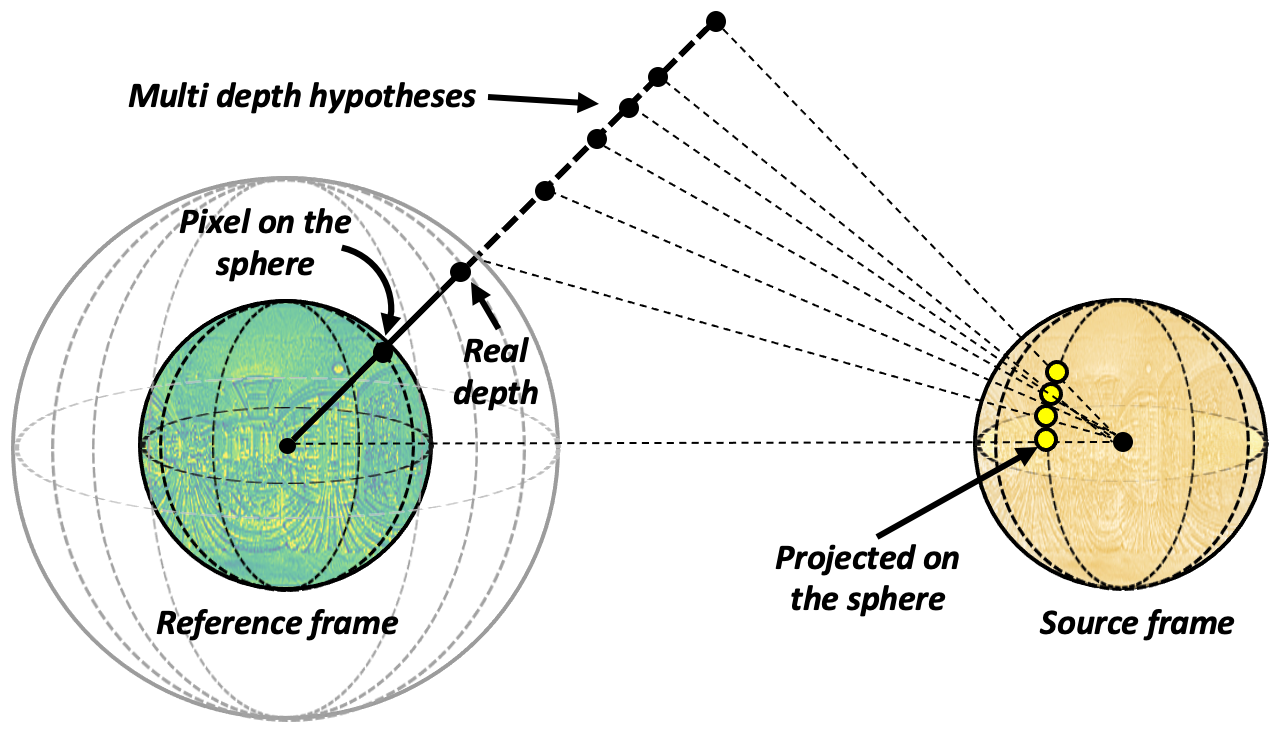} 
        \caption{\textbf{Illustration of spherical sweeping:} Based on multiple depth hypotheses, pixels from the reference frame are reprojected onto the source frame through spatial transformation to extract corresponding spherical features.}
    \label{fig:sweeping}
    \vspace{-0.4cm}
\end{figure}

\subsection{Spherical Sweeping}
\label{subsec:sweeping}
For multi-view depth estimation, how to effectively utilize the information from source views is crucial. We use depth hypotheses $r_i$ to back-project points from the reference frame $p^r$ into 3D space $P^r$, and then project these points onto the source frames $\bm{I}^s_i$ using the relative pose. The feature of the pixels on reference frame $p^r$ and the corresponding mapping points of source frames $p^s_i$ are used to construct the Cost Volume. The entire process, as shown in~\cref{fig:sweeping}, can be viewed as a feature matching process based on depth hypotheses.

Due to the characteristics of the ERP view's panoramic field of view, its intrinsic matrix $\bm{K}$ is determined by the image's height $H$ and width $W$.The intrinsic parameters and the process of transforming from the camera coordinate system to the world spherical coordinate system can be expressed as follows:
\begin{equation}
    \pi(p)=\begin{bmatrix}u\\v\\1\end{bmatrix}=\underbrace{\begin{bmatrix}W/2\pi&0&W/2\\0&-H/\pi&H/2\end{bmatrix}}_{\mathbf{K}}\begin{bmatrix}\theta\\\phi\\1\end{bmatrix}
\end{equation}
We follow the coordinate system conventions of ~\cite{360VIO}, the $P^r={[\theta,\phi,r_i]}^T$ is on the  camera space and parameterized by the longitude $\theta$, the latitude $\phi$, and the depth $r_i$, where the $-\pi <\theta < \pi$ and $-\pi/2 < \phi < \pi/2$.
The transformation between spherical coordinates and Cartesian coordinates can be expressed as follows:
\begin{equation}
      \resizebox{0.43\textwidth}{!}{$
   P_x = r_i\cdot\cos(\theta)\, \sin(\phi), \quad
   P_y = r_i\cdot(-\sin(\phi)), \quad
   P_z = r_i\cdot\cos(\theta)\, \cos(\phi).
   $}
\end{equation}

In the spherical coordinate system, the projection and back-projection process of panoramic images is a distortion-free process. For a pixel $p_r$ in the reference frame $\bm{I}^r$, its corresponding feature is $f^r$, and its depth can be assumed to be $r$. In the i-th source frame $\bm{I}^s_i$, the corresponding feature value $f^s_i$ under this depth hypothesis can be calculated as follows:
\begin{equation}
    f^s_i=\bm{F}^s(\bm{K} \bm{P}^s_i  (\bm{P}^r)^{-1} r_i(\bm{K}^{-1}  p_r))
\end{equation}
where $\bm{F}^s$ represents the feature map of source frames, $\bm{P_r}$ represents the pose of the reference frame, which transforms the points from the world to the camera, and $\bm{P}^s_i$ represents the pose of the source frame.

We take the dot product of the features $f_{si}$ from all the source frames with the features $f_r$ from the reference frame, and then concatenate them together with the original features to form a vector of dimension c:
\begin{equation}
    [f^r, f^s_1, \cdots, f^s_n, (f^r)^T\cdot f^s_1, \cdots, (f^r)^T\cdot f^s_n]
\end{equation}

By performing the above operation for all pixels and all depth hypotheses, we obtain a 4D Cost Volume with dimension $C\times D\times H\times W$, where C is the dimension of the dot features and wrap features, and D represents the number of depth hypotheses.

To avoid the expensive memory and time cost of a 3D CNN, we simply use an MLP network to reduce the 4D cost volume to a 3D probability volume with the shape $D \times H \times W$. This process can be viewed as considering the information from $c$ dimensions for each pixel at each depth hypothesis, i.e., the constraints from all source frames, ultimately obtaining the confidence of the depth hypothesis.

 It is worth noting that for ERP  images, the depth value generally refers to the absolute distance from a 3D point to the camera center, rather than the planar depth commonly used in pinhole camera models. 
 This is due to the omnidirectional nature of ERP images, which makes it impossible to simply use the Z-coordinate to represent depth. Using the absolute distance is more consistent with the spherical model of panoramic images. The characteristics of the ERP image camera model make the equidistant plane at a given depth hypothesis a complete sphere rather than a plane.
 
\subsection{Depth Fusion}
\label{sec:tsdf-reconstruction}

When performing the final depth estimation, we not only consider the constraints between multiple views, which are stored in the 3D probability volume but also incorporate the geometric structural information of the reference frame itself. This approach ensures both the continuity and scale consistency of the depth predictions within the image. To achieve this, we employ a multi-scale Encoder-Decoder architecture, which effectively fuses the information and generates the final depth prediction.

After obtaining the depth estimation values, we aim to reconstruct the 3D structure on a real scale using the color images, depth data, and camera pose information. But to the best of our knowledge, there is currently no depth fusion algorithm specifically designed for ERP images. This is because the TSDF fusion methods are typically designed to work with RGB-D images, requiring intrinsic and extrinsic parameters for projection transformations based on the pinhole camera model. These methods cannot be directly adapted to the panoramic camera model, as the pinhole model and ERP camera model have fundamentally different projection characteristics.

To address this problem, we enhanced the existing unwrapped TSDF module algorithm from ~\cite{simplerecon} to adapt it for the ERP image projection model introduced in ~\cref{subsec:sweeping}. This enables the projection of voxels from the world coordinate system into the panoramic camera coordinate system, while simultaneously updating both the TSDF values and color information. In summary, we have implemented an incremental 3D reconstruction algorithm based on depth fusion that is compatible with the ERP camera model, enabling more accurate and efficient 3D scene reconstruction.

\subsection{Loss Design}
\label{subsec:Loss}
Our loss followed the loss design of ~\cite{simplerecon}, which consists of four parts: Depth regression loss, Multi-scale gradient, normal losses, and Multi-view depth regression loss. 

The first part is the depth regression loss $\mathcal{L}_\mathrm{depth}$, where we supervise the depth estimation values by applying the log of the depth predictions at each depth prediction scale. The multi-scale $\mathcal{L}_\mathrm{grad}$ gradient loss is the difference of the first-order spatial gradients between the predicted depth map and the ground truth. A simple normal loss $\mathcal{L}_\mathrm{normals}$ defines the difference between the predicted normal vectors, calculated using the intrinsic parameters and depth values, and the ground truth normal vectors. As a consistency constraint for multi-frame depth prediction, which is vital for 3D reconstruction, the ground truth depth maps of each source frame are used in $\mathcal{L}_\mathrm{mv}$ as additional supervision.  We supervise the model by projecting the predicted depth values from the reference frame onto the source frame, calculating the predicted depth, and comparing it with the ground truth depth values in the source frame. Our total loss is:
\begin{equation}
\resizebox{0.9\linewidth}{!}{$\mathcal{L}=\mathcal{L}_\mathrm{depth}+\alpha_\mathrm{grad}\mathcal{L}_\mathrm{grad}+\alpha_\mathrm{normals}\mathcal{L}_\mathrm{normals}+\alpha_\mathrm{mv}\mathcal{L}_\mathrm{mv}$}
\end{equation}
\vspace{-\baselineskip}
\begin{figure*}[t]  
    \centering
    \includegraphics[width=1.0\textwidth]{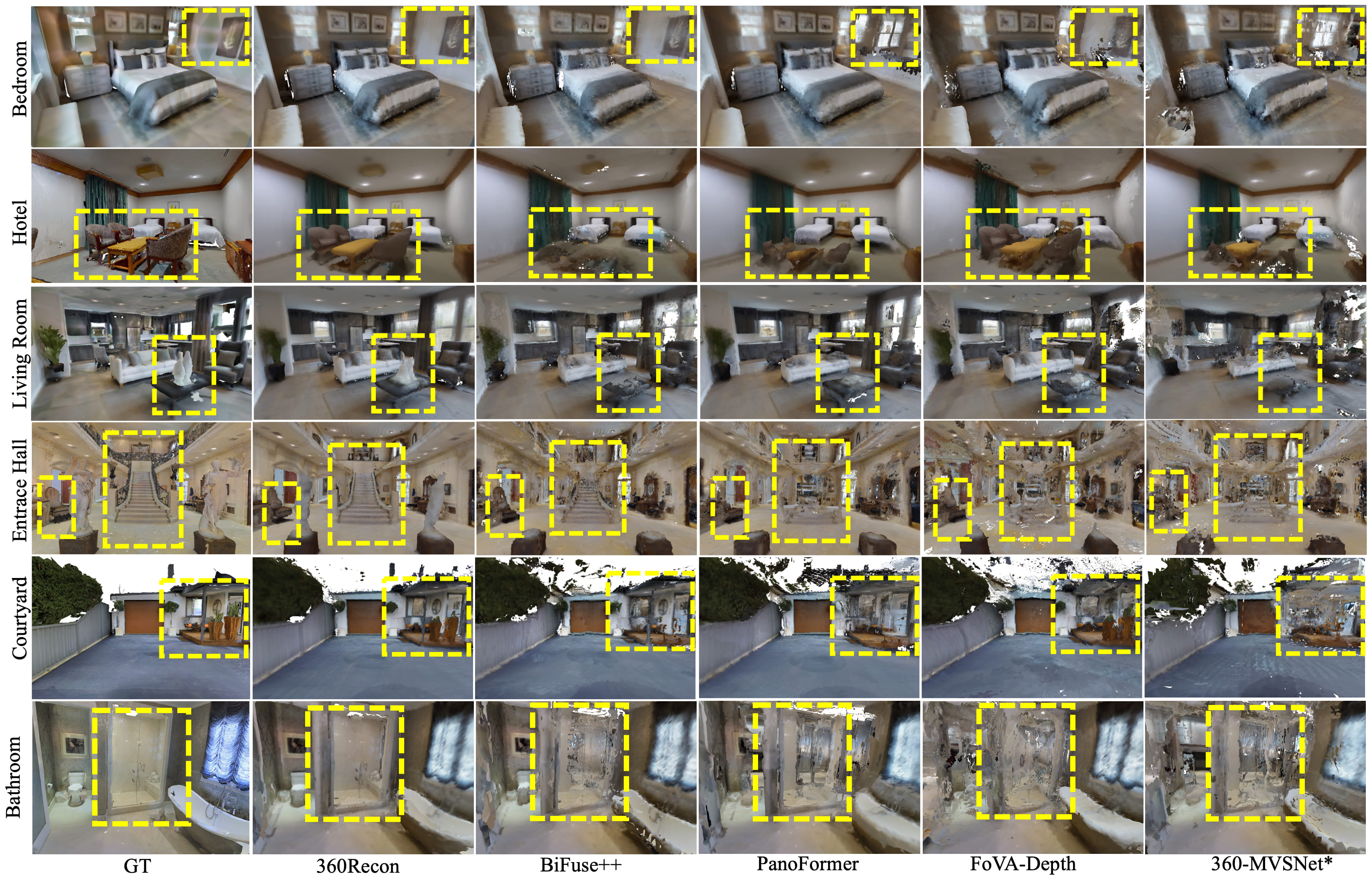}
    \vspace{-0.7cm}
    \caption{\textbf{3D Reconstruction Comparison.} The 3D reconstruction results are compared with various methods in different scenes. The reconstruction results of our algorithm in these scenarios are more accurate and complete.}
    \vspace{-0.6cm}
    \label{fig:reconstructions}
\end{figure*}
\begin{table}[tp]	
\centering
\fontsize{5.3}{4}\selectfont
\renewcommand{\arraystretch}{1.5} 
\label{tab:reconstructions}
\resizebox{\linewidth}{!}{
\begin{tabular}{cccccc}
	\toprule
	\multirow{3}{*}{\textbf{Dataset}}&\multirow{3}{*}{\shortstack{\textbf{Method}} }&\multicolumn{4}{c}{\textbf{Reconstruction Metrics}}\cr
	\cmidrule(l){3-6} 
	&&Comp$\downarrow$(cm) &Acc$\downarrow$(cm)&Chamfer$\downarrow$(cm)&F-Score$\uparrow$\cr
	\midrule
	
        \multirow{5}{*}{M3D\cite{Matterport3D}}&BiFuse++\cite{Bifusev2}&16.6&16.1&16.4&21.0\cr
	&PanoFormer\cite{panoformer}&15.7&21.7&18.7&18.6\cr
	&FoVA-Depth\cite{fova}&13.07&19.63&16.34&23.2\cr
	&360-MVSNet\textsuperscript{*}\cite{360mvsnet,panogrf} & 24.95 & 26.37 & 25.66 & 11.44 \cr
        &\textbf{360Recon}&\textbf{8.65}&\textbf{8.69}&\textbf{8.67}&\textbf{47.2}\cr\midrule

        \multirow{5}{*}{S2D3D\cite{S2D3D}} & BiFuse++\cite{Bifusev2}    & 29.79 & 25.68 & 27.73  & 2.9 \\
                              & PanoFormer\cite{panoformer}  & 27.80 & 28.85 & 28.33  & 2.9 \\
                              & FoVA-Depth\cite{fova}  & \textbf{23.64} & 29.43 & 26.53  & 3.4 \\
                              & 360-MVSNet\textsuperscript{*}\cite{360mvsnet,panogrf} & 30.15 & 27.99 & 29.08 & 2.5\\
                                & \textbf{360Recon} & 25.82 & \textbf{20.95} & \textbf{23.38}& \textbf{6.0} \\

	\midrule
        \multirow{5}{*}{OmniScenes\cite{OmniScenes}} & BiFuse++\cite{Bifusev2} & 13.21 & 21.47 & 17.29 & 23.1 \\
	& PanoFormer\cite{panoformer} & 17.81 & 26.66 & 22.2  & 10.4 \\
	& FoVA-Depth\cite{fova} & 18.02 & 29.51 & 23.76  & 23.1 \\
	& 360-MVSNet\textsuperscript{*}\cite{360mvsnet} & 22.79 & 30.06 & 26.42 & 13.76 \\
	& \textbf{360Recon} & \textbf{7.54} & \textbf{14.71} & \textbf{11.13} & \textbf{43.1} \\
\bottomrule
\end{tabular}}
\vspace{-0.2cm}
\caption{The qualitative comparison of 3D reconstruction metrics.The best metric in each column is highlighted in bold.}

\vspace{-0.2cm}
\end{table}

\begin{table}[tp]	
\centering
\fontsize{5.3}{4}\selectfont
\renewcommand{\arraystretch}{1.5} 
\label{table:depth}
\resizebox{\linewidth}{!}{
\begin{tabular}{ccccccc}
	\toprule
	\multirow{3}{*}{\textbf{Dataset}}&\multirow{3}{*}{\shortstack{\textbf{Method}} }&\multicolumn{5}{c}{\textbf{Depth Metrics}}\cr
	\cmidrule(l){3-7} 
	&&MVS&MAE(cm)&MRE(\%)&RMSE(cm)&$\delta_1(\%)$ \cr
	\midrule
	
        \multirow{5}{*}{M3D\cite{Matterport3D}} & BiFuse++\cite{Bifusev2}    & ×     & 43.75 & 15.79 & 76.12 & 79.09 \\
                            & PanoFormer\cite{panoformer}  & ×     & 33.65 & 14.13 & 60.98 & 85.66 \\
                            & FoVA-Depth\cite{fova}  & \checkmark     & 39.00 & 20.13 & 78.93 & 80.15 \\
                            & 360-MVSNet\textsuperscript{*}\cite{360mvsnet,panogrf} & \checkmark & 57.20 & 24.24 & 83.87 & 63.64 \\
                            & \textbf{360Recon}  & \checkmark  & \textbf{14.35} & \textbf{5.86} & \textbf{37.57} & \textbf{95.27} \\
        \midrule
	
        \multirow{5}{*}{S2D3D\cite{S2D3D}} & BiFuse++\cite{Bifusev2}    & ×     & 27.47 & 12.40 & 49.84 & 80.81 \\
                              & PanoFormer\cite{panoformer} & ×     & 21.31 & 11.29 & 39.72  & 86.27 \\
                              & FoVA-Depth\cite{fova}  & \checkmark & 30.57 & 16.60 & 60.23 & 82.52 \\
                              & 360-MVSNet\textsuperscript{*}\cite{360mvsnet,panogrf} & \checkmark & 30.97  & 15.77   & 51.79 & 74.80 \\
                              & \textbf{360Recon}  & \checkmark & \textbf{9.74} & \textbf{4.56} & \textbf{27.28} & \textbf{96.74} \\
	\bottomrule
\end{tabular}
}

\vspace{-0.2cm}
\caption{The qualitative comparison of depth estimation metrics. The best metric in each column is highlighted in bold.}

\vspace{-0.7cm}
\end{table}

\section{Experiments}
\label{experimets}
The experiment is divided into three parts. First, we assess the algorithm's performance by comparing its output with the ground truth, focusing on the overall reconstruction quality. Second, we validate the accuracy and consistency of local detail estimation by comparing the estimated depth maps with the ground truth depth. Lastly, we conduct ablation studies on the key components of the algorithm to analyze the impact of each module on its accuracy. The results of all experiments support the two claims outlined in the introduction (\cref{sec:intro}):

\begin{enumerate}
    \item The proposed algorithm, 360Recon achieves high reconstruction accuracy across various scenarios.
    \item The spherical feature extraction module, which has been carefully designed, helps reduce the distortion effects in the image data, thereby improving the accuracy of the algorithm.
    \item The integration of the geometric properties of 360 metadata can enhance depth quality and achieve improved spatial consistency in 3D scene structures.
\end{enumerate}

The entire algorithm is developed in PyTorch, and all experimental results were tested on a desktop system equipped with an NVIDIA RTX 3090 with 24GB of memory.
\vspace{-0.1cm}
\subsection{Experimental Setup}
\paragraph{Dataset.} We evaluate our algorithm on three public datasets—Matterport3D~\cite{Matterport3D}, Stanford2D3D~\cite{S2D3D}, and OmniScenes~\cite{OmniScenes}—and compare its performance with state-of-the-art (SOTA) methods. These datasets consist of real indoor panoramic images, accurate camera poses, and ground truth 3D scene reconstructions. For the Matterport3D dataset, we selected 61 scenes based on the standard split for model training, including 6,946 pairs of color and depth images. In contrast to the Matterport3D and Stanford2D3D datasets, which were captured using multi-camera stitching, OmniScenes employs a back-to-back fisheye 360° panoramic camera, offering a greater number of images and denser spatial capture sampling. This enables a more comprehensive analysis of the algorithm's generalization capabilities, demonstrating that the algorithm can achieve more accurate reconstruction results and better depth estimation performance across various indoor scenes and devices. As OmniScenes lacks ground truth depth, we evaluate depth estimation performance on the Matterport3D and Stanford2D3D datasets.
\vspace{-0.5cm}

\paragraph{Competing methods.} We compare the scene reconstruction capability of the proposed algorithm with several state-of-the-art methods, including BiFuse++, PanoFormer, FoVA-Depth, and 360-MVSNet$^{*}$, on three public datasets. Among them, BiFuse++ and PanoFormer are the top-performing monocular depth estimation algorithms, while FoVA-Depth is the first to extend MVS networks, and enables depth estimation of the large field of view images using the pre-trained pinhole camera model. Since the 360-MVSNet algorithm is not publicly available, we evaluate the 360-MVSNet network reproduced by PanoGRF, a depth-based novel view image synthesis method, and label it as 360-MVSNet$^{*}$ in the following experiments.

\vspace{-0.5cm}

\paragraph{Metrics.} For the evaluation of reconstruction capability, we adopt the methodology outlined in~\cite{simplerecon,transformerfusion}, selecting metrics such as \textit{Comp}, \textit{Acc}, \textit{Chamfer} and \textit{F-Score} to systematically assess the completeness, accuracy, and overall performance of the reconstruction. To evaluate the algorithm's depth prediction capability, we employ metrics such as \textit{MAE}, \textit{MRE}, \textit{RMSE}, and $\delta_1$ to assess the quality of the local geometric structure in the reconstruction, based on the predicted depth maps. 
Following the evaluation process proposed in~\cite{simplerecon}, we disregard invalid points (such as NaN values) and only evaluate the valid points.

\begin{figure*}[!h]
    \centering
    \newcommand{\mywidth}{\textwidth / 7}
    \setlength\tabcolsep{0.05em}
    \newcolumntype{P}[1]{>{\centering\arraybackslash}m{#1}}
    \def\arraystretch{0.30}
    \begin{tabular}{P{\mywidth} P{\mywidth} P{\mywidth} P{\mywidth} P{\mywidth} P{\mywidth} P{\mywidth}}
        \renewcommand{\arraystretch}{0.05}
        
        \includegraphics[width=\mywidth]{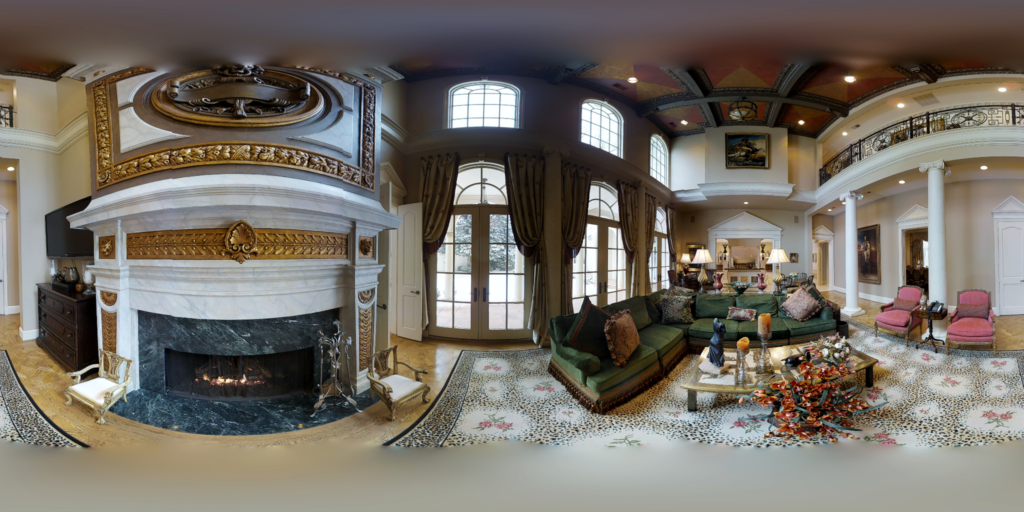}\hspace{-1cm} &
        \includegraphics[width=\mywidth]{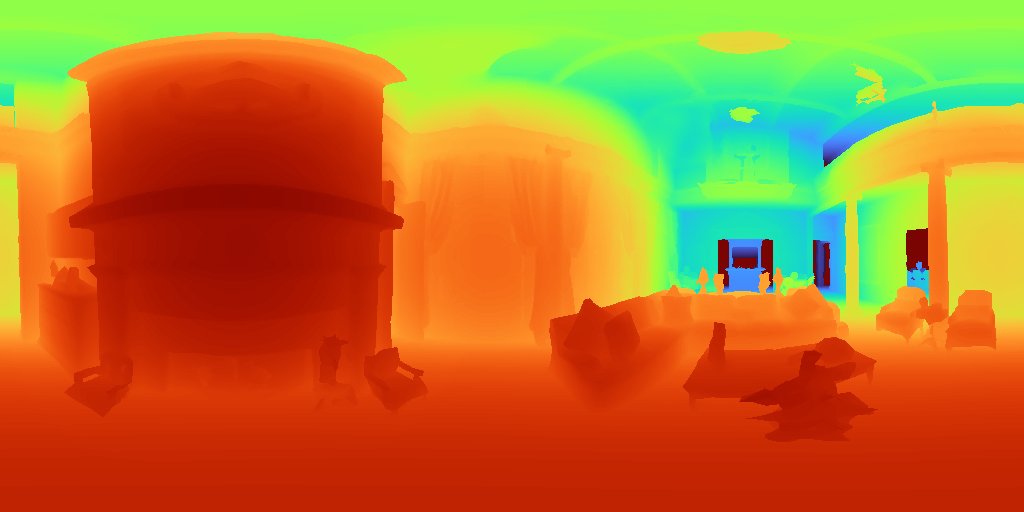}\hspace{-1cm} &
        \includegraphics[width=\mywidth]{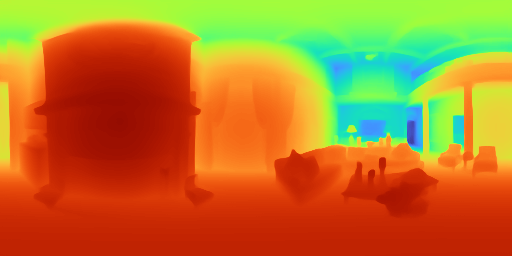} &
        \includegraphics[width=\mywidth]{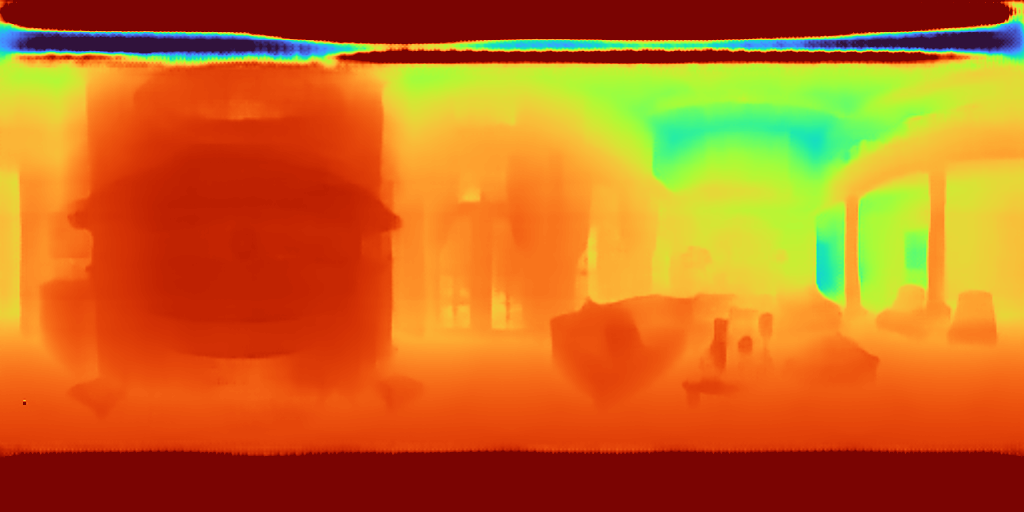}\hspace{-1cm} &
        \includegraphics[width=\mywidth]{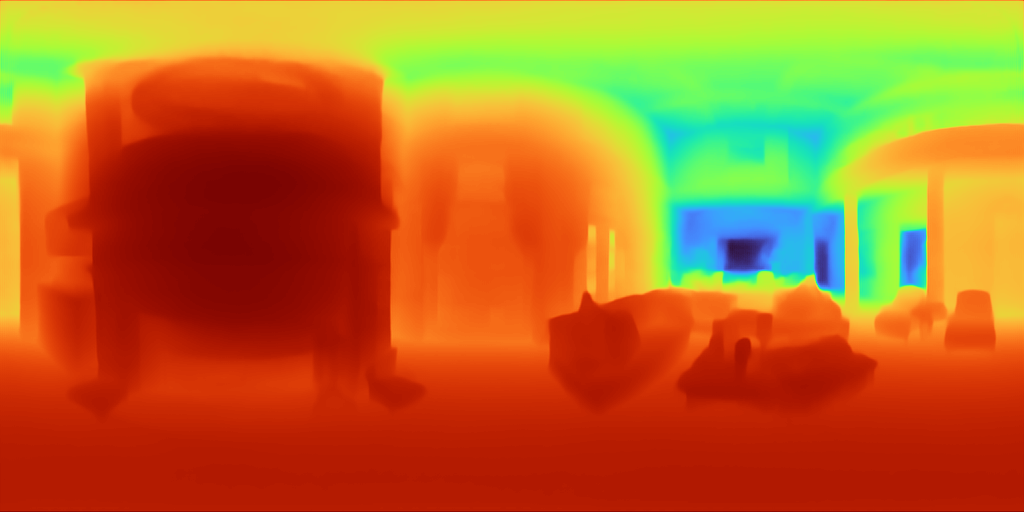}\hspace{-1cm} &
        \includegraphics[width=\mywidth]{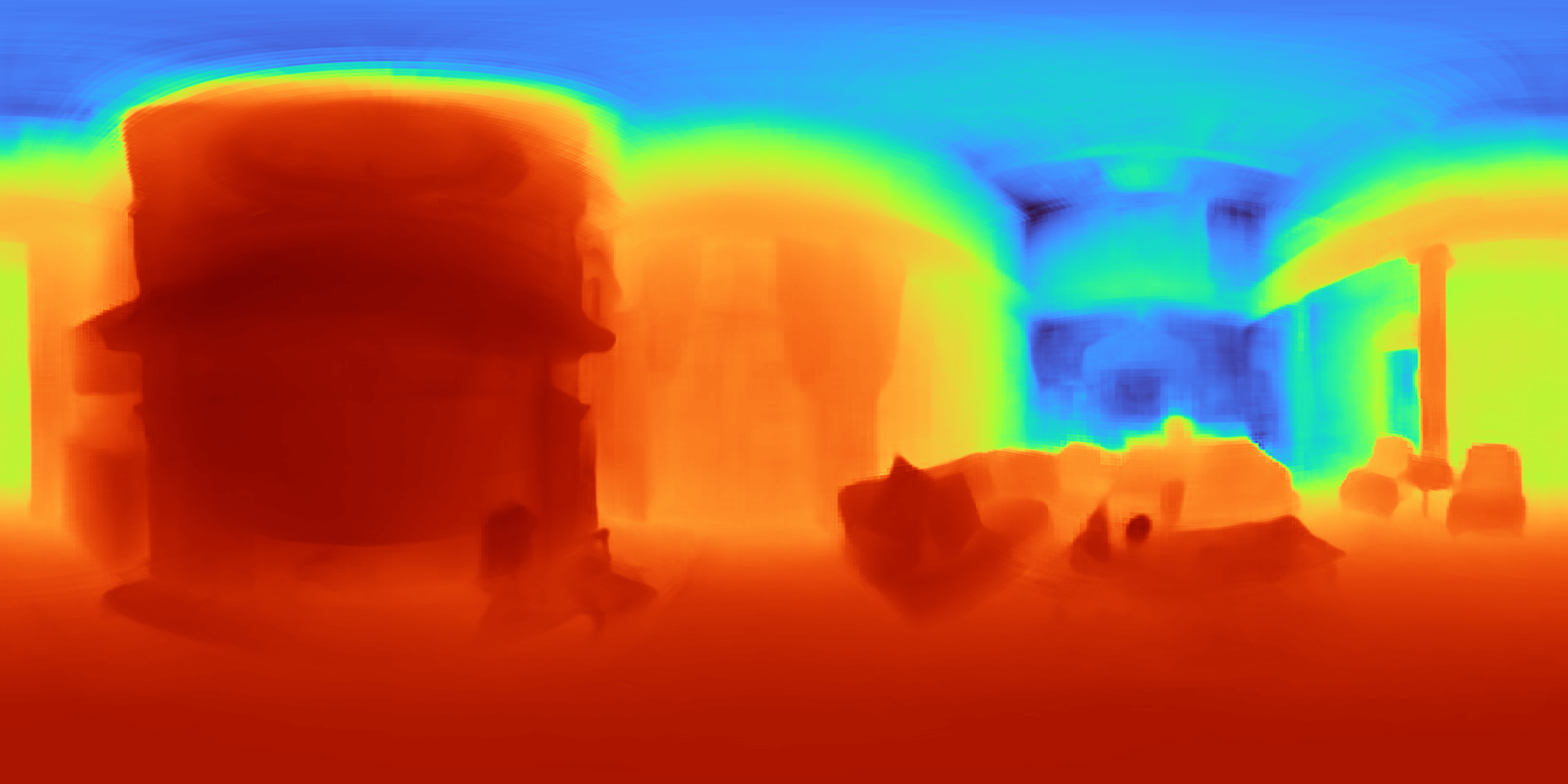}\hspace{-1cm} &
        \includegraphics[width=\mywidth]{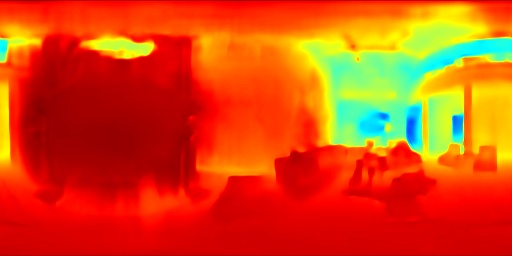}\hspace{-0.8cm} 
         \\

        \includegraphics[width=\mywidth]{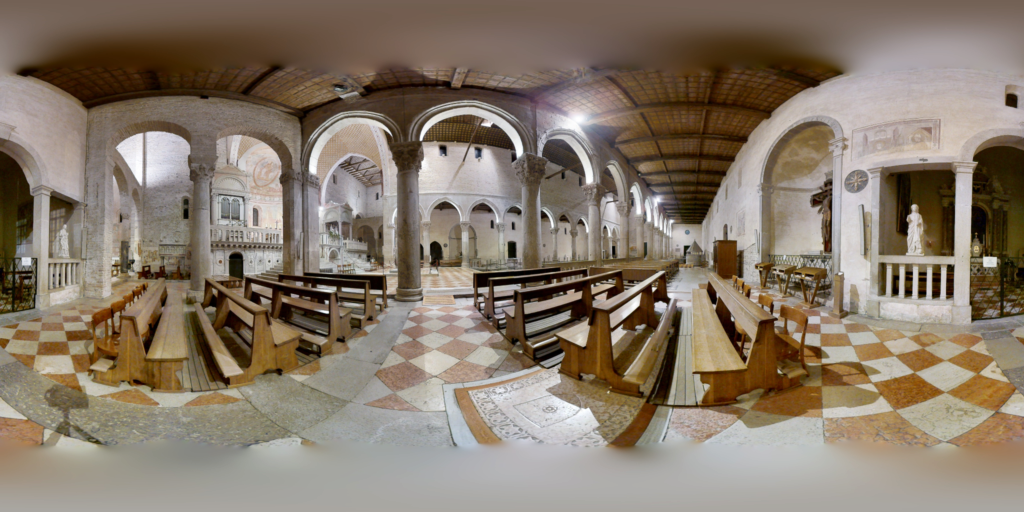} &
        \includegraphics[width=\mywidth]{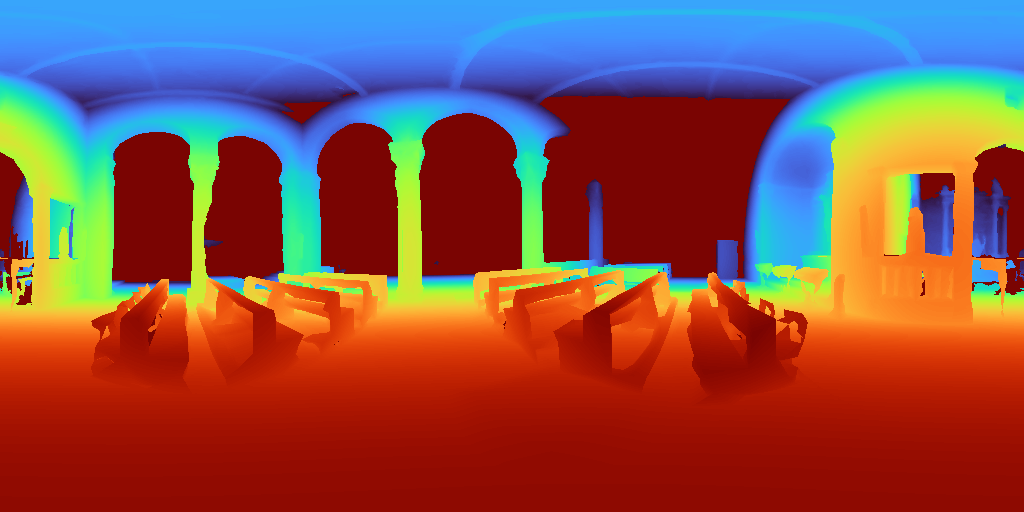} &
        \includegraphics[width=\mywidth]{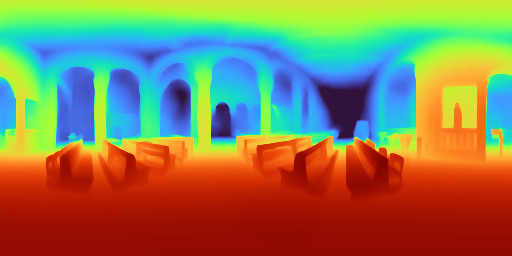} &
        \includegraphics[width=\mywidth]{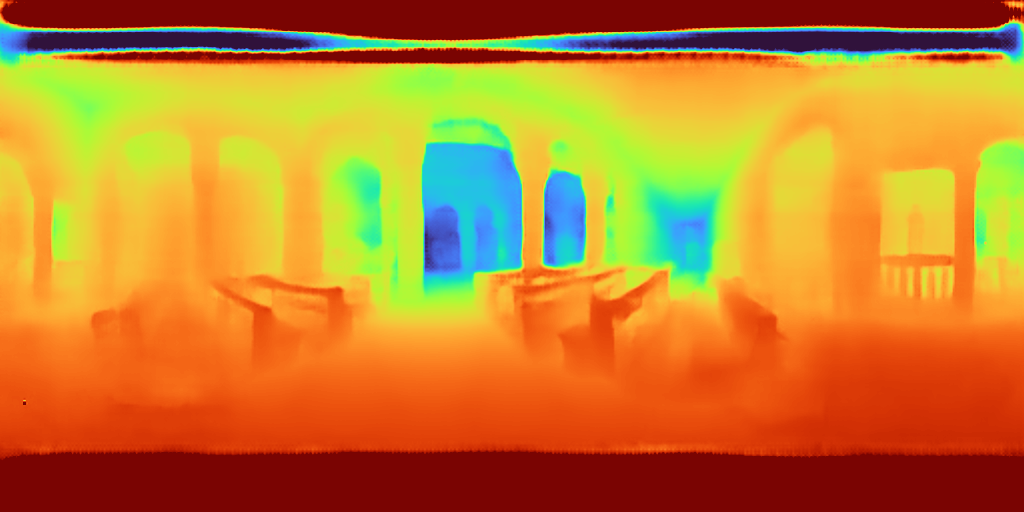} &
        \includegraphics[width=\mywidth]{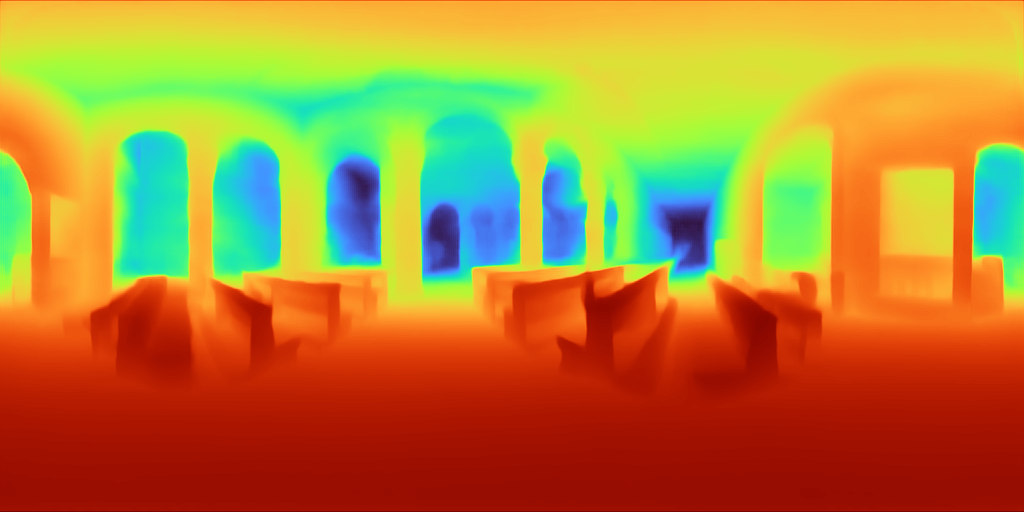} &
        \includegraphics[width=\mywidth]{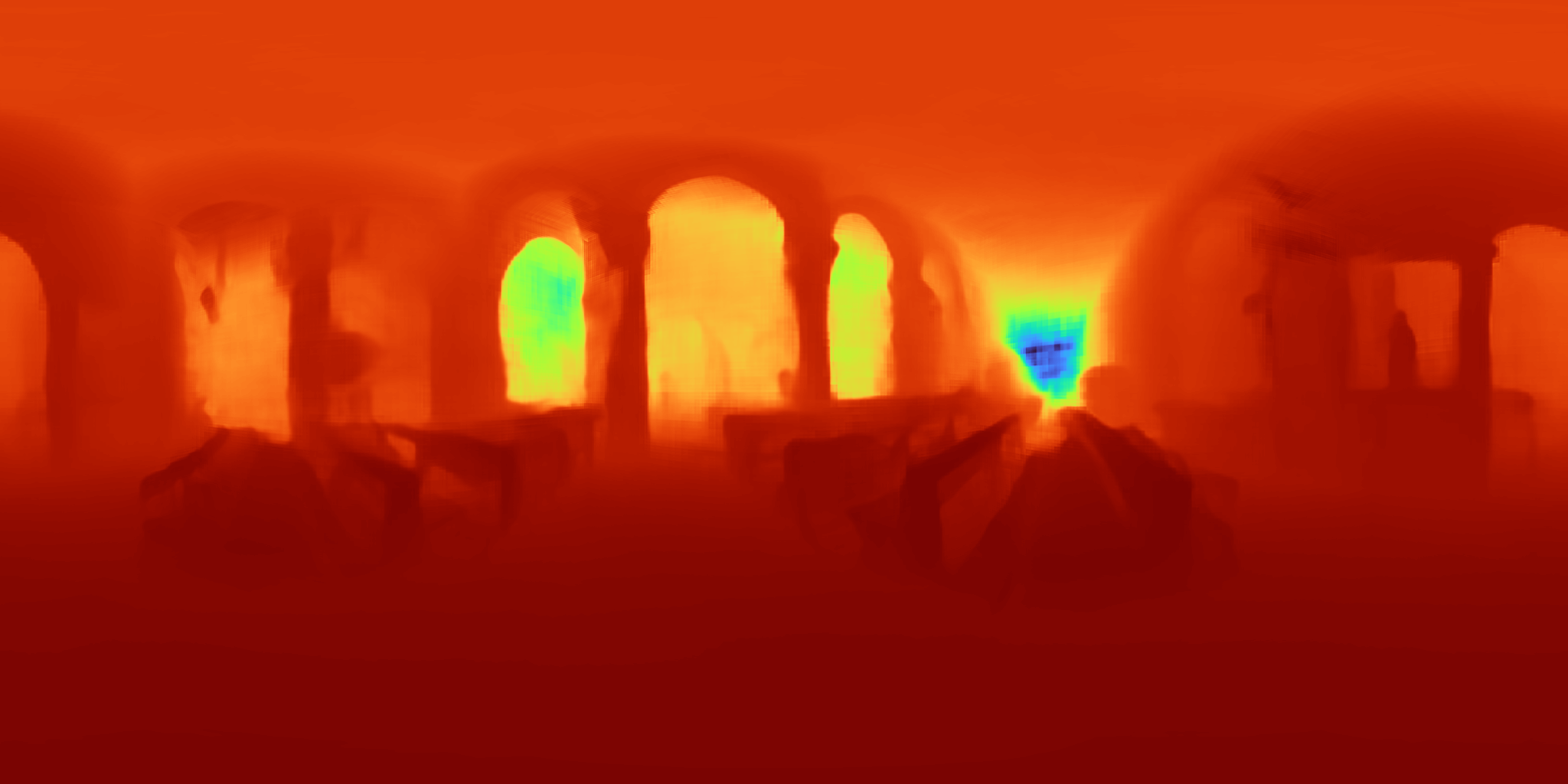} &
        \includegraphics[width=\mywidth]{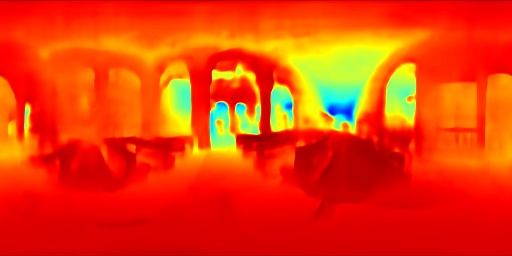} 
 \\

        \includegraphics[width=\mywidth]{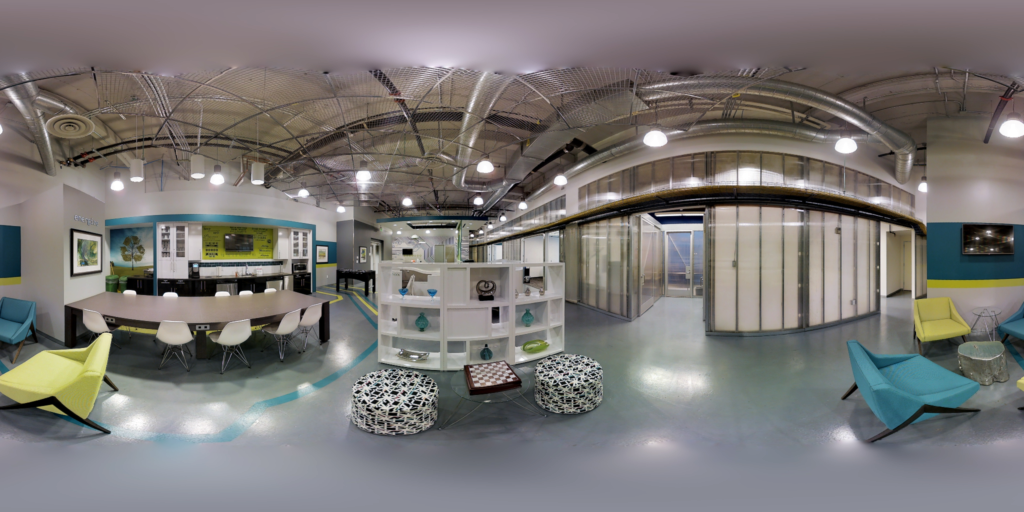} &
        \includegraphics[width=\mywidth]{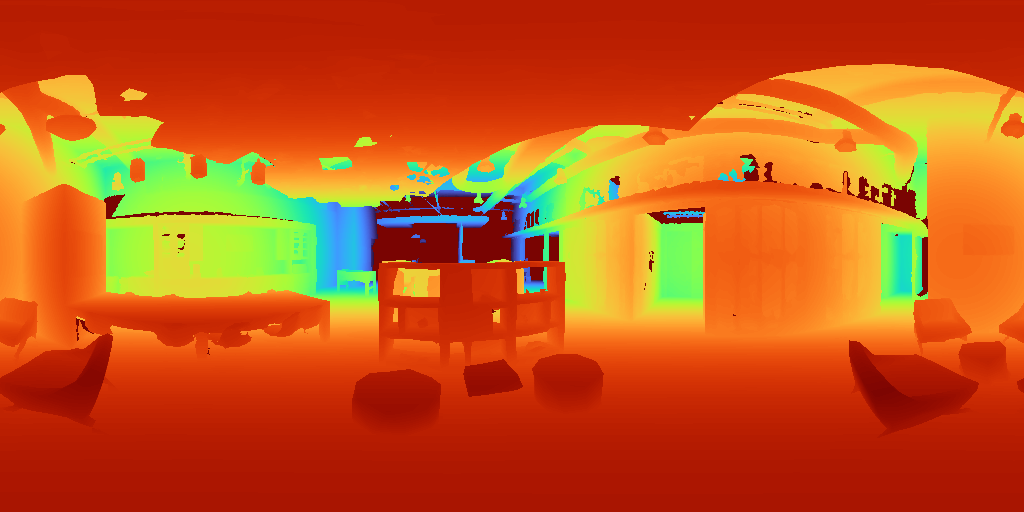} &
                \includegraphics[width=\mywidth]{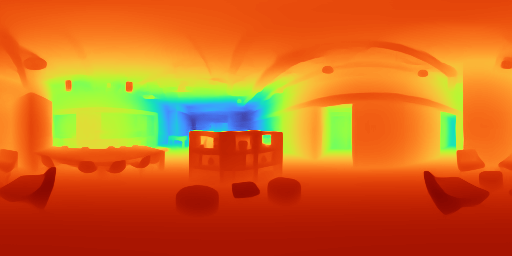}  &
        \includegraphics[width=\mywidth]{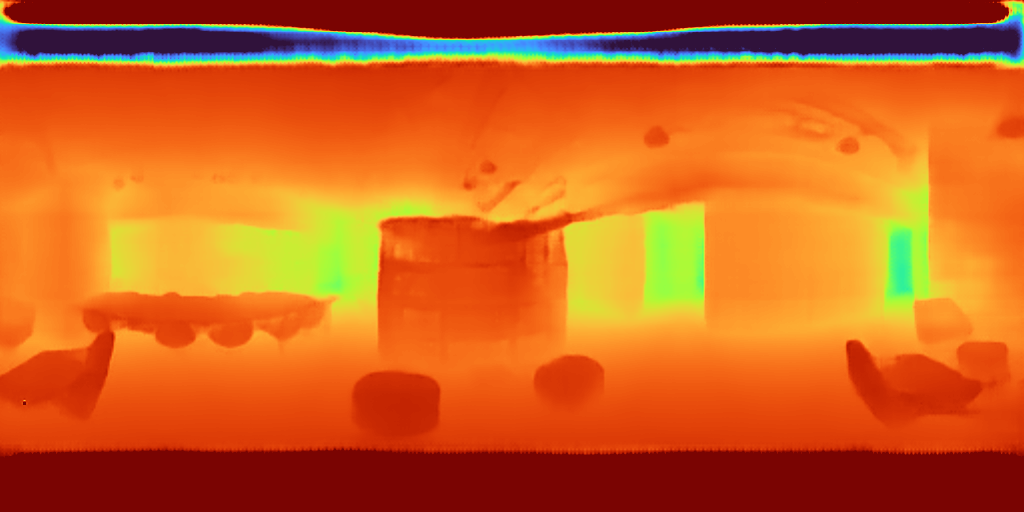} &
        \includegraphics[width=\mywidth]{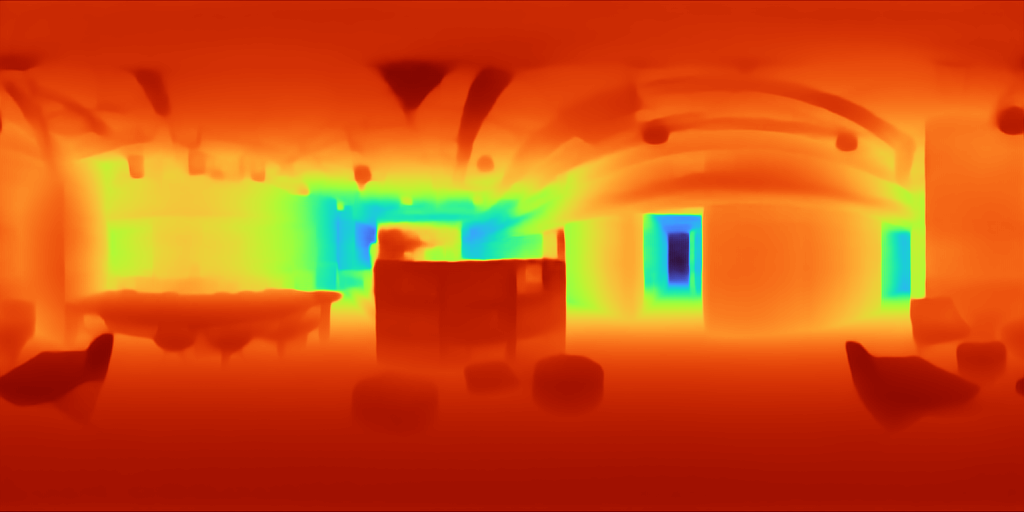} &
        \includegraphics[width=\mywidth]{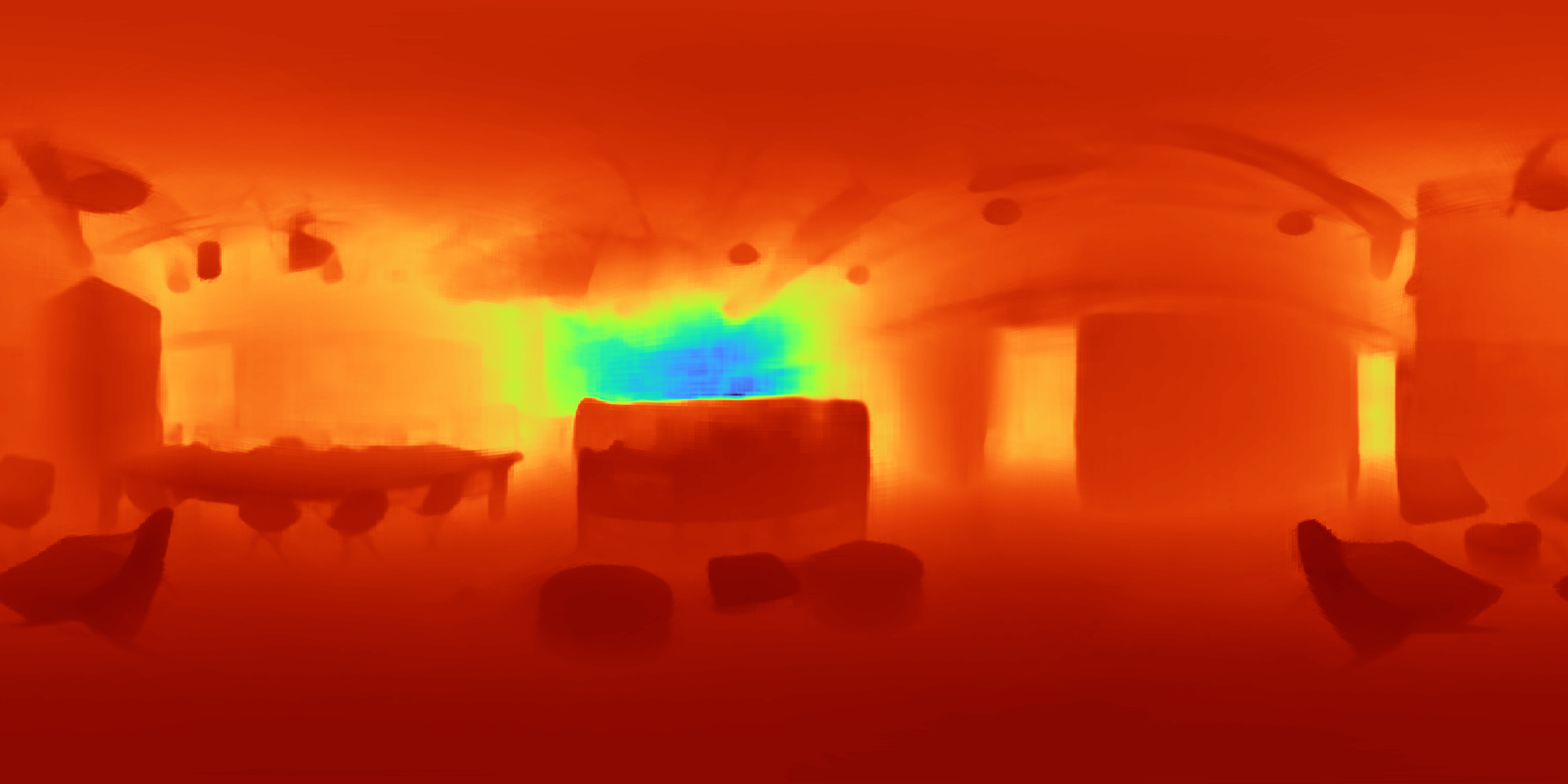} &
        \includegraphics[width=\mywidth]{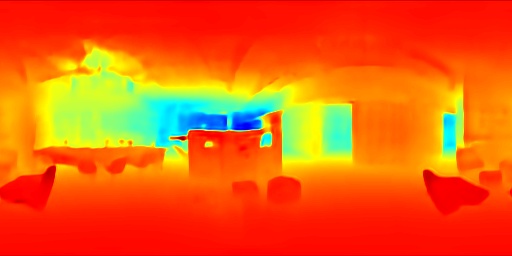} \\

        \includegraphics[width=\mywidth]{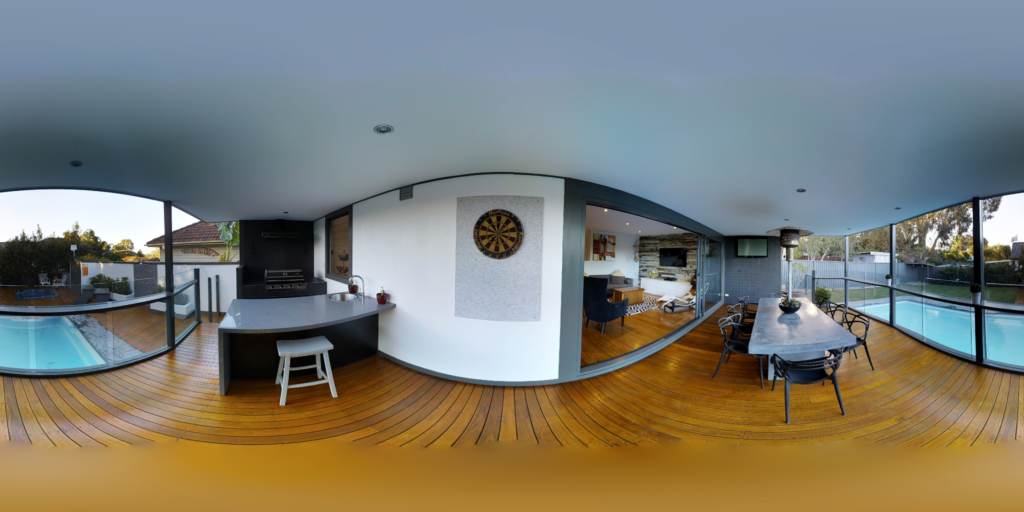} &
        \includegraphics[width=\mywidth]{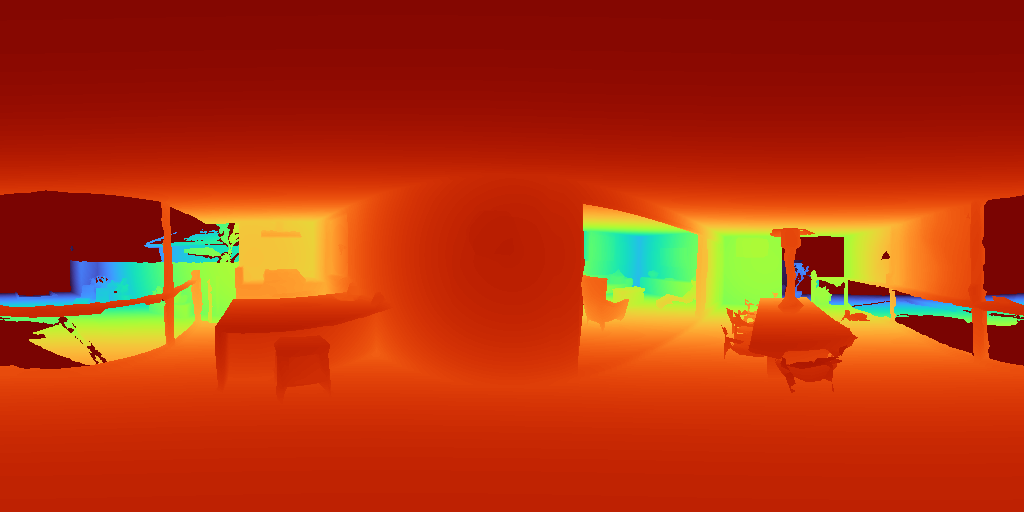} &
                \includegraphics[width=\mywidth]{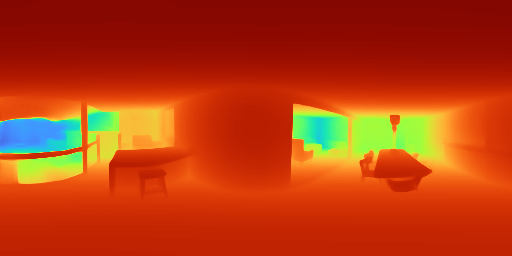} &
        \includegraphics[width=\mywidth]{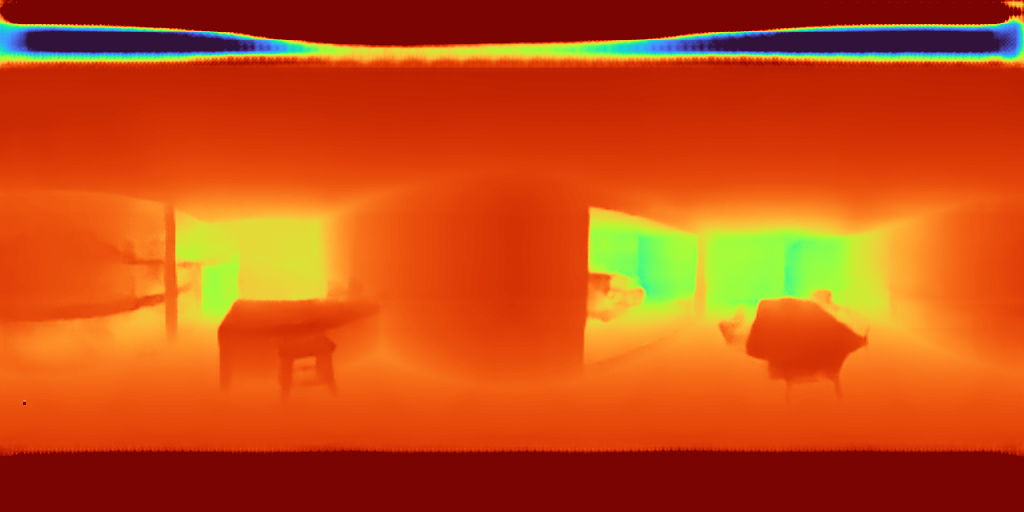} &
        \includegraphics[width=\mywidth]{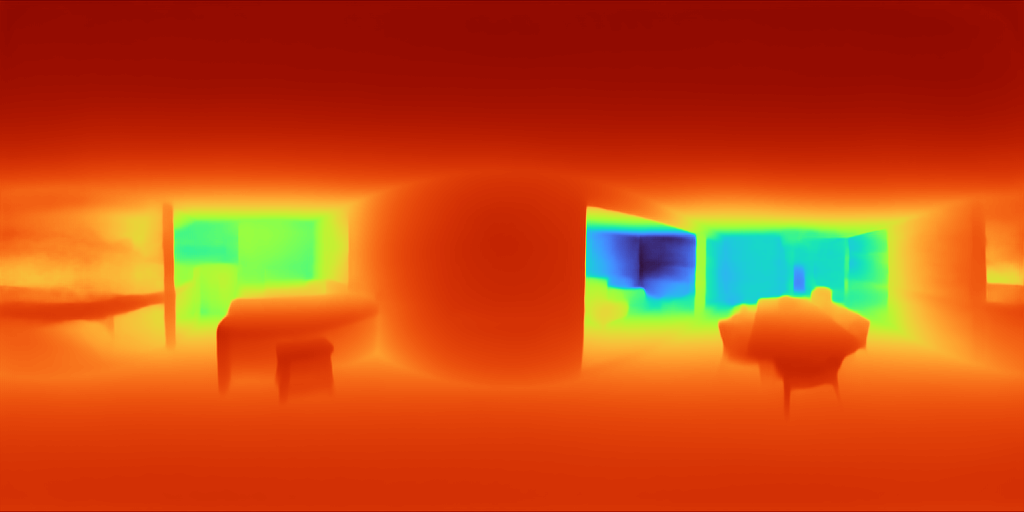} &
        \includegraphics[width=\mywidth]{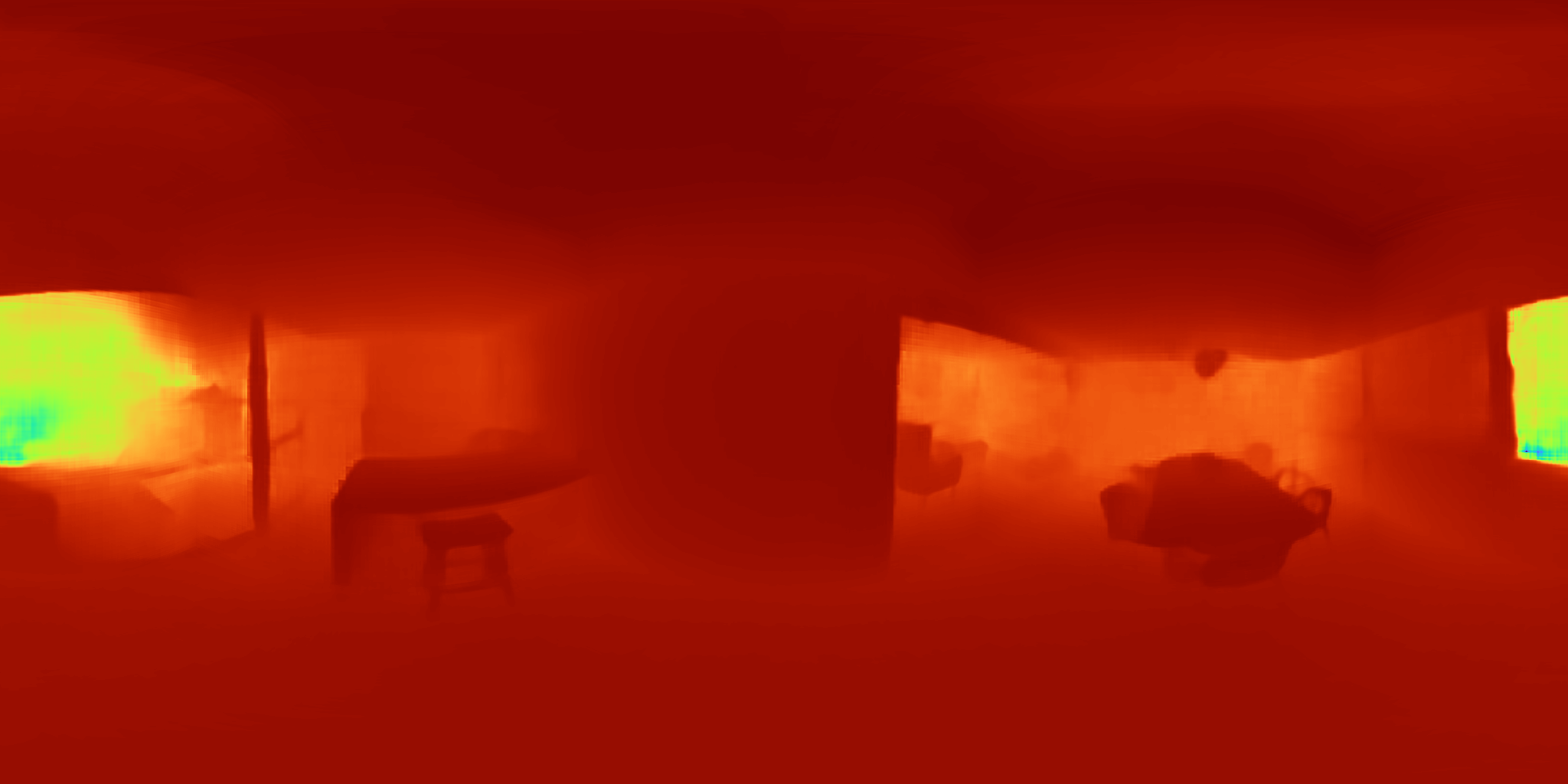} &
        \includegraphics[width=\mywidth]{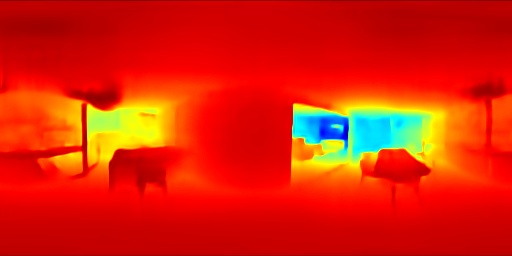} 
 \\

        \includegraphics[width=\mywidth]{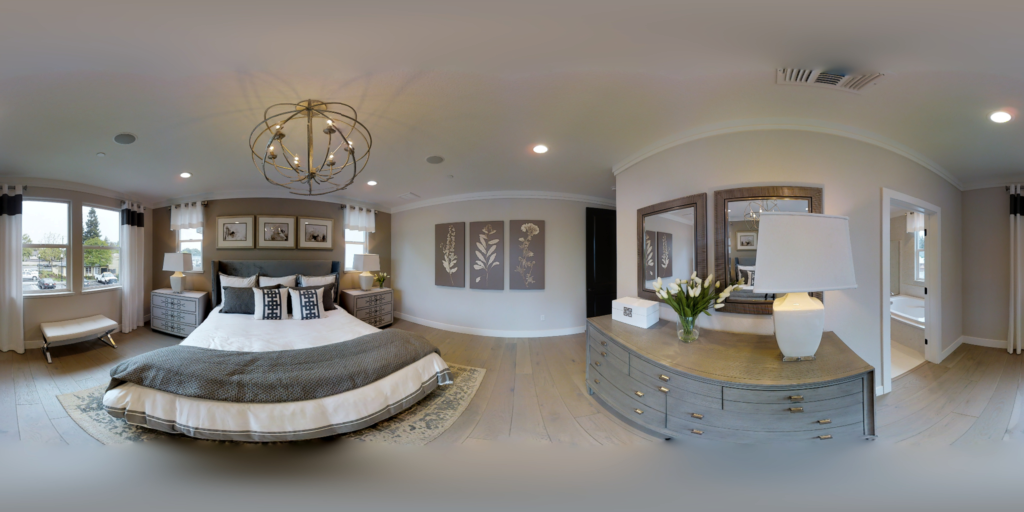} &
        \includegraphics[width=\mywidth]{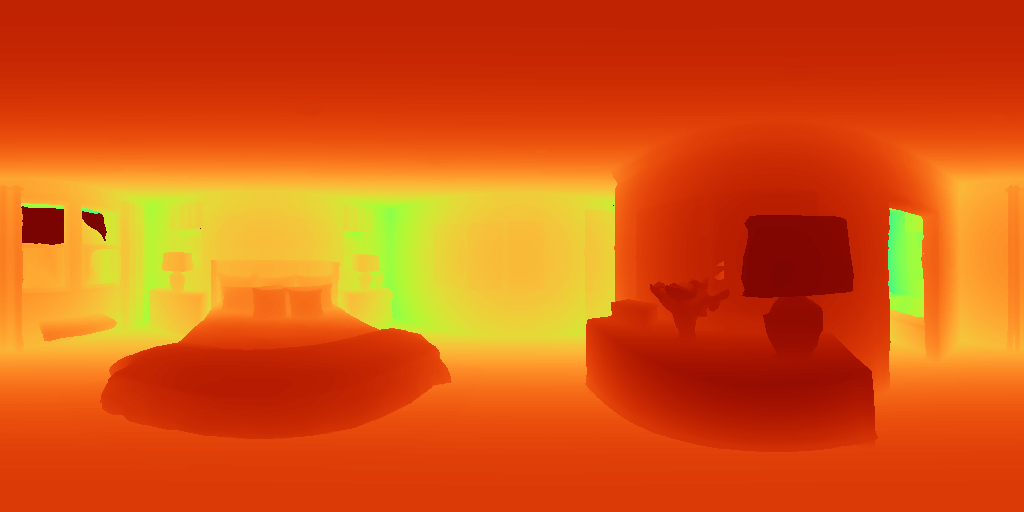} &
                \includegraphics[width=\mywidth]{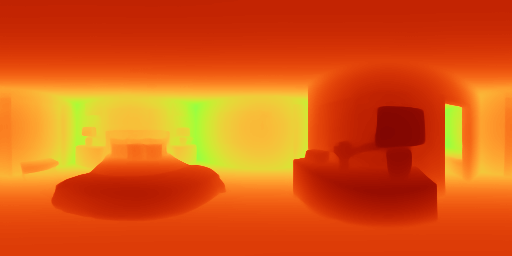} &
        \includegraphics[width=\mywidth]{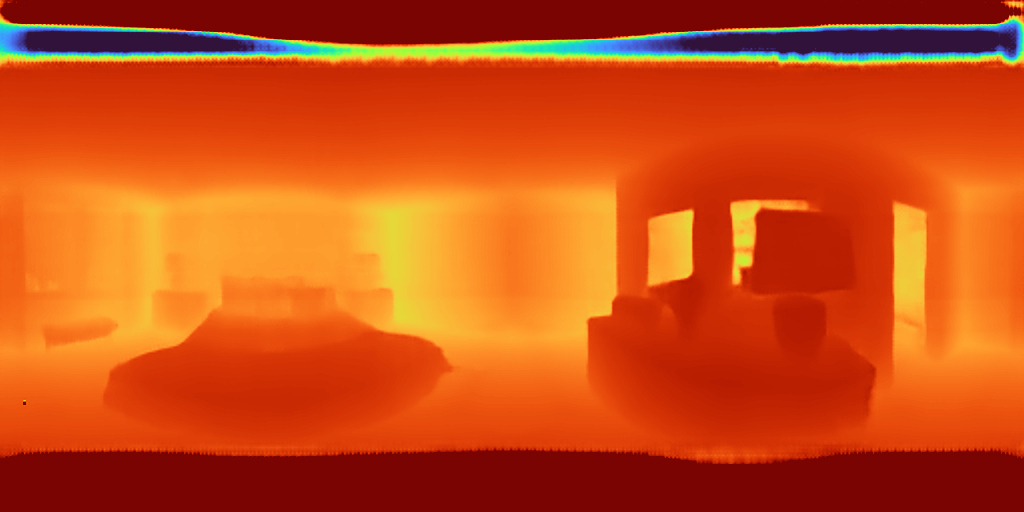} &
        \includegraphics[width=\mywidth]{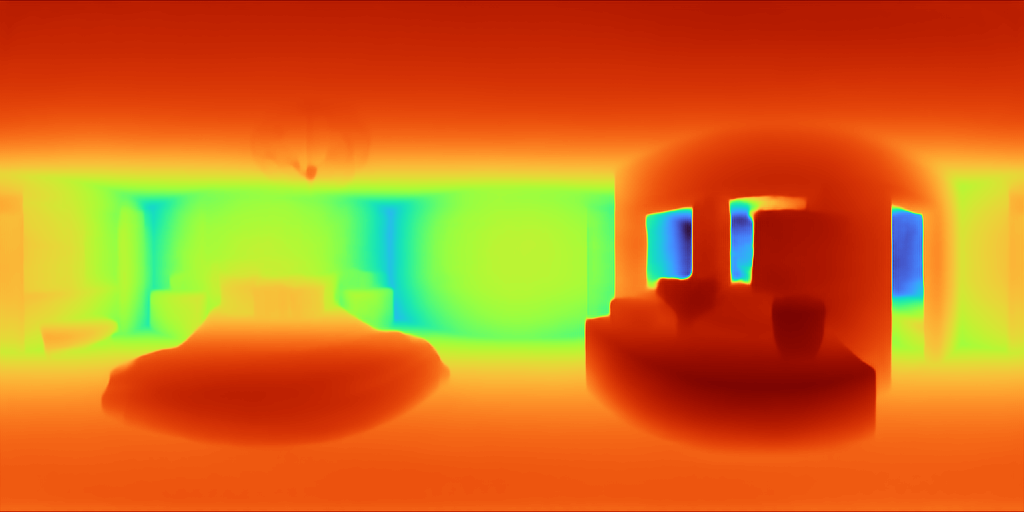} &
        \includegraphics[width=\mywidth]{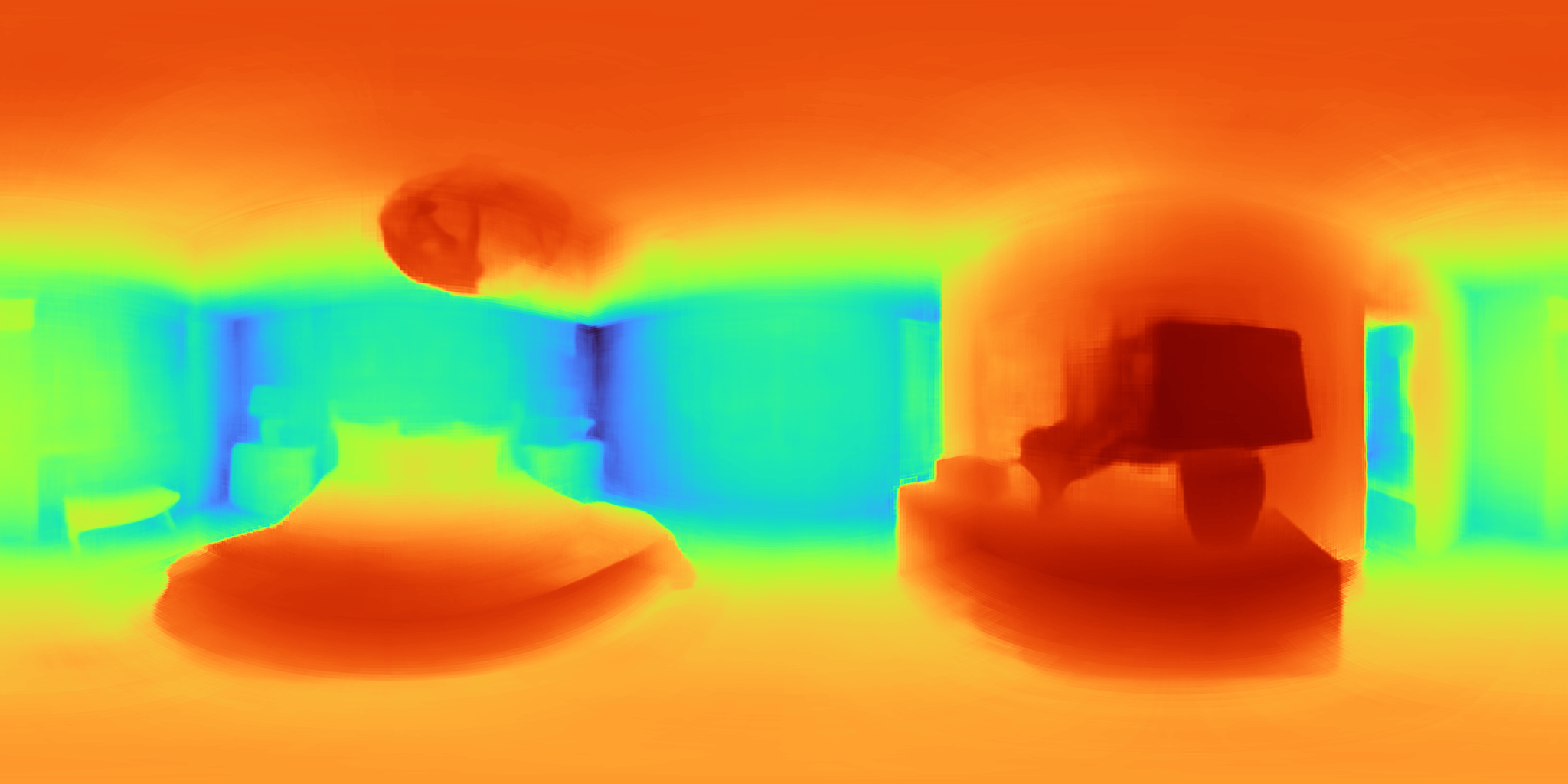} &
        \includegraphics[width=\mywidth]{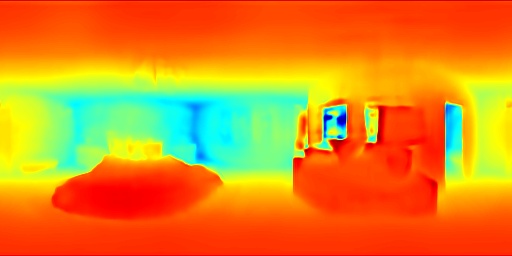} \\

        \includegraphics[width=\mywidth]{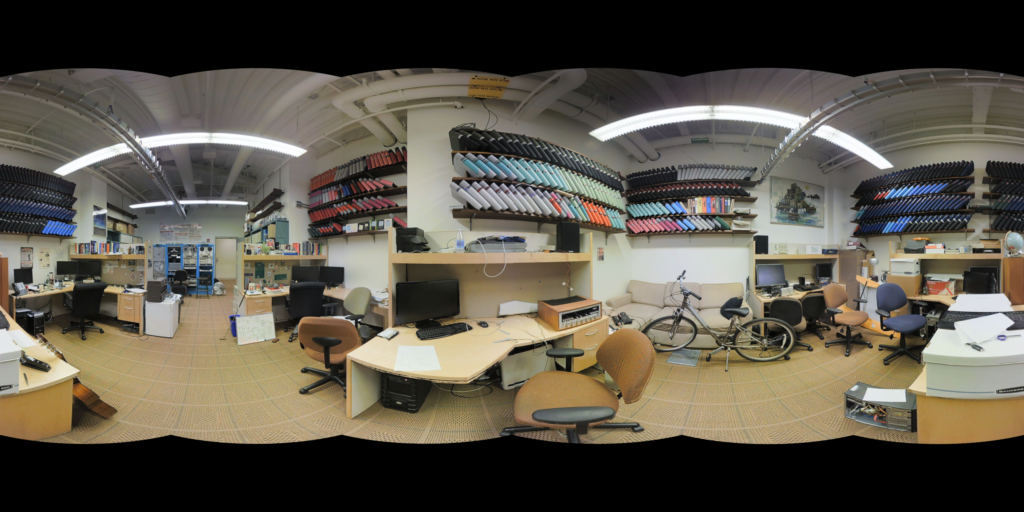} &
        \includegraphics[width=\mywidth]{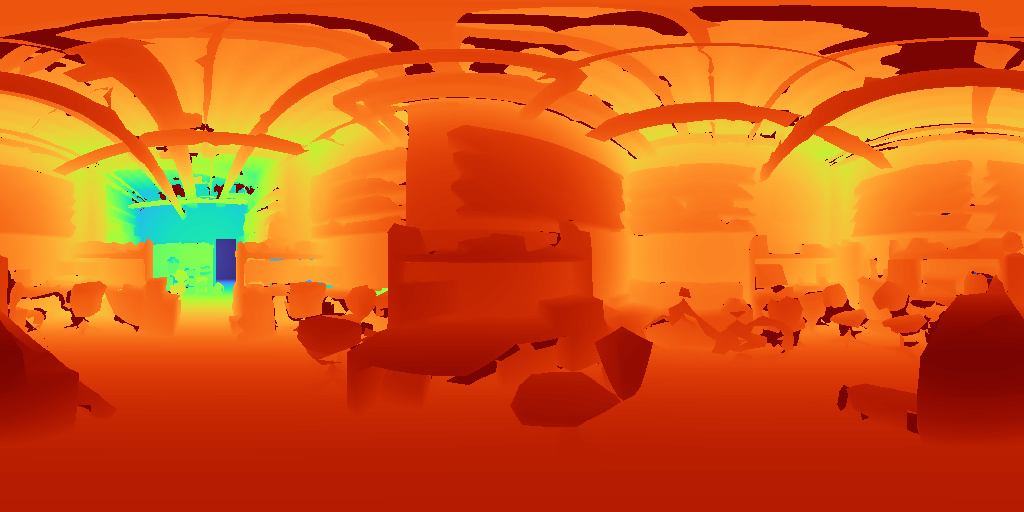} &
                \includegraphics[width=\mywidth]{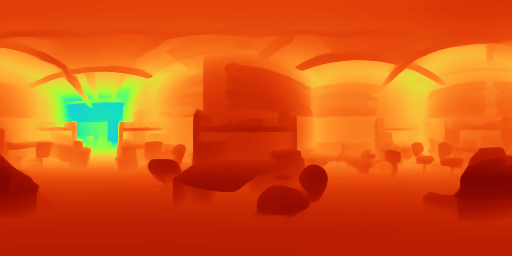} &
        \includegraphics[width=\mywidth]{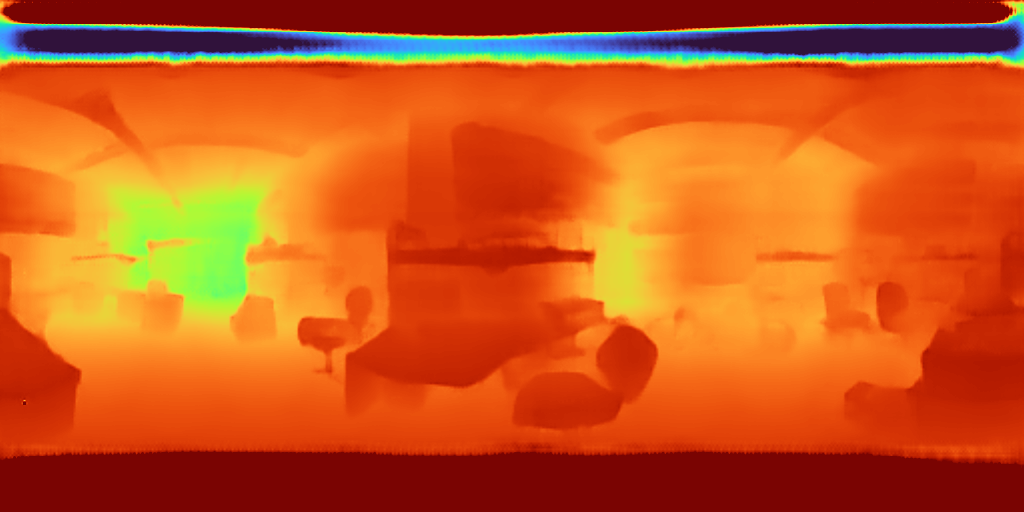} &
        \includegraphics[width=\mywidth]{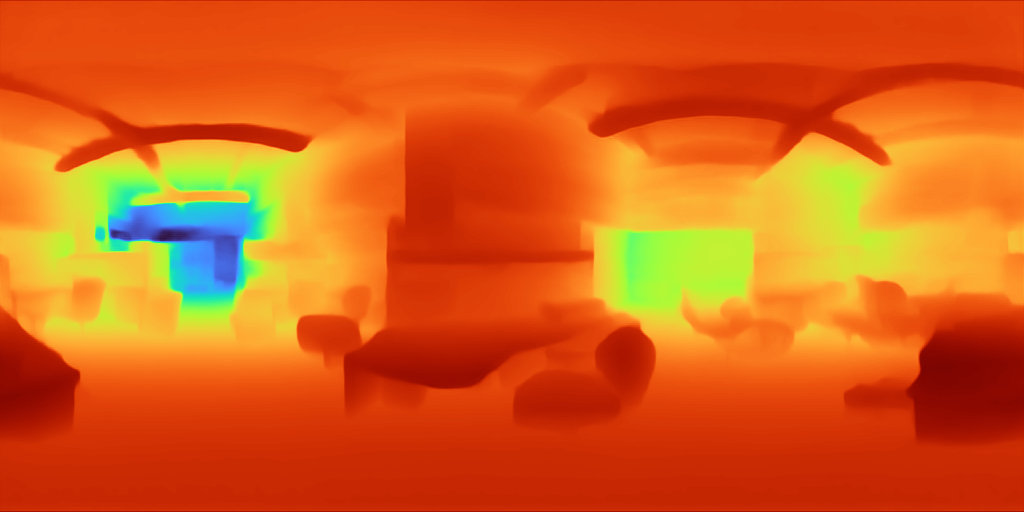} &
        \includegraphics[width=\mywidth]{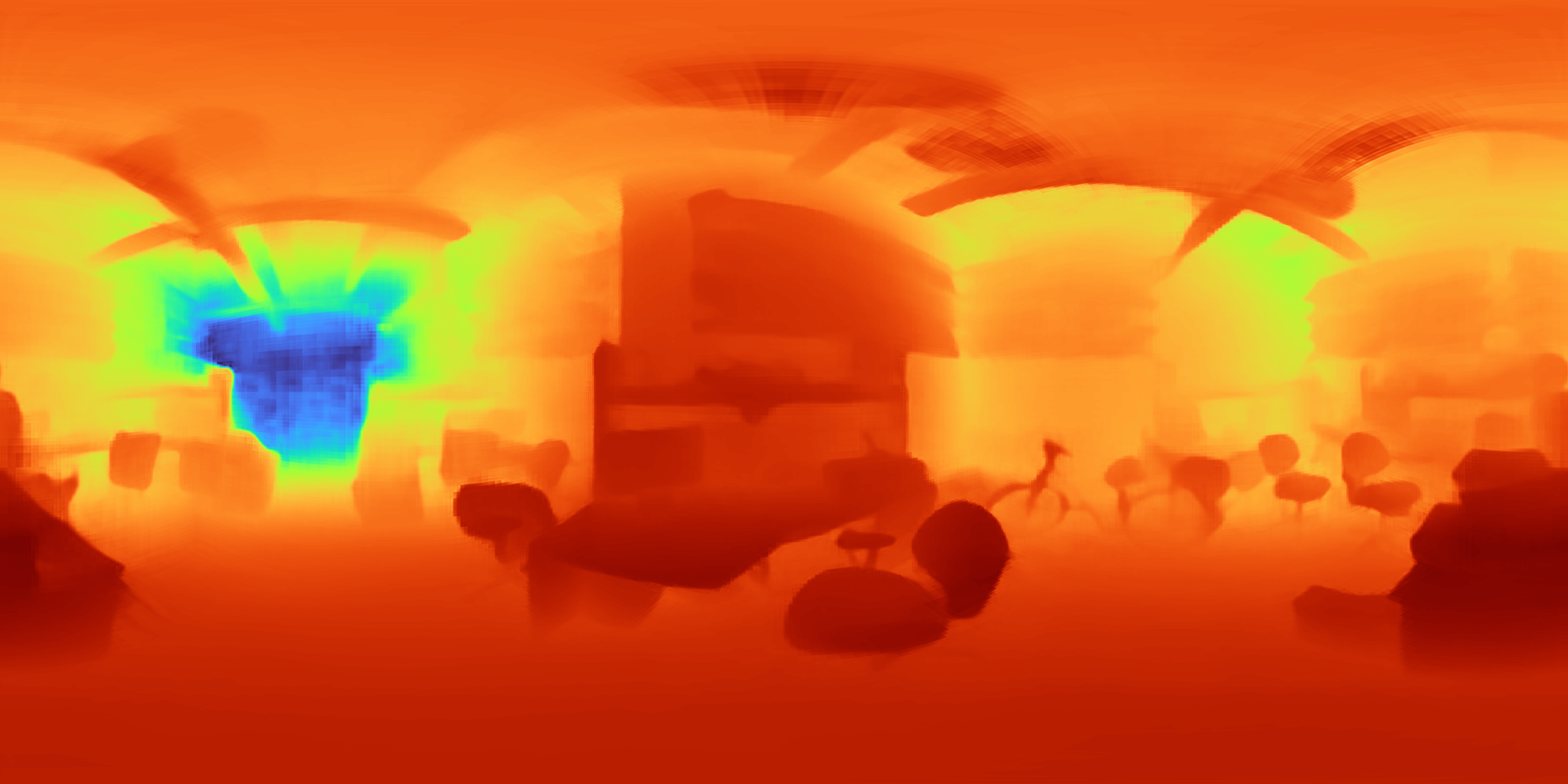} &
        \includegraphics[width=\mywidth]{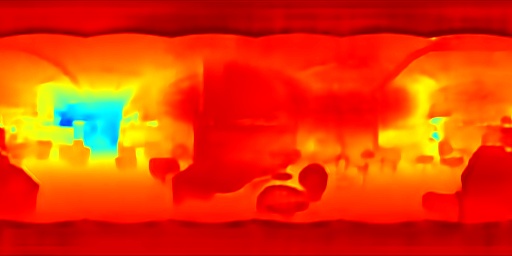} 
 \\
         RGB & GT & 360Recon & Bifuse++ & PanoFormer & FoVA-Depth & 360-MVSNet$^{*}$
    \end{tabular}
    \vspace{-8pt}
    \caption{\textbf{Depth Estimation Comparison.} The predicted depth maps are compared with various methods in different scenes. Our method provides more accurate and detailed depth estimates across these datasets.}
    \label{fig:depth_comparison}
    \vspace{-0.4cm}
\end{figure*}
\vspace{-0.25cm}
\subsection{3D Reconstruction}
\vspace{-0.15cm}
In this section, we evaluate the algorithm's reconstruction performance and compare it with state-of-the-art methods. Since no algorithm currently provides direct and complete reconstruction results, we input the depth maps estimated by these methods into the reconstruction module mentioned in \cref{sec:tsdf-reconstruction}, performing scene reconstruction and comparing the results with the ground truth mesh. 

\vspace{-0.08cm}
\textbf{Quantitative Experimental Analysis.} We compare our method with the current SOTAs, and the results are demonstrated in Tab.1. The F-scores, which represent the overall reconstruction performance, demonstrate that 360Recon achieves the best results across all datasets. Given that the Omniscenes dataset uses different 360° sensors compared to the Matterport3D dataset, this shows that our algorithm exhibits strong generalization and can be applied to different sensors. The Chamfer metric demonstrates that our algorithm exhibits superior geometric consistency, especially in comparison to monocular depth estimation methods like BiFuse++ and PanoFormer. Since 360-MVSNet$^{*}$ is trained on a self-constructed synthetic dataset under specific baseline settings, and the datasets used here consist of real-world images that may not align with these conditions, its performance is less effective, particularly as the scene scale increases.

\textbf{Qualitative Experimental Analysis.} As shown in~\cref{fig:reconstructions}, 360Recon demonstrates strong reconstruction performance across various environments. In small indoor scenes, such as bedrooms, its performance is comparable to other state-of-the-art algorithms in terms of overall reconstruction quality. However, 360Recon excels in capturing finer details, such as paintings on room walls, which other methods may miss. As the scene size increases, for example in hotel and living rooms, monocular depth estimation methods like BiFuse++ and PanoFormer struggle with depth consistency, leading to the loss of critical scene details—such as furniture and decorative items in hotel rooms and living spaces. In larger environments, such as the entrance hall and courtyard, existing algorithms fail to balance global scene reconstruction with local detail accuracy. For instance, the staircase in the entrance hall and the sofa in the corner are poorly reconstructed by all methods except 360Recon. Notably, 360Recon also excels in outdoor and complex environments, such as bathrooms with glass walls, where it achieves exceptional accuracy, even capturing challenging features like glass surfaces, which are typically difficult for reconstruction algorithms.
Additional qualitative results will be provided in the supplementary materials.

\vspace{-0.3cm}
\subsection{Panorama Depth Estimation}
\vspace{-0.2cm}
\begin{figure*}[t]  
    \centering
    \includegraphics[width=\textwidth]{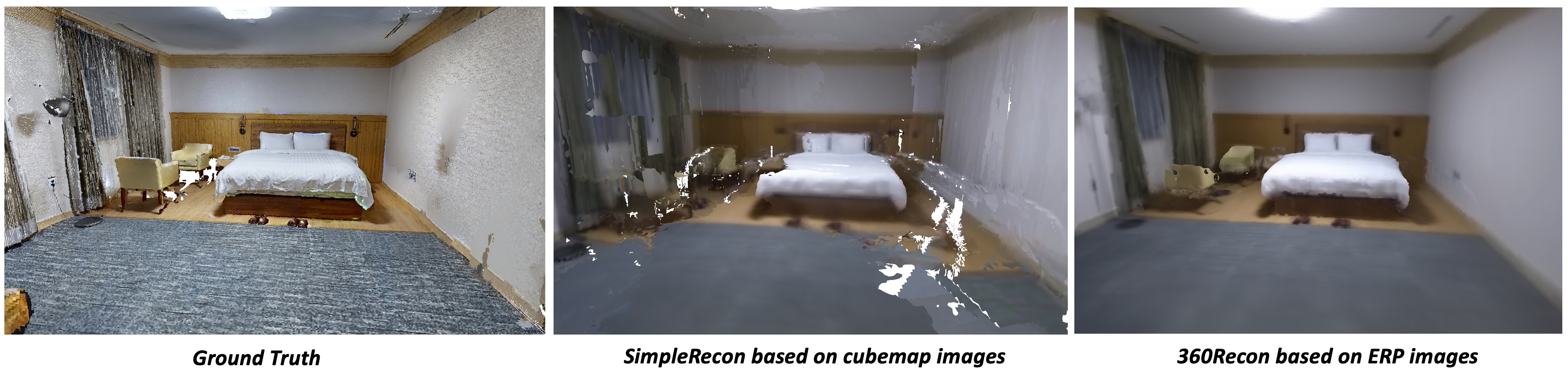}
    \caption{\textbf{Scene reconstruction performance under different 360° representations:} We use SimpleRecon for the cubemap representation and use 360Recon for the ERP format. }
    \label{fig:reconstructions_ablation}
    \vspace{-0.5cm}
\end{figure*}

To assess the algorithm's geometric consistency in local information, this section primarily analyzes the model's depth prediction capability. We compare the 360Recon with the other SOTAs. As shown in Tab. 2, our algorithm outperforms all other state-of-the-art methods across nearly all depth evaluation metrics. The RMSE metric, in particular, strongly demonstrates the precision of our algorithm in depth estimation, effectively highlighting that the proposed Spherical Feature Extraction module mitigates the impact of wide-angle lens distortion on the reconstruction algorithm.

As shown in~\cref{fig:depth_comparison}, our algorithm demonstrates better depth detail representation and improved geometric consistency. Since the data distribution does not always satisfy the specific baseline assumption required by 360-MVSNet, its performance is unstable, with some blurry regions in fine details. Meanwhile, both 360-MVSNet and BiFuse++ lack specific handling for image distortion areas, resulting in more blurriness at the top and bottom of the images. In terms of depth prediction, PanoFormer shows clearer local region details, but its monocular nature amplifies issues related to the lack of geometric consistency in the scene reconstruction, leading to poorer mesh quality. Our algorithm, on the other hand, eliminates distortion effects through the Spherical Feature Extraction on a single image, while fully leveraging the camera's large field of view through spherical sweeping. This ensures both detail accuracy and better geometric consistency in the final reconstruction. Further qualitative results can be found in the supplementary materials.

An additional key point to highlight is that our algorithm is both lightweight and efficient, with an average per-frame processing time of 112ms. The total model size is only 49.8M, which is even smaller than the parameter size of the monocular depth estimation method BiFuse++. Further tests regarding the algorithm’s efficiency will be provided in the supplementary materials.

\vspace{-0.2cm}
\subsection{Ablation Study}

\vspace{-0.2cm}
 \begin{table}[tp]	
\centering
\fontsize{5.3}{4}\selectfont
\renewcommand{\arraystretch}{1.5} 
\resizebox{0.9\linewidth}{!}{
\begin{tabular}{ccc}
	\toprule
	Method&without Spherical CNN&360Recon \cr
	\midrule
	
                             MAE$\downarrow$(cm)    &14.41 & \textbf{13.68}  \\
                             MRE$\downarrow$(cm)  &5.87  & \textbf{5.50}  \\
                             RMSE$\downarrow$(cm)   &37.71  & \textbf{36.22}  \\
                             $\delta_1\uparrow$(\%)  & 95.26 & \textbf{95.35} \\
                             chamfer$\downarrow$(cm)    &9.90& \textbf{9.68}  \\
                             F-score$\uparrow$    & \textbf{43.0} & 42.1  \\
        \midrule
	
\end{tabular}
}
\vspace{-0.3cm}
\caption{The qualitative comparison of depth estimation metrics. The best metric in each column is highlighted in bold.}
\vspace{-0.6cm}
\label{table:ablation_modules}
\end{table}

In this section, we evaluate our system from two aspects: (1) the impact of 360-degree image representation on overall reconstruction performance, and (2) the effect of different system modules on the overall reconstruction results.

For the first part, we convert 360-degree images from public datasets into Cubemap format and input them into an MVS network adapted for pinhole cameras to test the reconstruction performance. Since each image in the Cubemap is derived from the 360-degree image, it is free from lens distortion. The experimental results show that this reconstruction method fails in scenarios with low camera density or weak scene texture. 
Since our method is inspired by SimpleRecon, we chose it for the cubemap representation to better assess the effectiveness of this approach. We tested SimpleRecon on the Room 6 sequence from the OmniScene dataset, which features rich textures and a higher image capture frequency. However, both qualitative and quantitative results demonstrate that ERP achieves superior reconstruction performance, as shown in~\cref{fig:reconstructions_ablation} and \cref{table:ablation_cubemap}.

In the Cubemap representation, the smaller field of view of each image complicates the identification of common-view relationships, which results in numerous artifacts in the scene. In contrast, the ERP representation enables the algorithm to fully utilize the wide field of view, leading to improved reconstruction performance. This confirms that, for reconstruction tasks, the ERP format provides superior results for 360 images.

For the second part, we test the impact of each module of our algorithm on the overall reconstruction quality. The results presented at~\cref{table:ablation_modules} demonstrate that all modules contribute to the algorithm’s performance, significantly improving its geometric consistency and overall effectiveness. The RMSE value of the depth metric indicates that spherical features effectively mitigate the impact of distortion. Although the F-Score shows a slight decrease of 2.09\%, indicating a minor drop in reconstruction performance, the improvements in Chamfer distance (2.22\% increase) and RMSE (1.49\% improvement) demonstrate that the extracted spherical features play a significant role in enhancing geometric consistency in the final reconstruction. Additional ablation study results will be provided in the supplementary materials.

 \begin{table}[tp]	
\centering
\fontsize{5.3}{4}\selectfont
\renewcommand{\arraystretch}{1.5} 
\label{table:ablation_cubemap}
\resizebox{0.9\linewidth}{!}{
\begin{tabular}{ccc}
	\toprule
	Method&SimpleRecon&360Recon \cr
	\midrule
	
                             Comp$\downarrow$(cm)    &40.41 & \textbf{6.54}  \\
                             Acc$\downarrow$(cm)  &34.60  & \textbf{14.13}  \\
                             Chamfer$\downarrow$(cm)   &37.51  & \textbf{10.29}  \\
                             F-Score$\uparrow$  & 6.39 &  \textbf{49.18} \\
        \midrule
	
\end{tabular}
}
\vspace{-0.4cm}
\caption{The qualitative comparison of depth estimation metrics. The best metric in each column is highlighted in bold.}
\vspace{-0.6cm}
\end{table}

\vspace{-0.3cm}
\section{Conclusion}
\vspace{-0.2cm}
In this paper, we present a novel 360-image-based multi-view reconstruction approach. The algorithm demonstrates state-of-the-art performance in evaluations on publicly available datasets. The proposed spherical feature extraction module effectively mitigates the distortion effects caused by wide-angle lenses. Additionally, the fusion of the 3D cost volume with image-enhanced features obtained through 2D CNNs further improves the reconstruction quality, providing both high accuracy and better local detail representation.

\textbf{Limitation and future work.} 360Recon currently cannot achieve complete scene reconstruction. When objects in the scene are occluded, it is unable to rely on the partially observed data to reconstruct the missing information. In future work, we plan to integrate generative algorithms to enable more comprehensive scene reconstruction.

{
    \small
    \bibliographystyle{ieeenat_fullname}
    \bibliography{main}
}
\clearpage
\setcounter{page}{1}
\setcounter{figure}{0}
\setcounter{table}{0}
\setcounter{section}{0}
\renewcommand\thesection{\Alph{section}}
\maketitlesupplementary
This supplementary material provides additional information to complement our main paper. \cref{section:Details} outlines the network architecture and implementation details of our approach. \cref{Datasets} offers an overview of the datasets utilized in our research. \cref{efficiency} presents an analysis and comparison of the model's efficiency. \cref{ablation} presents extended ablation studies, encompassing further analysis of component design. \cref{Qualitative} features additional qualitative comparisons between our model and methods. In \cref{collected}, we tested the generalization and robustness of our model on a dataset we collected. Finally, \cref{disscusion} summarizes our work, discusses its limitations, and proposes potential directions for future research.

\section{Implementation Details}
\label{section:Details}
As mentioned in \cref{experimets}, we used an NVIDIA 3090 GPU for training. We set the batch size to 4 and the learning rate to \(1 \times 10^{-4}\). We used half-precision computations during both training and testing. The total number of training steps was 50,000, corresponding to approximately 25 epochs. We reduced the learning rate by a factor of ten at steps 30,000 and 40,000. The total training time was approximately 12 hours.

To enhance image features in the encoder, we employed a pre-trained EfficientNetV2 network~\cite{efficientNet}. For the matching feature extractor, we incorporated spherical CNN layers into ResNet~\cite{Resnet}. Portions of the spherical convolution code were referenced from MODE~\cite{MODE} and utilized CUDA acceleration to ensure computational efficiency. For the Encoder-Decoder network structure that integrates the cost volume and image-enhanced features to produce the final depth estimation, we modified the U-Net architecture~\cite{Unet} accordingly.

The input tuple to our model contains three images: one reference frame for depth estimation and two source frames for multi-view constraints. All input images to the model are standardized to a resolution of 512×1024 pixels. During depth estimation, the resolution is halved. Regarding the construction of the cost volume based on depth hypotheses, we assume a minimum depth of 0.25 meters and a maximum depth of 8 meters. We hypothesize 64 depth planes, with depth values following a logarithmic uniform distribution.
\section{Datasets Details}
\label{Datasets}
\begin{table}[tp]	
\centering
\fontsize{5.3}{4}\selectfont
\renewcommand{\arraystretch}{1.5} 
\resizebox{0.9\linewidth}{!}{
\begin{tabular}{cccc}
	\toprule
	Method&Parameter(M)&Per-Frame Time(ms)&Memory(MB) \cr
	\midrule
	
                             Bifuse++\cite{Bifusev2}    &52.49 & 14 &902 \\
                             PanoFormer\cite{panoformer}  &20.38  & 36 &2204 \\
                             FoVA-Depth\cite{fova}   &37.31  & 59&5928  \\
                             360MVSNet\textsuperscript{*}\cite{360mvsnet,panogrf}    &61.82& 154&3956  \\
                             Ours    & 49.19 & 112&2892  \\
        \midrule
	
\end{tabular}
}
\vspace{-0.3cm}
\caption{\textbf{Efficiency comparison between our algorithm and other algorithms.} Comparison items include the size of parameters, per-frame inference time during depth estimation, and total GPU memory consumption during model inference.}
\vspace{-0.2cm}
\label{table:efficiency}
\end{table}
For Matterport3D~\cite{Matterport3D} and Standford2D3D~\cite{S2D3D} datasets, we used the original panoramic images downloaded from the official website, rather than the re-rendered versions. For the Matterport3D~\cite{Matterport3D} and Standford2D3D~\cite{S2D3D} datasets, we strictly adhere to the dataset splits used in previous works~\cite{BiFuse,Bifusev2,panoformer}. We trained on the 61 training scenes of Matterport3D~\cite{Matterport3D} and conducted depth testing on the test sets of both datasets. For the OmniScenes dataset~\cite{OmniScenes}, since it lacks depth information that directly corresponds to each frame, we do not use it for evaluation of depth estimation.

When evaluating 3D reconstruction, we employed the method in ~\cite{transformerfusion} across all three datasets. We do not use any masks and set the threshold distance to 5 cm when calculating the F-Score. For the Matterport3D dataset~\cite{Matterport3D}, we selected 16 scenes out of the 18 test scenes for 3D reconstruction evaluation. For the Stanford2D3D dataset~\cite{S2D3D}, we evaluated both scenes in the test set. For OmniScenes~\cite{OmniScenes}, we selected 17 scenes for evaluation and performed reconstruction using data collected by the TurtleBot.

\section{Efficiency Analysis}
\label{efficiency}

\begin{figure*}[!h]

  \centering
  \captionsetup{aboveskip=-8pt}
  \begin{tabular}{@{}lcccc@{}}
    \raisebox{1cm}{\rotatebox{90}{\textbf{RGB}}} &
    \begin{subfigure}[b]{0.22\textwidth}
      \includegraphics[width=\linewidth]{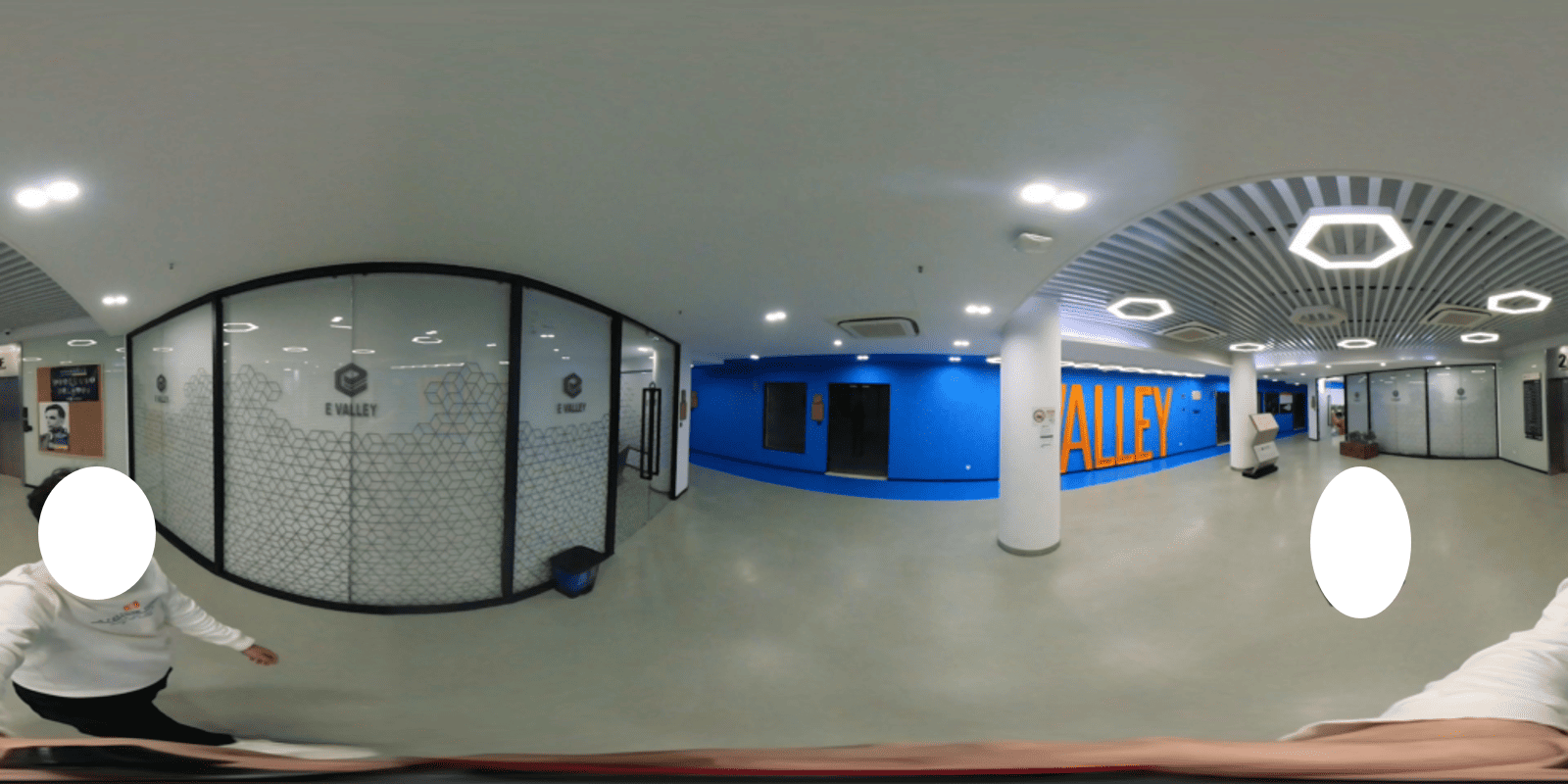}
      \label{fig:img1}
    \end{subfigure} &
    \begin{subfigure}[b]{0.22\textwidth}
      \includegraphics[width=\linewidth]{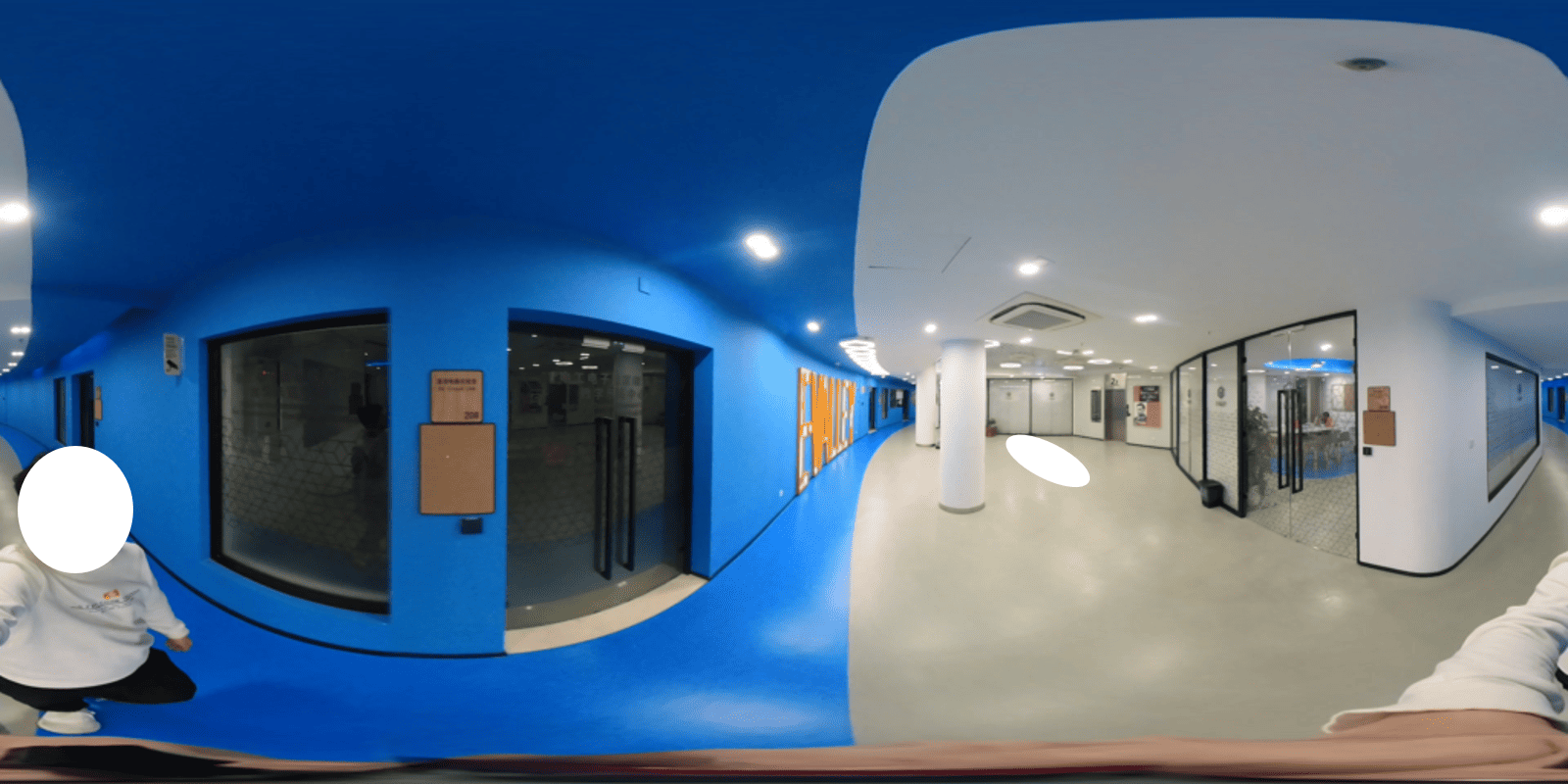}
      \label{fig:img2}
    \end{subfigure} &
    \begin{subfigure}[b]{0.22\textwidth}
      \includegraphics[width=\linewidth]{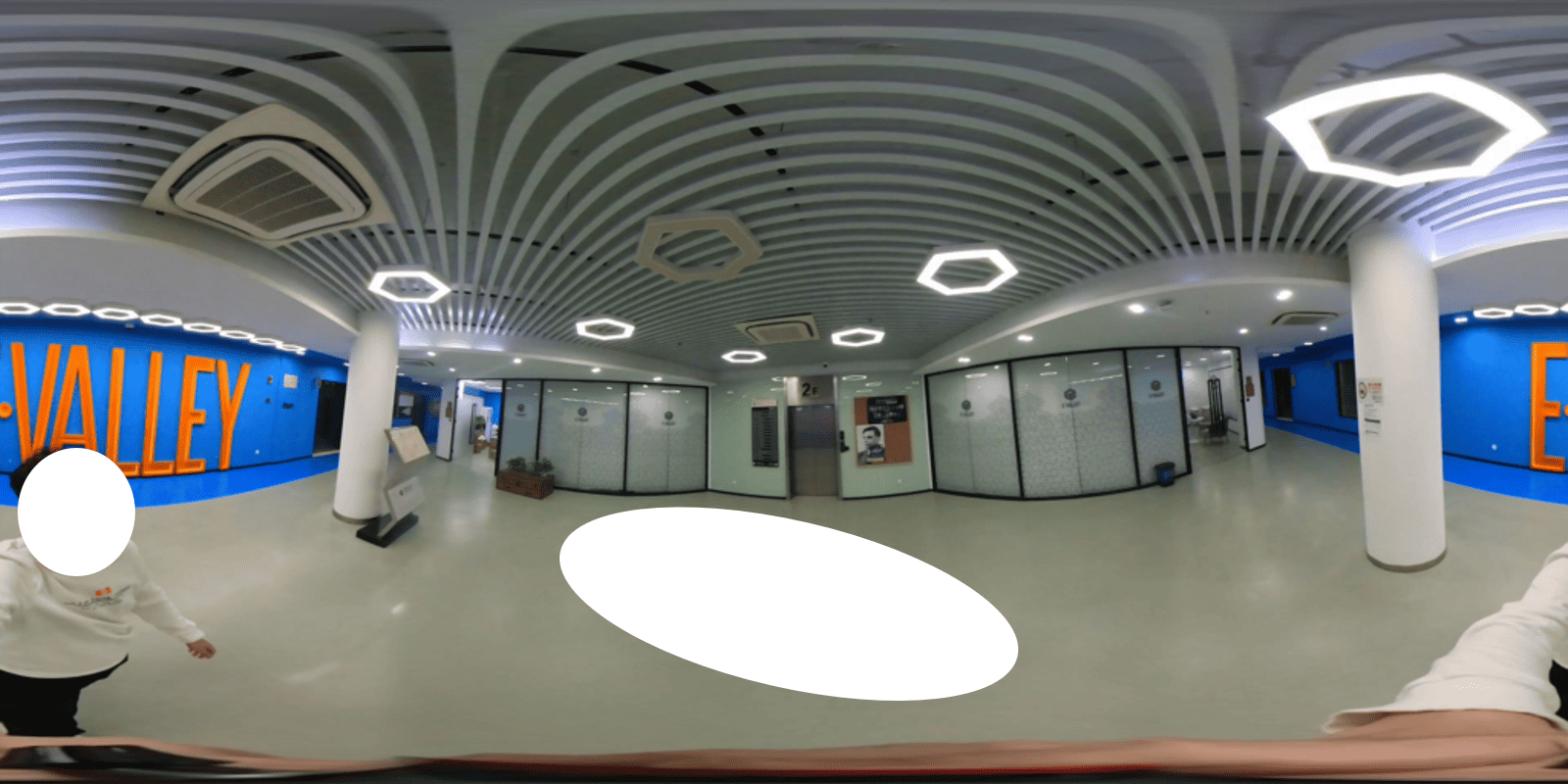}
      \label{fig:img3}
    \end{subfigure} &
    \begin{subfigure}[b]{0.22\textwidth}
      \includegraphics[width=\linewidth]{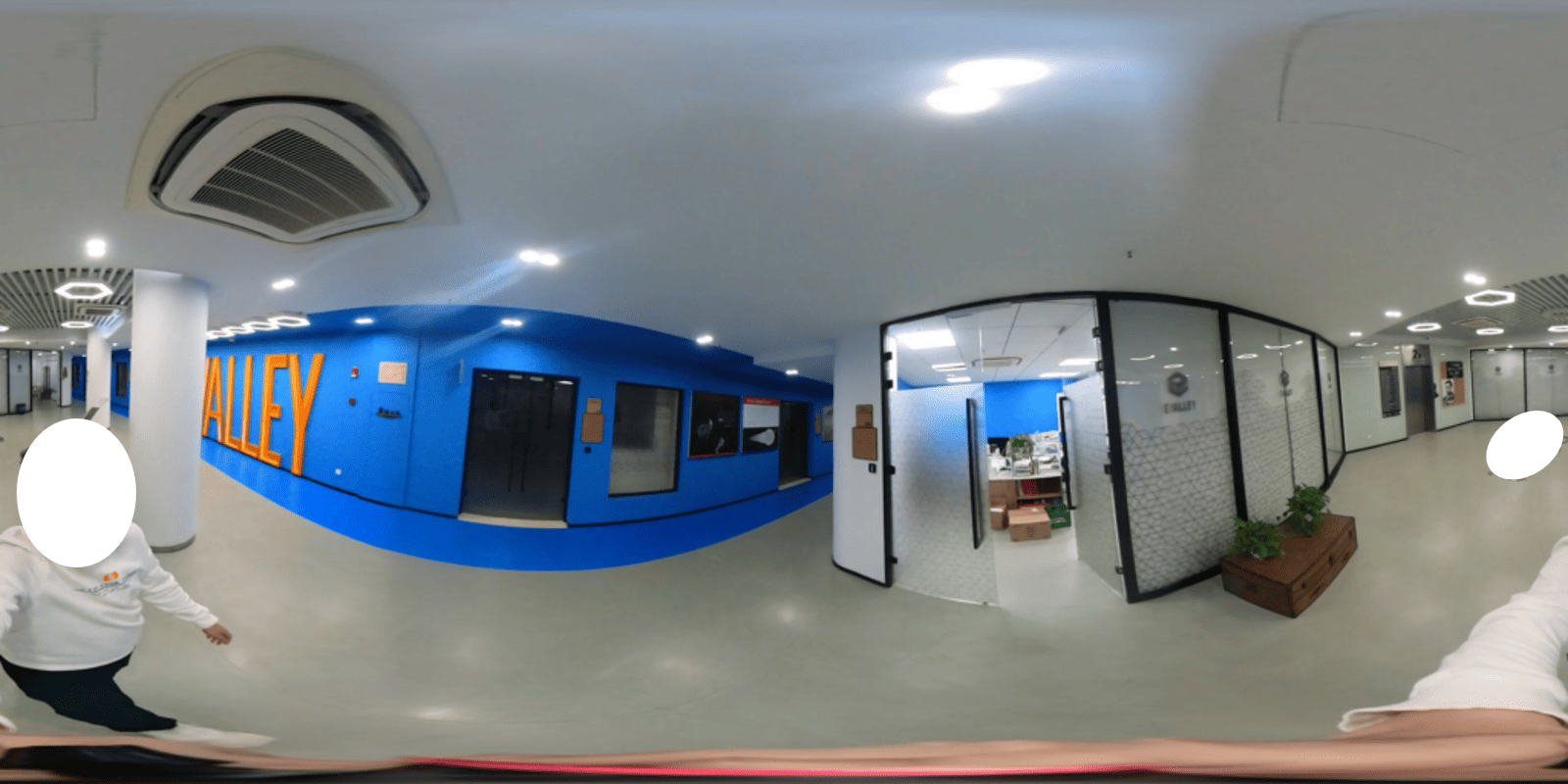}
      \label{fig:img4}
    \end{subfigure} \\
    
    \raisebox{0.9cm}{\rotatebox{90}{\textbf{Depth}}} &
    \begin{subfigure}[b]{0.22\textwidth}
      \includegraphics[width=\linewidth]{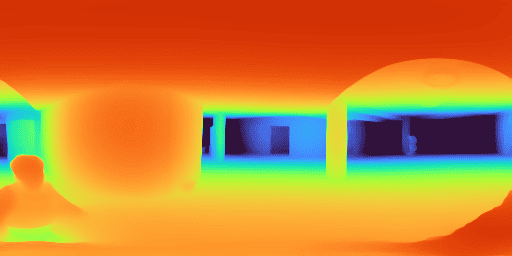}
      \label{fig:img5}
    \end{subfigure} &
    \begin{subfigure}[b]{0.22\textwidth}
      \includegraphics[width=\linewidth]{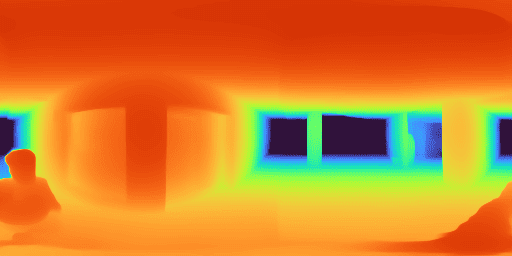}
      \label{fig:img6}
    \end{subfigure} &
    \begin{subfigure}[b]{0.22\textwidth}
      \includegraphics[width=\linewidth]{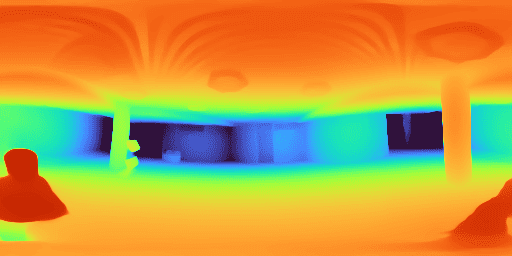}
      \label{fig:img7}
    \end{subfigure} &
    \begin{subfigure}[b]{0.22\textwidth}
      \includegraphics[width=\linewidth]{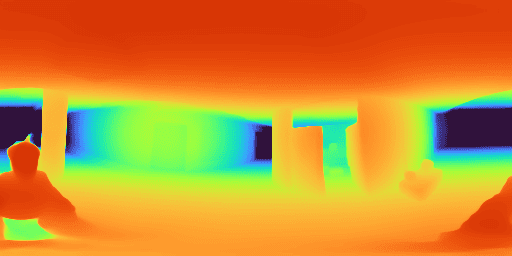}
      \label{fig:img8}
    \end{subfigure} \\

  \end{tabular}
  
  \caption{\textbf{Visualization of Depth Estimation on the Self-Collected Dataset}. The first row displays panoramic color images, and the second row shows the corresponding depth estimation visualizations.}
       \label{fig:cdepth}
\end{figure*}

\begin{figure*}[!h]
  \centering
  \captionsetup{aboveskip=-8pt}
  \begin{tabular}{@{}cccc@{}}
    \begin{subfigure}[b]{0.22\textwidth}
      \includegraphics[width=\linewidth]{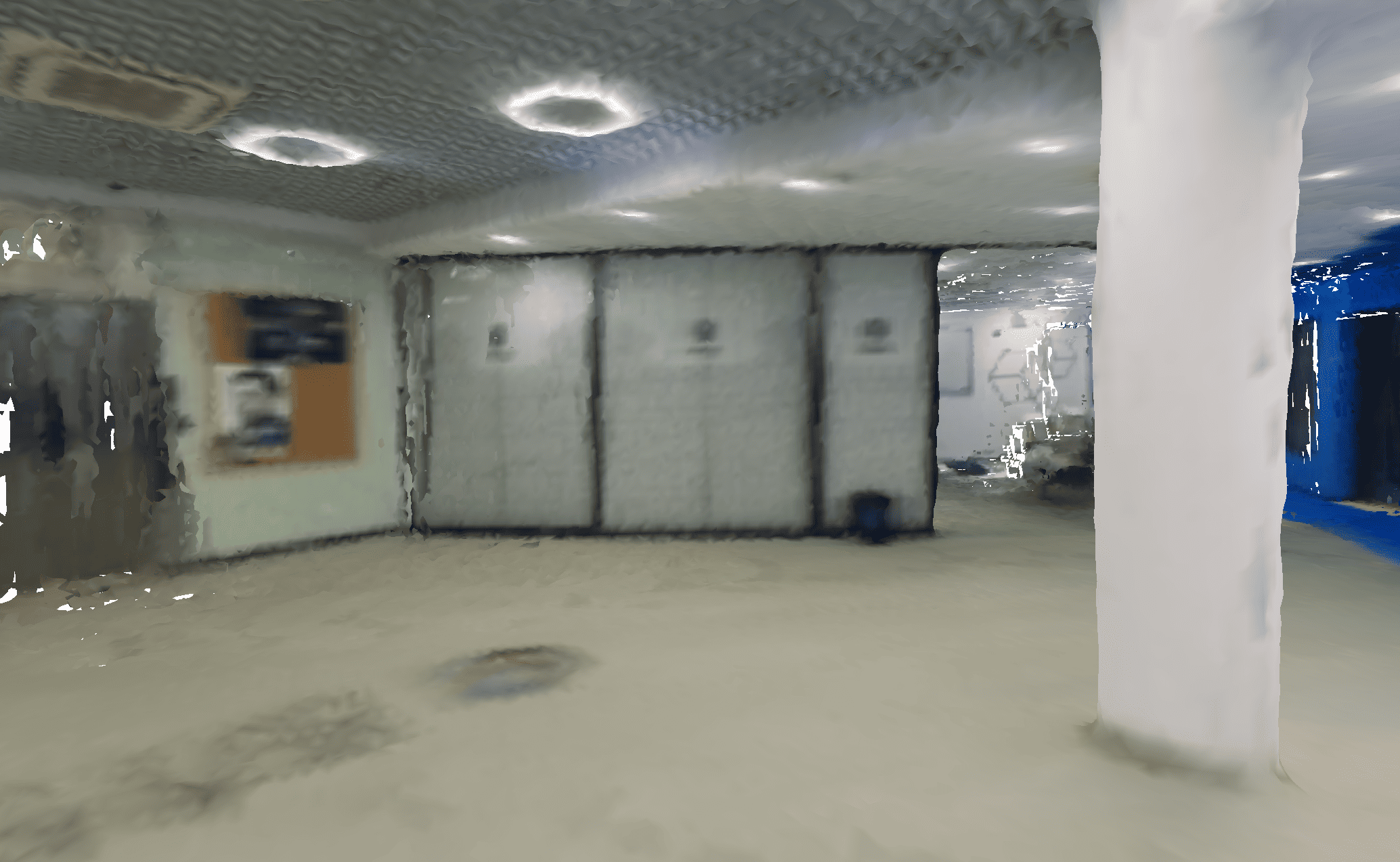}
      \label{fig:img1}
    \end{subfigure} &
    \begin{subfigure}[b]{0.22\textwidth}
      \includegraphics[width=\linewidth]{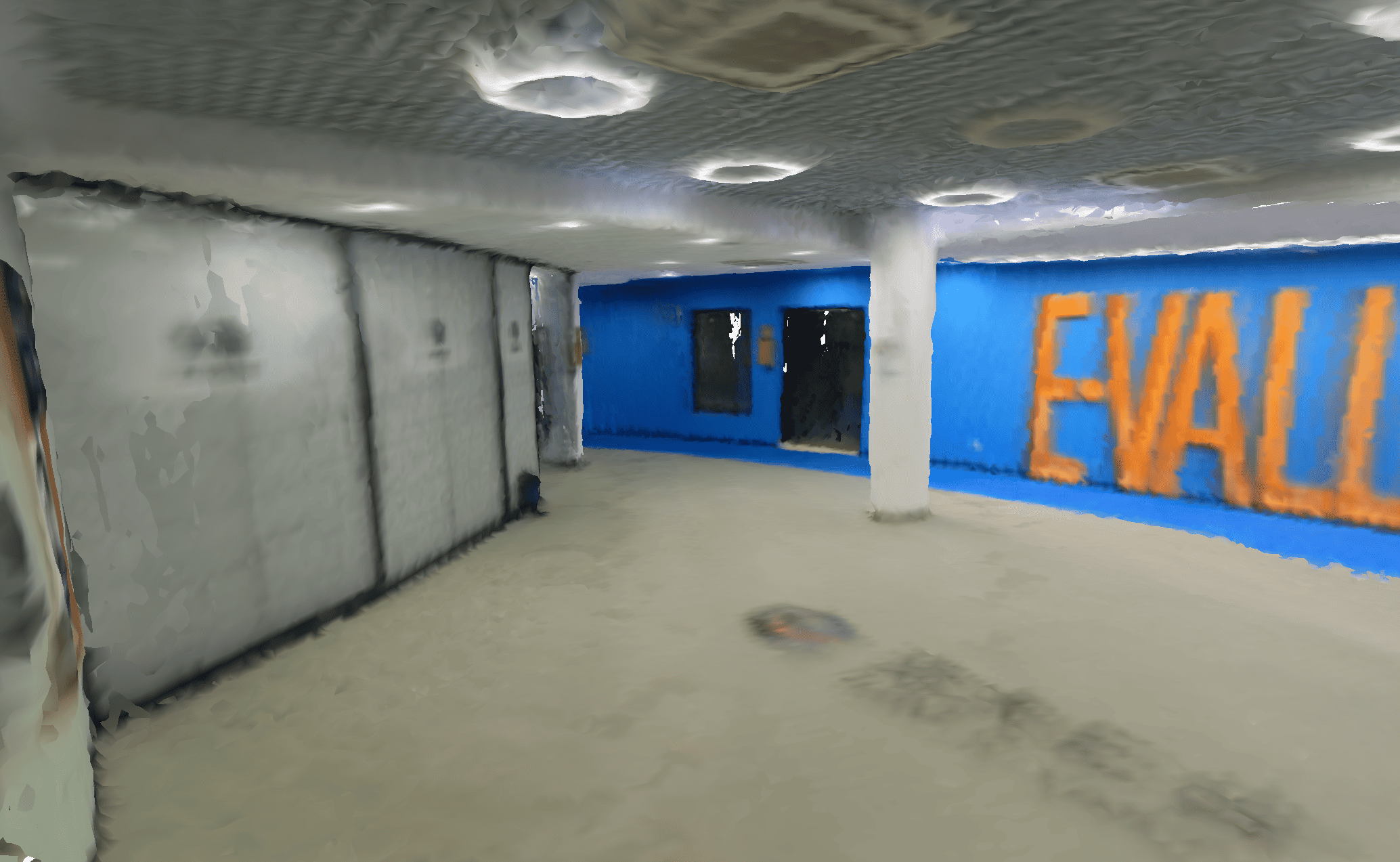}
      \label{fig:img2}
    \end{subfigure} &
    \begin{subfigure}[b]{0.22\textwidth}
      \includegraphics[width=\linewidth]{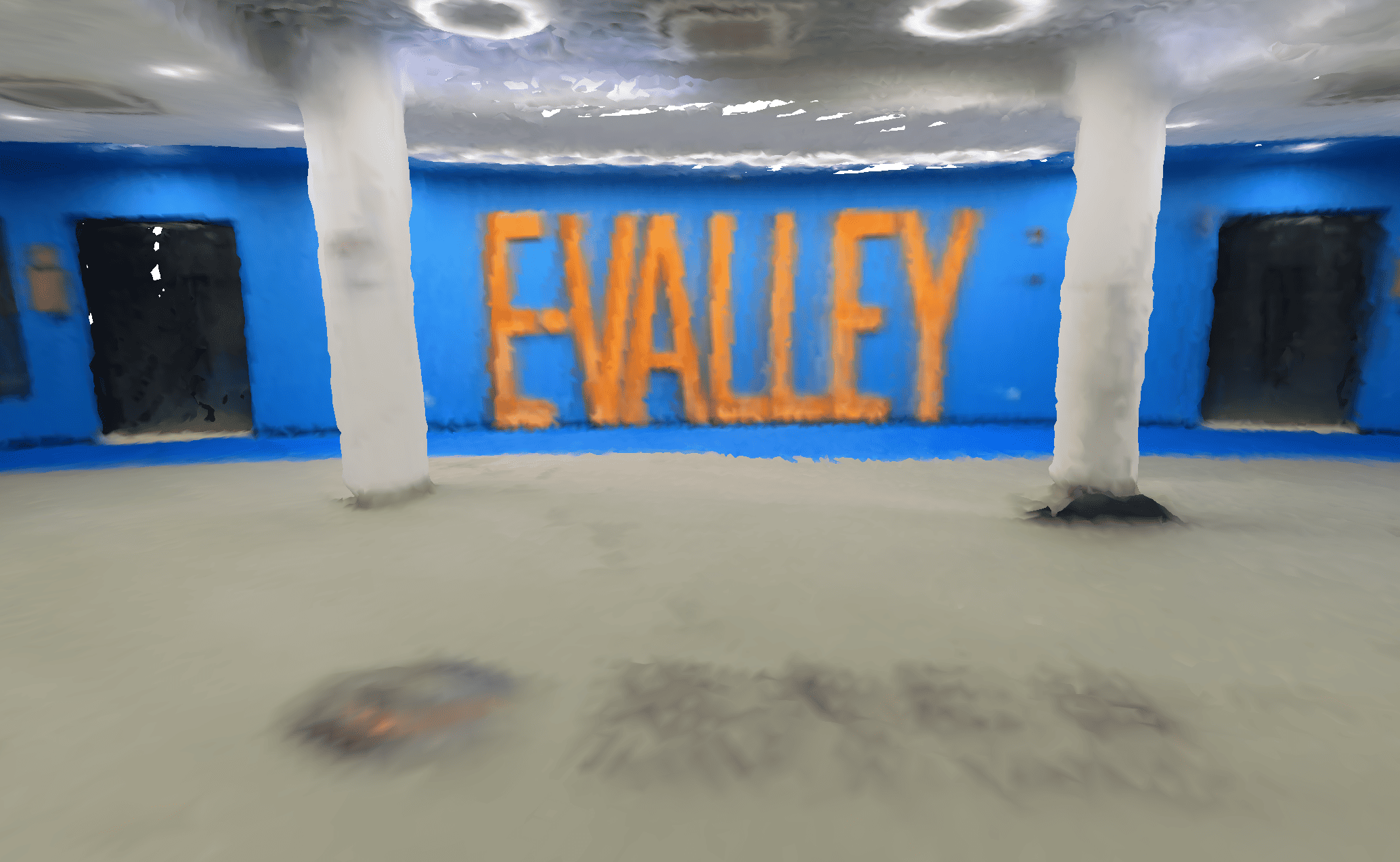}
      \label{fig:img3}
    \end{subfigure} &
    \begin{subfigure}[b]{0.22\textwidth}
      \includegraphics[width=\linewidth]{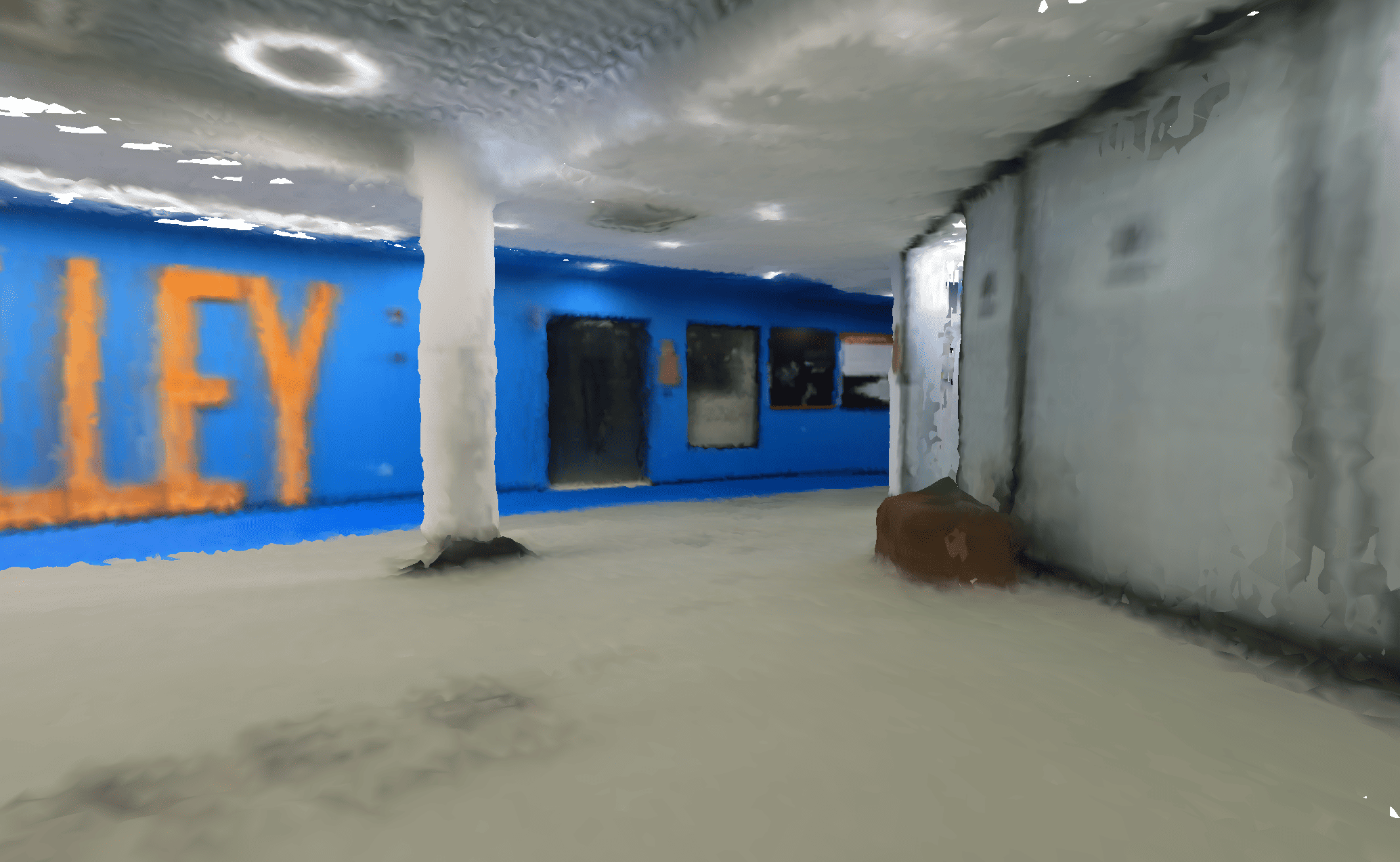}
      \label{fig:img4}
    \end{subfigure} \\
    
  \end{tabular}
  
  \caption{\textbf{Visualization of 3D Reconstruction Results on the Self-Collected Dataset.} We selected four images observed from different views of the reconstructed model using our self-collected dataset for display. }
  \label{fig:crecon}
\end{figure*}
To evaluate the efficiency of our model, we tested the size of parameters of our model and various other models, measured the inference time for depth prediction on each frame, and assessed the GPU memory consumption during runtime. The results are presented in \cref{table:efficiency}.

Single-frame image depth estimation models, such as Bifuse++~\cite{Bifusev2} and PanoFormer~\cite{panoformer}, achieve faster inference speeds and lower GPU memory consumption because they do not require computations involving multi-frame constraints, particularly spatial projection and back-projection operations.

For multi-view depth estimation models, FoVA-Depth~\cite{fova} and 360MVSNet~\cite{panogrf} require substantial GPU memory during runtime due to their use of 3D CNN architectures. In contrast, our multi-view model has an advantage in terms of memory consumption.

In terms of parameter size and runtime, our algorithm outperforms 360MVSNet~\cite{panogrf} and achieves a parameter size comparable to that of Bifuse++~\cite{Bifusev2}. This is attributed to our use of a simple MLP network to simplify the 4D cost volume, and employing an Encoder-Decoder network structure based on 2D CNNs to generate depth predictions, instead of using computationally expensive 3D CNNs.

These experiments demonstrate that our algorithm not only ensures accurate depth estimation and 3D reconstruction but also achieves relatively good computational efficiency and comparatively acceptable GPU memory consumption.

\section{Ablation Studies}
\label{ablation}

\begin{table}[tp]	
\centering
\fontsize{5.3}{4}\selectfont
\renewcommand{\arraystretch}{1.5} 
\resizebox{0.9\linewidth}{!}{
\begin{tabular}{ccccccc}
	\toprule
	Method&w/o SphereCNN,Image-Ecoder&w/o Image-Ecoder&w/o SphereCNN&360Recon \cr
	\midrule
	
                             MAE$\downarrow$(cm)    &19.25&16.22&14.41 & \textbf{13.68}  \\
                             MRE$\downarrow$(cm)  &8.31&7.34&5.87  & \textbf{5.50}  \\
                             RMSE$\downarrow$(cm)   &45.30&41.27&37.71  & \textbf{36.22}  \\
                             $\delta_1\uparrow$(\%)  &92.71&94.06& 95.26 & \textbf{95.35} \\
                             chamfer$\downarrow$(cm)    &10.75&11.02&9.90& \textbf{9.68}  \\
                             F-score$\uparrow$    &37.6&38.1& \textbf{43.0} & 42.1  \\
        \midrule
	
\end{tabular}
}
\vspace{-0.3cm}
\caption{Quantitative  comparison of the metrics for depth estimation and 3D reconstruction among different ablation experiment models on the Matterport3D~\cite{Matterport3D} test set. 'w/o' indicates 'without' a specific module.}
\vspace{-0.6cm}
\label{table:ablation_supp}
\end{table}
In this section, we demonstrate the contributions of both the Sphere Feature Extractor and the image feature enhancement network to the overall algorithm through additional ablation experiments.

As mentioned in \cref{method}, we utilize a designed Sphere CNN layer to ensure that the features involved in matching are not affected by the pixel latitude positions. We also employ a pre-trained image encoder to enhance image features, which are then input into the Encoder-Decoder structure for depth prediction. The information in the cost volume can be regarded as geometric information obtained based on multi-view constraints, while the enhanced image features provided by the image encoder offer more detailed texture information for depth prediction.

We separately removed the SphereCNN structure and the image encoder structure to test the impact of these two modules on the overall algorithm. We conducted ablation experiments on the test set of Matterport3D~\cite{Matterport3D}. The metrics for depth estimation and 3D reconstruction of different models are presented in \cref{table:ablation_supp}. It is evident that both the Image Encoder and the SphereCNN structures have positively impacted the model. Compared to the network model without both the SphereCNN and image encoder, our complete model achieved a 28.9\% reduction in RMSE and a 12.0\% increase in F-Score.

\section{Qualitative Results}
\label{Qualitative}
In this section, we will present more qualitative results of depth estimation and 3D reconstruction, along with comparisons to other algorithms~\cite{Bifusev2,panoformer,fova,panogrf}.

\begin{figure*}[t]
    \centering
    \newcommand{\mywidth}{0.14\textwidth }
    \setlength\tabcolsep{0.05em}
    \newcolumntype{P}[1]{>{\centering\arraybackslash}m{#1}}
    \def\arraystretch{0.30}
    \begin{tabular}{P{\mywidth} P{\mywidth} P{\mywidth} P{\mywidth} P{\mywidth} P{\mywidth} P{\mywidth}}
        \renewcommand{\arraystretch}{0.05}
        
        \includegraphics[width=\mywidth]{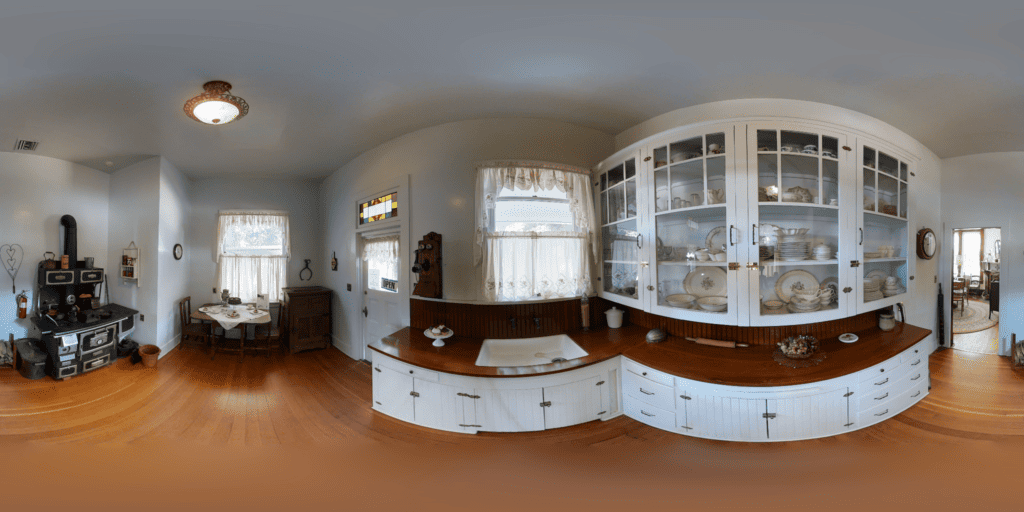}\hspace{-1cm} &
        \includegraphics[width=\mywidth]{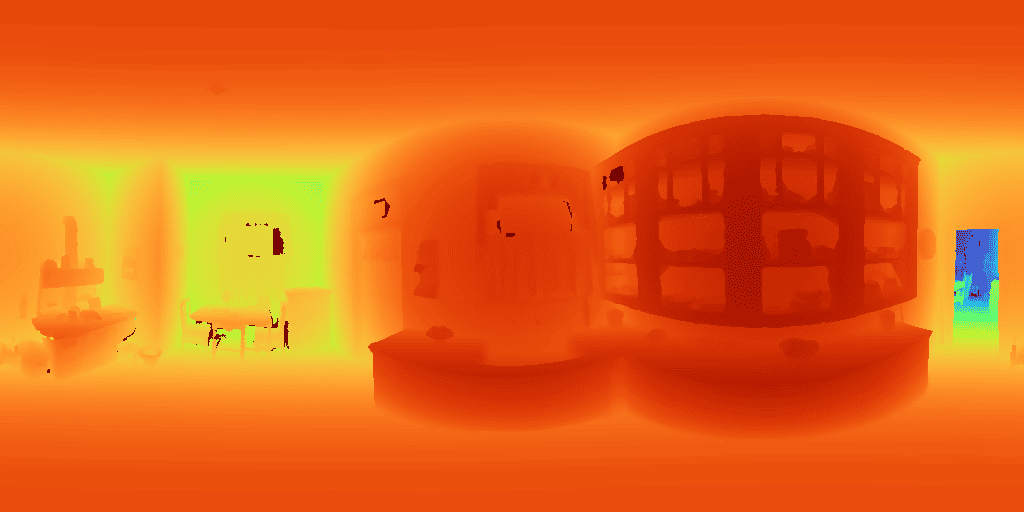}\hspace{-1cm} &
        \includegraphics[width=\mywidth]{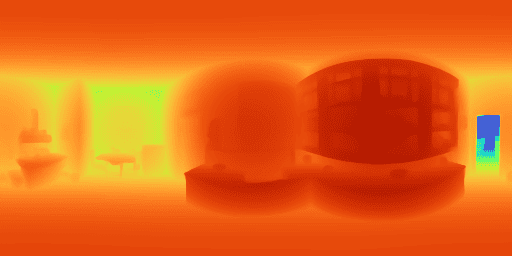} &
        \includegraphics[width=\mywidth]{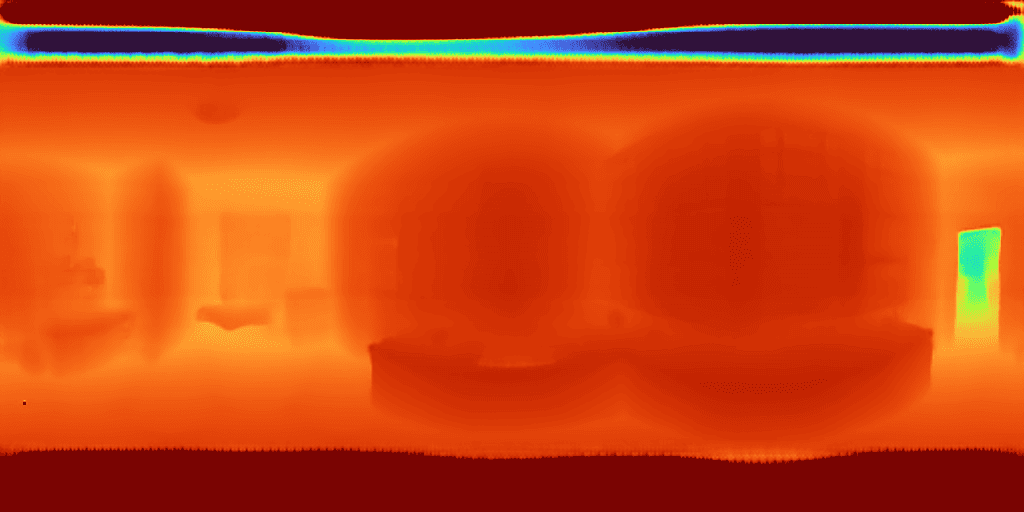}\hspace{-1cm} &
        \includegraphics[width=\mywidth]{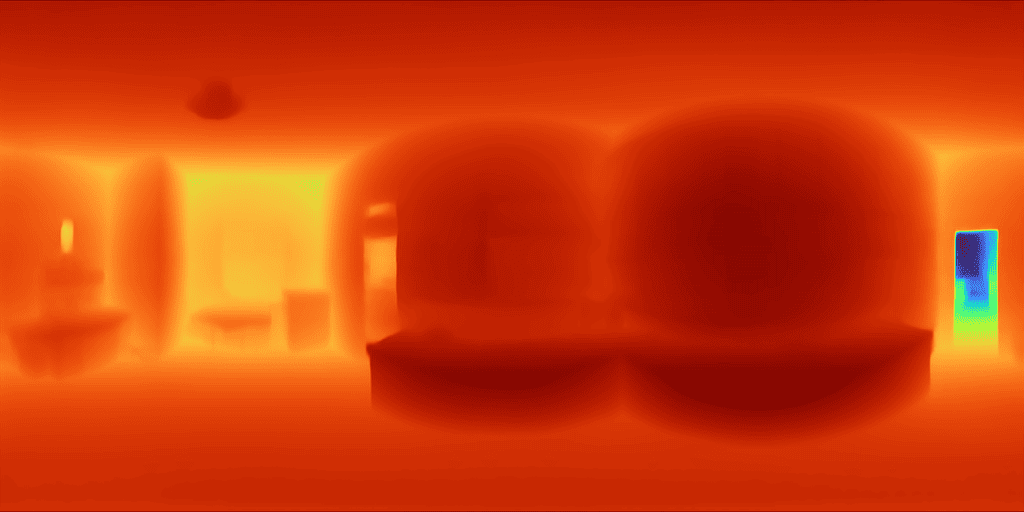}\hspace{-1cm} &
        \includegraphics[width=\mywidth]{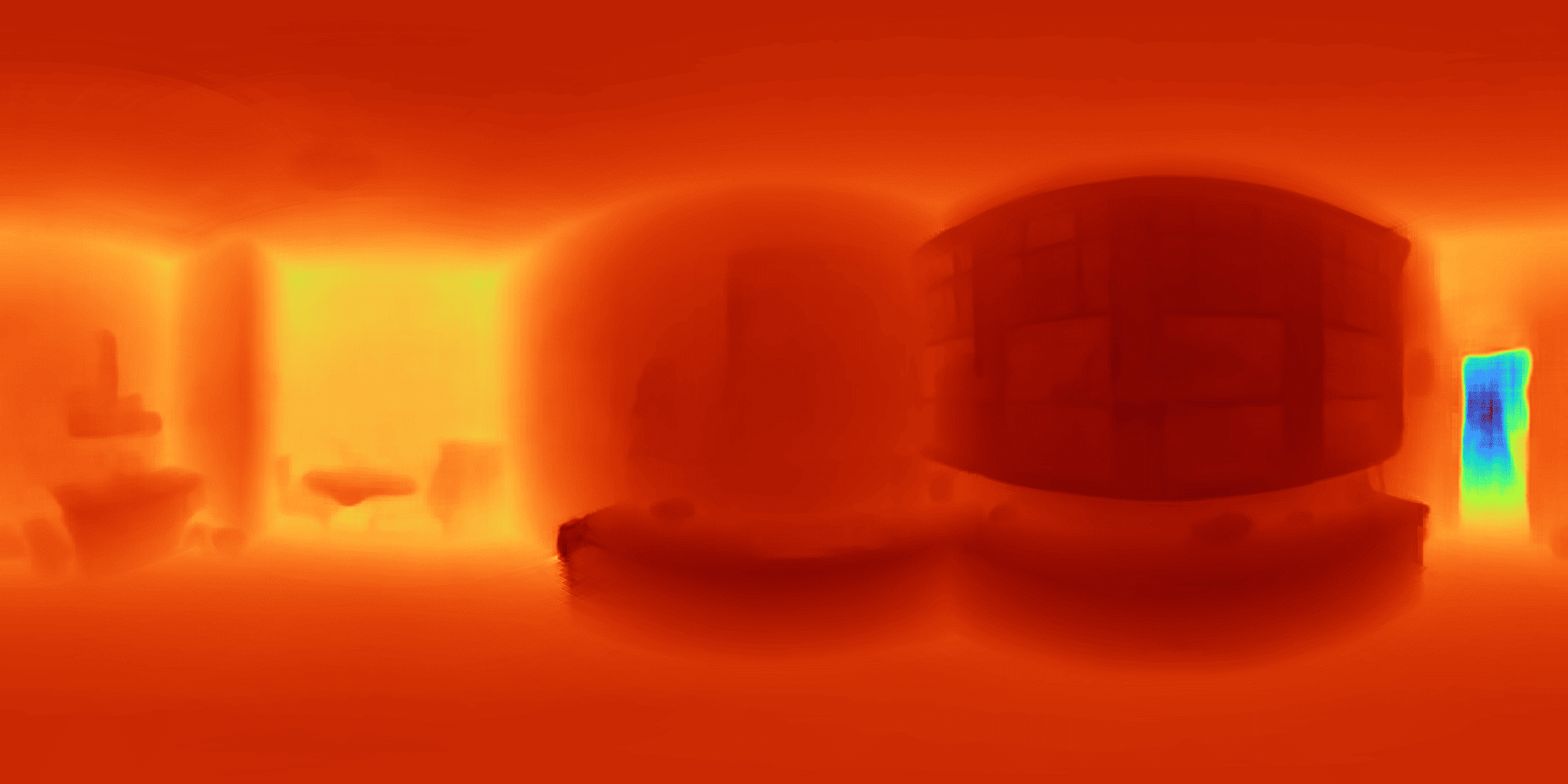}\hspace{-1cm} &
        \includegraphics[width=\mywidth]{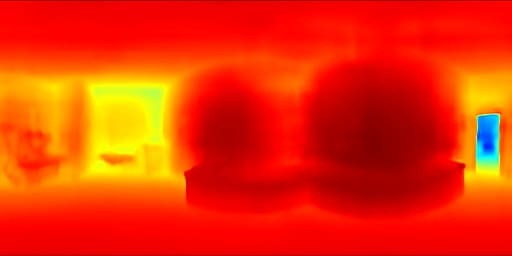}\hspace{-0.8cm} 
         \\

        \includegraphics[width=\mywidth]{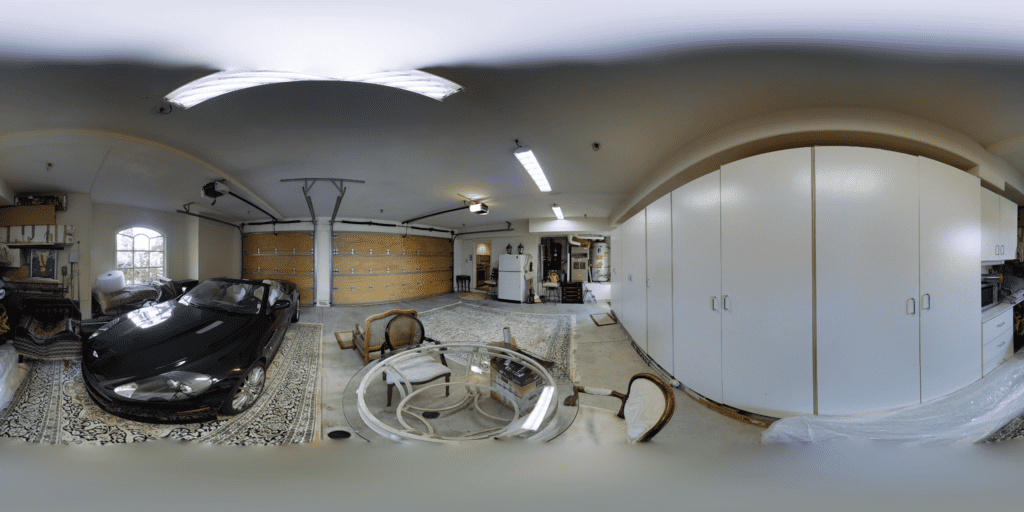} &
        \includegraphics[width=\mywidth]{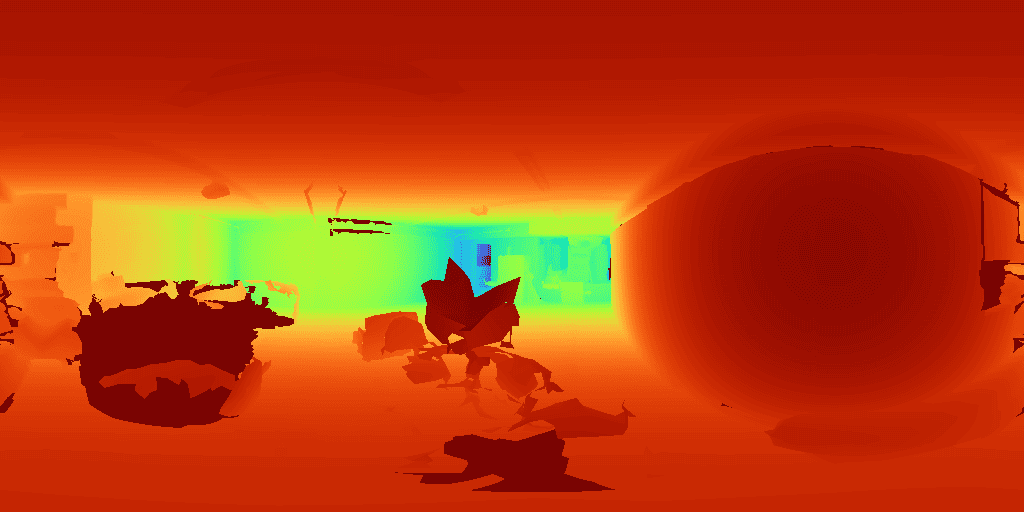} &
        \includegraphics[width=\mywidth]{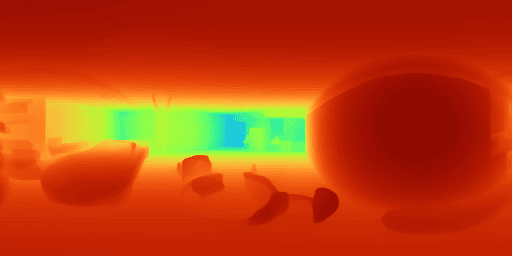} &
        \includegraphics[width=\mywidth]{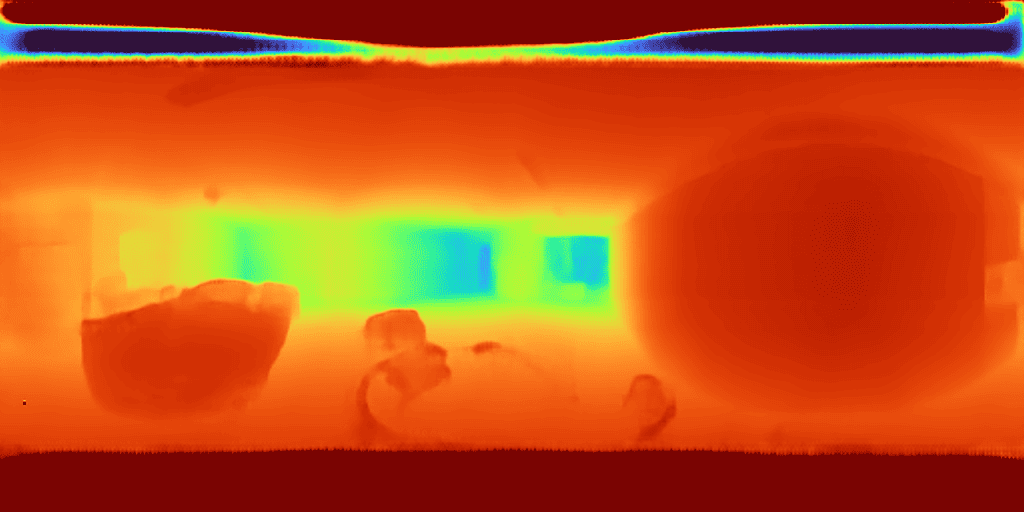} &
        \includegraphics[width=\mywidth]{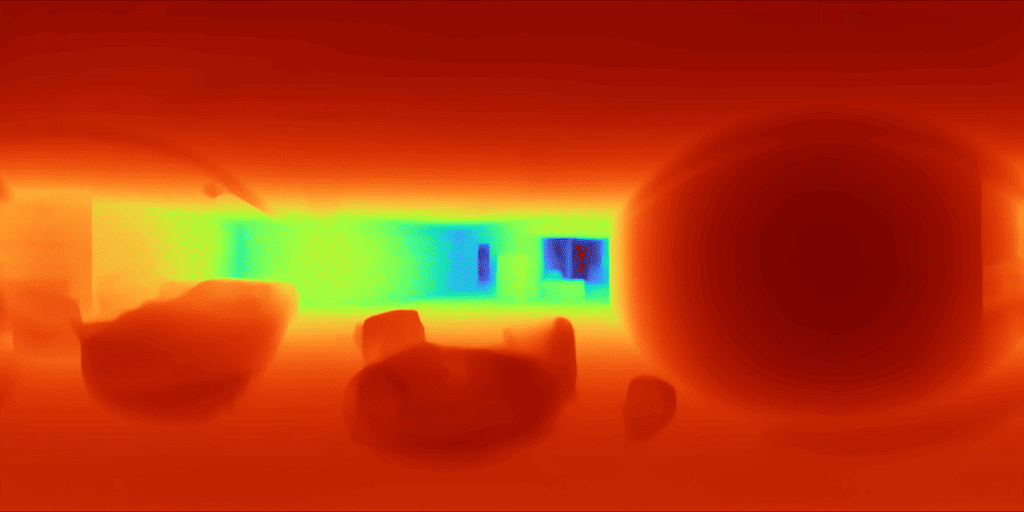} &
        \includegraphics[width=\mywidth]{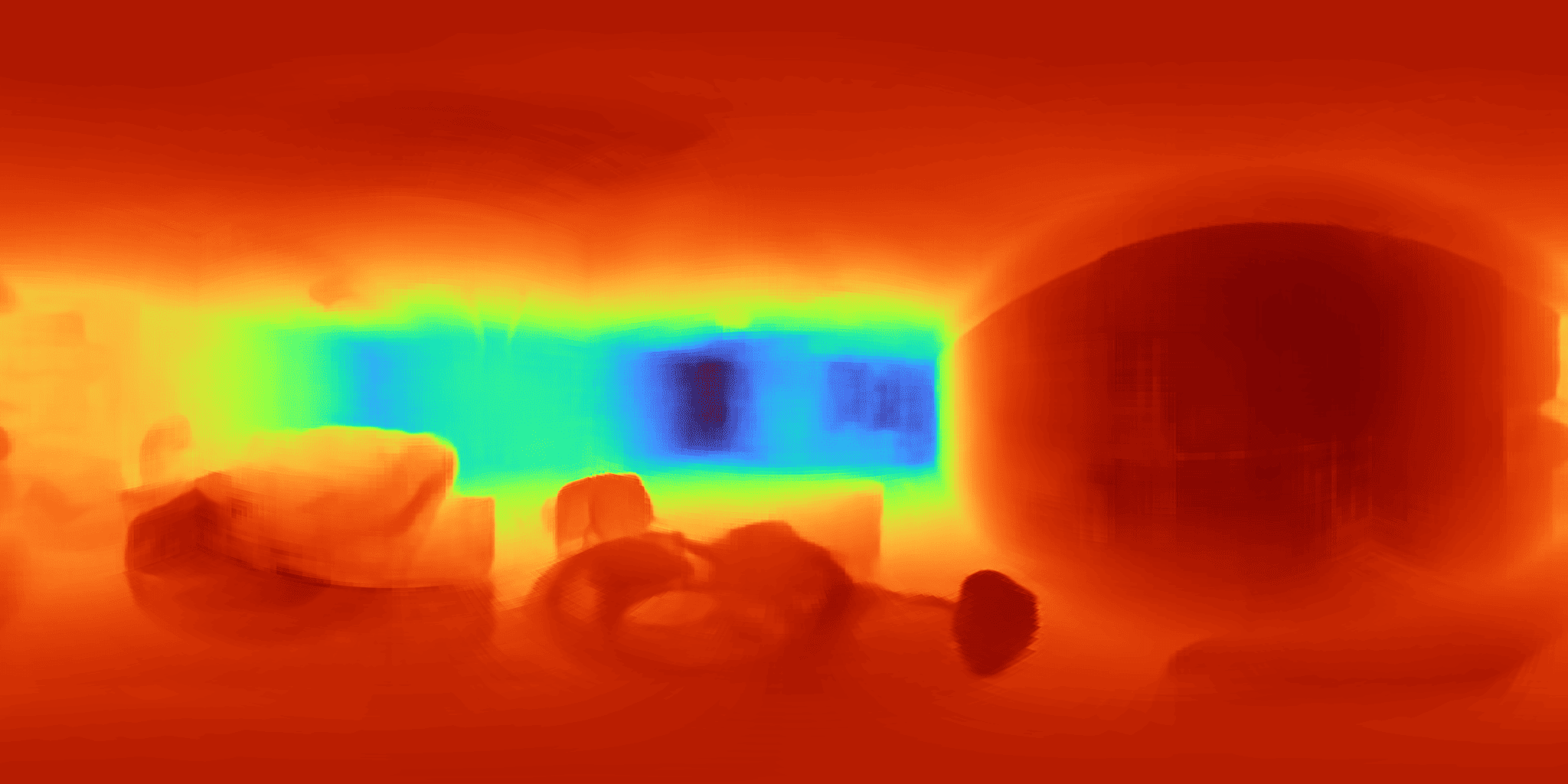} &
        \includegraphics[width=\mywidth]{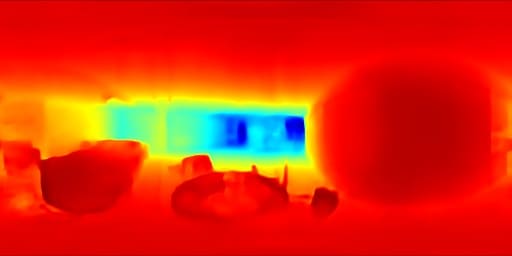} 
 \\

        \includegraphics[width=\mywidth]{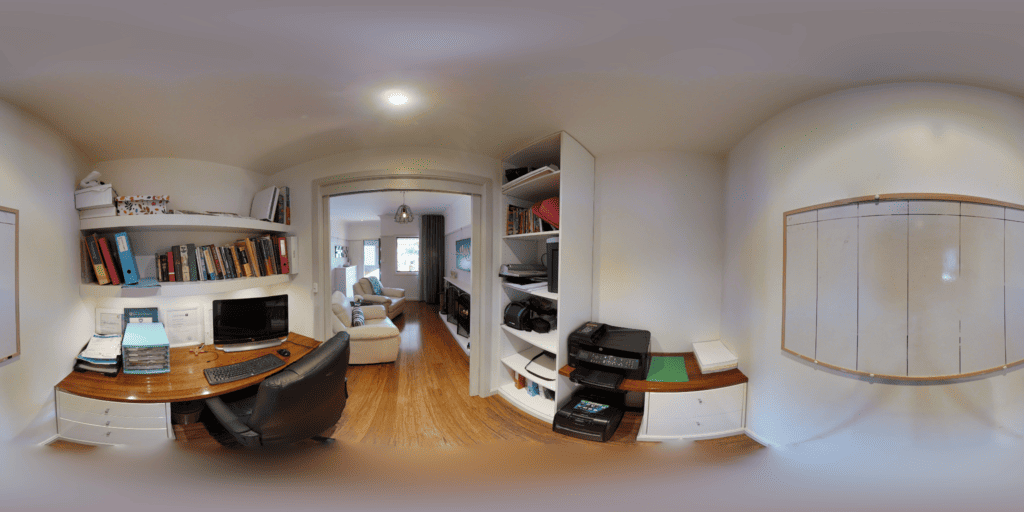} &
        \includegraphics[width=\mywidth]{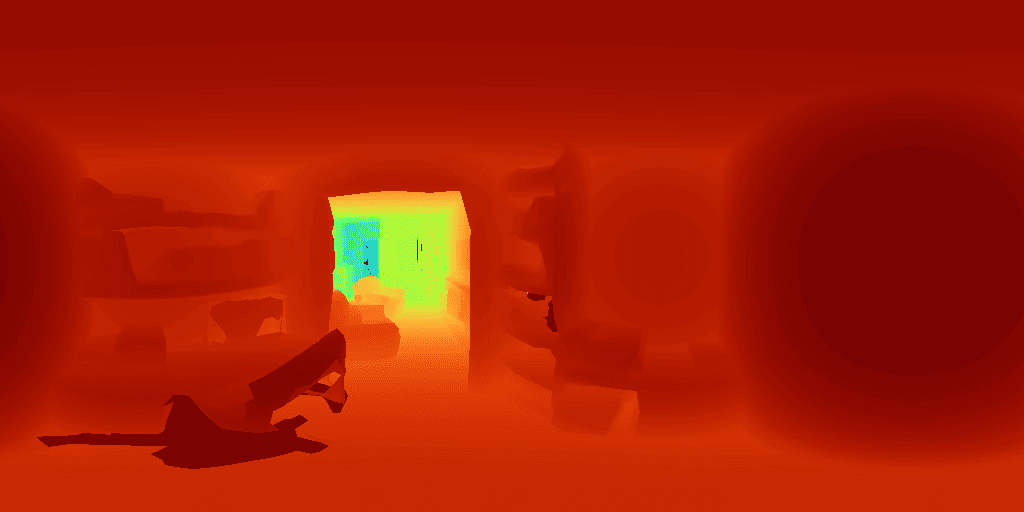} &
                \includegraphics[width=\mywidth]{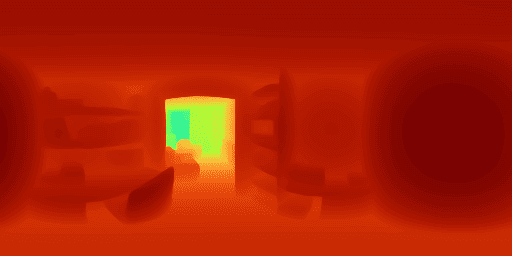}  &
        \includegraphics[width=\mywidth]{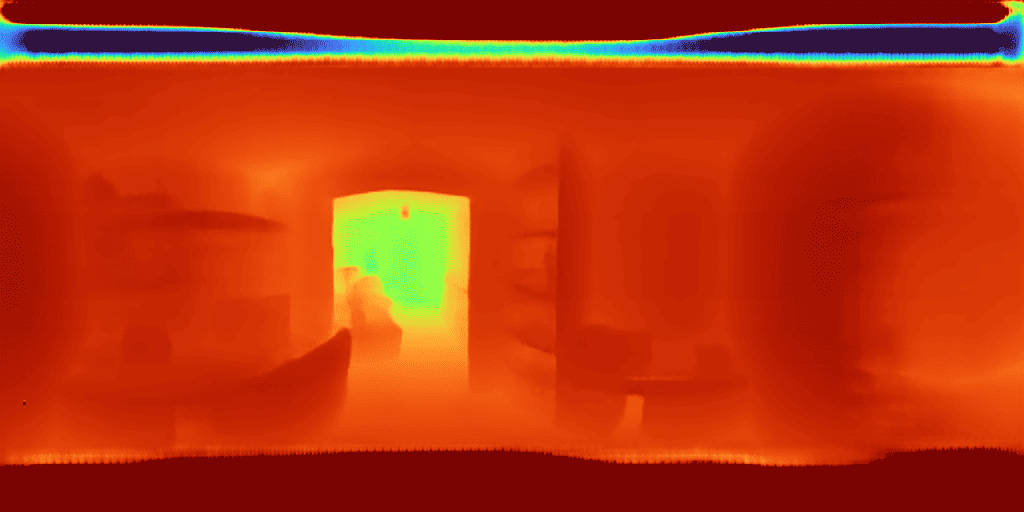} &
        \includegraphics[width=\mywidth]{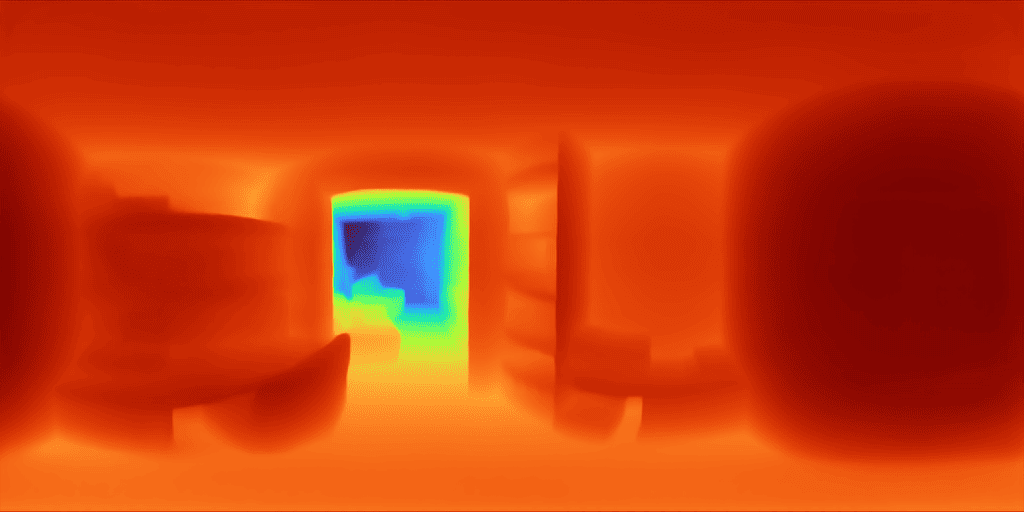} &
        \includegraphics[width=\mywidth]{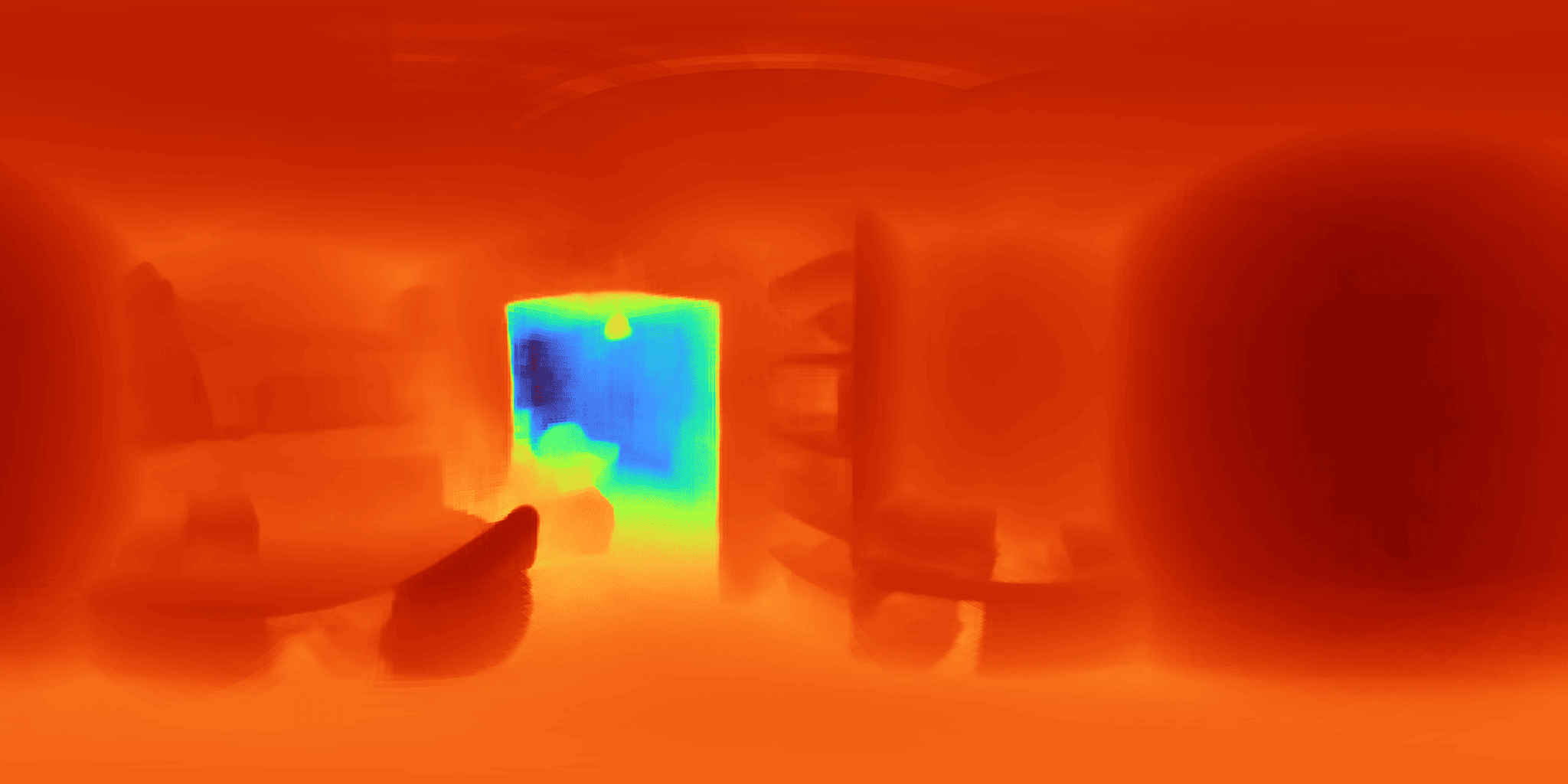} &
        \includegraphics[width=\mywidth]{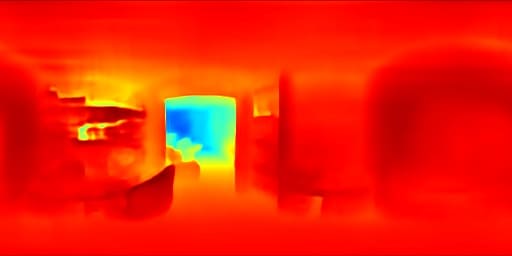} \\

        \includegraphics[width=\mywidth]{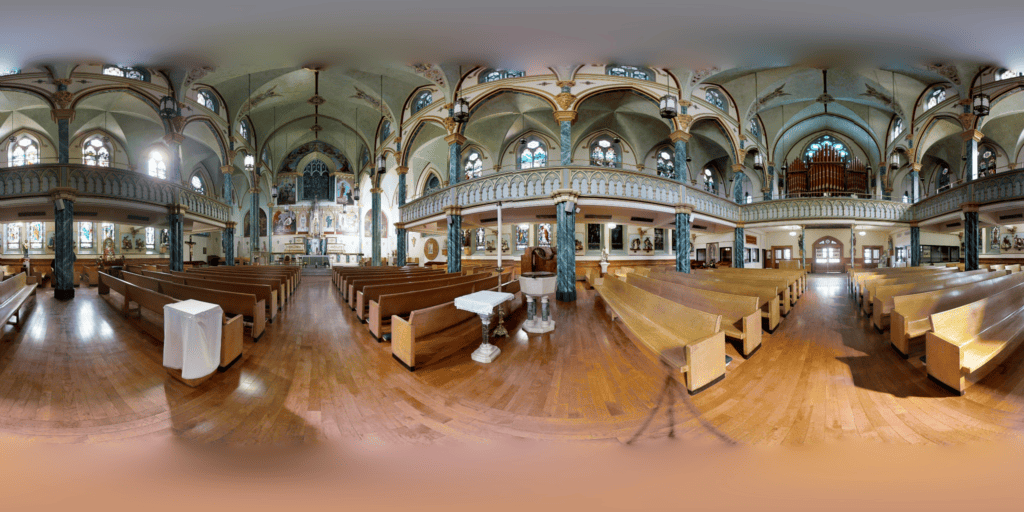} &
        \includegraphics[width=\mywidth]{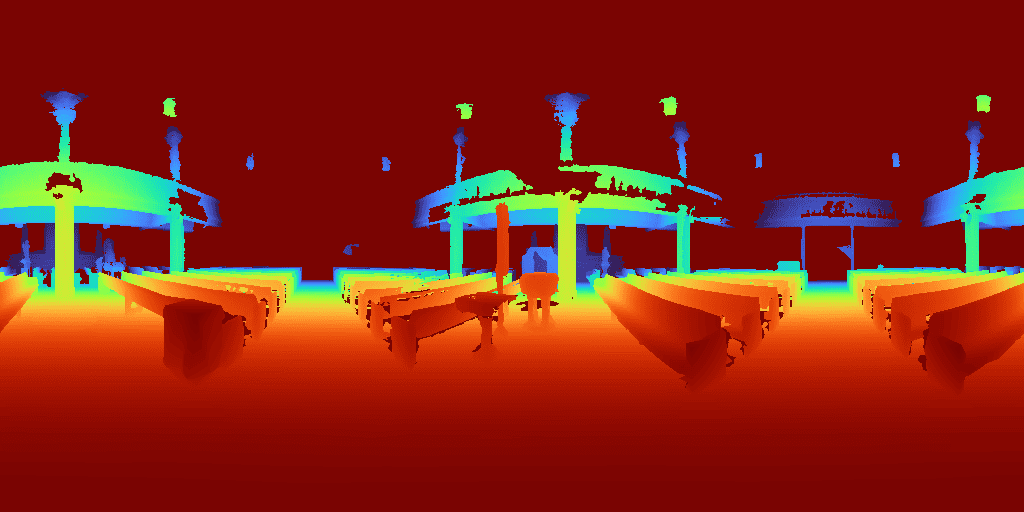} &
                \includegraphics[width=\mywidth]{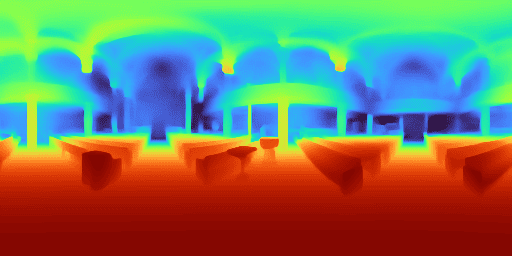} &
        \includegraphics[width=\mywidth]{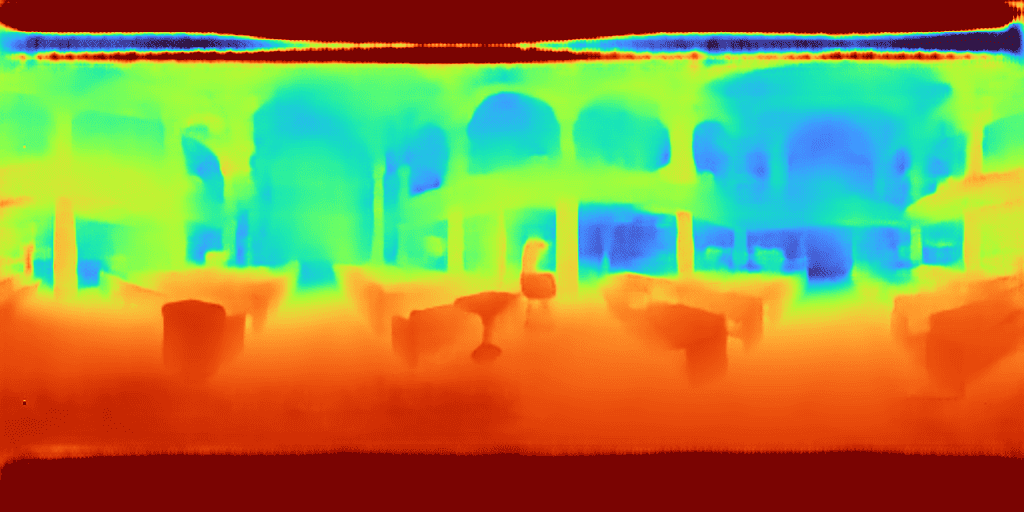} &
        \includegraphics[width=\mywidth]{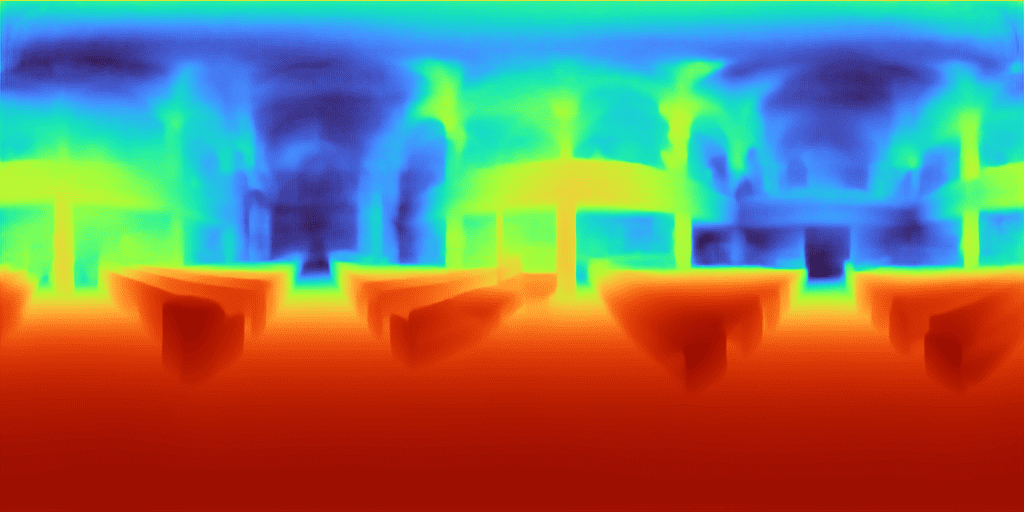} &
        \includegraphics[width=\mywidth]{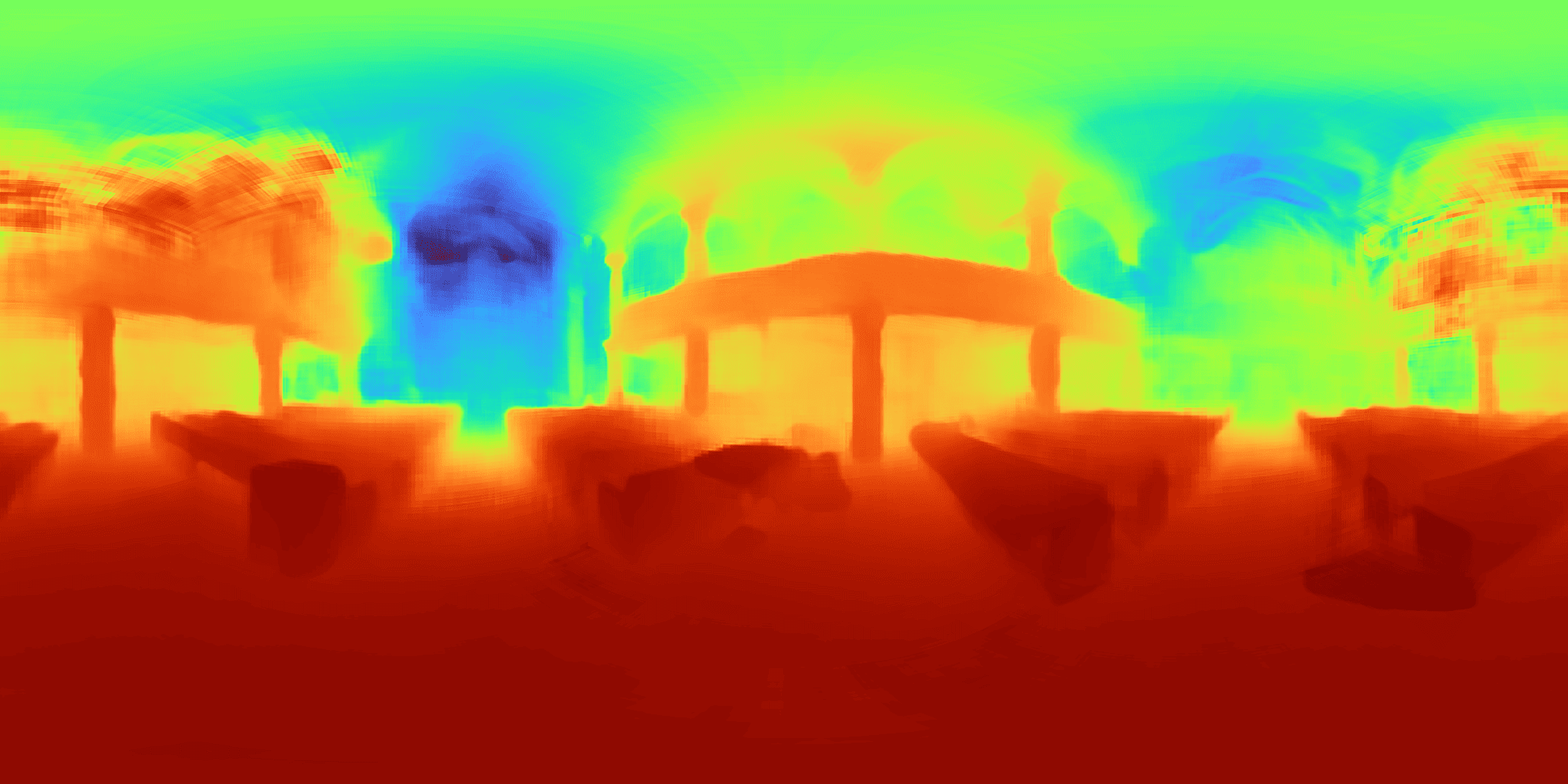} &
        \includegraphics[width=\mywidth]{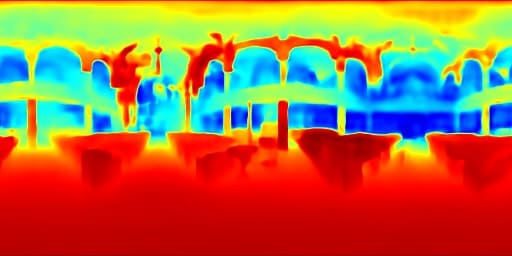} 
 \\

        \includegraphics[width=\mywidth]{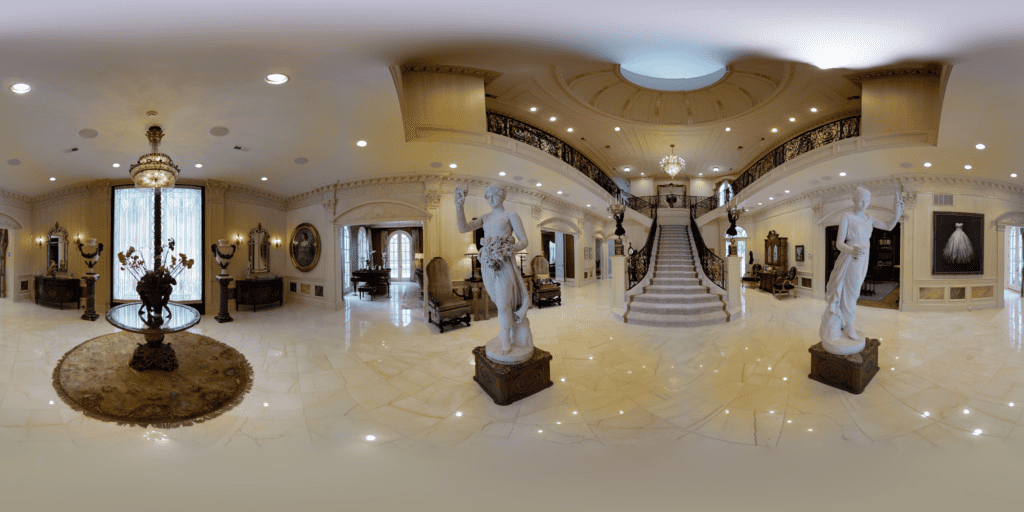} &
        \includegraphics[width=\mywidth]{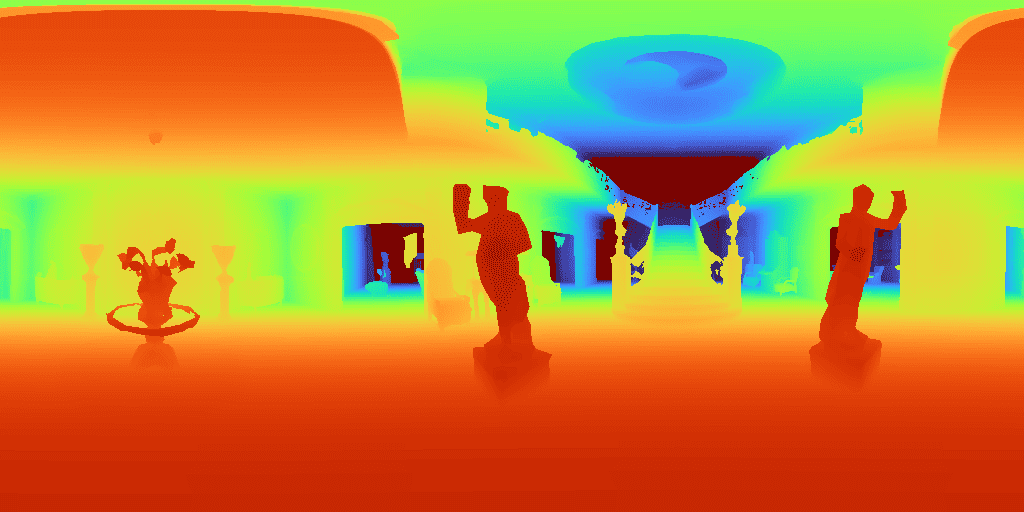} &
                \includegraphics[width=\mywidth]{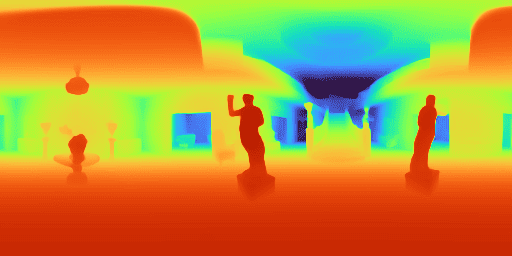} &
        \includegraphics[width=\mywidth]{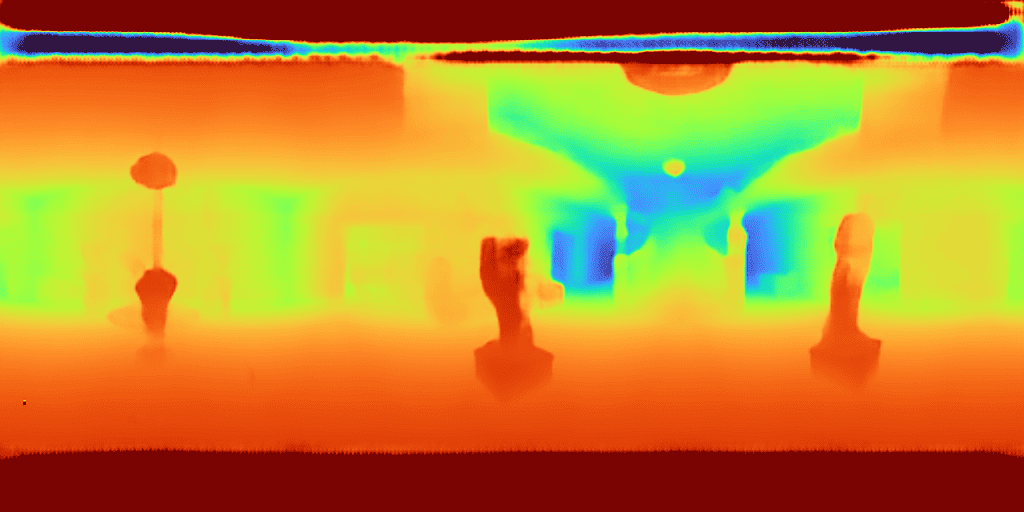} &
        \includegraphics[width=\mywidth]{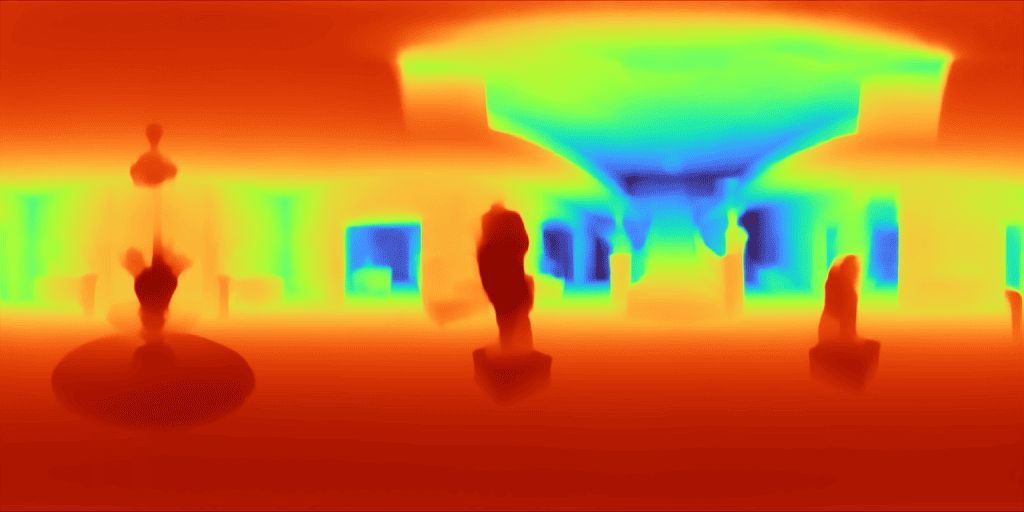} &
        \includegraphics[width=\mywidth]{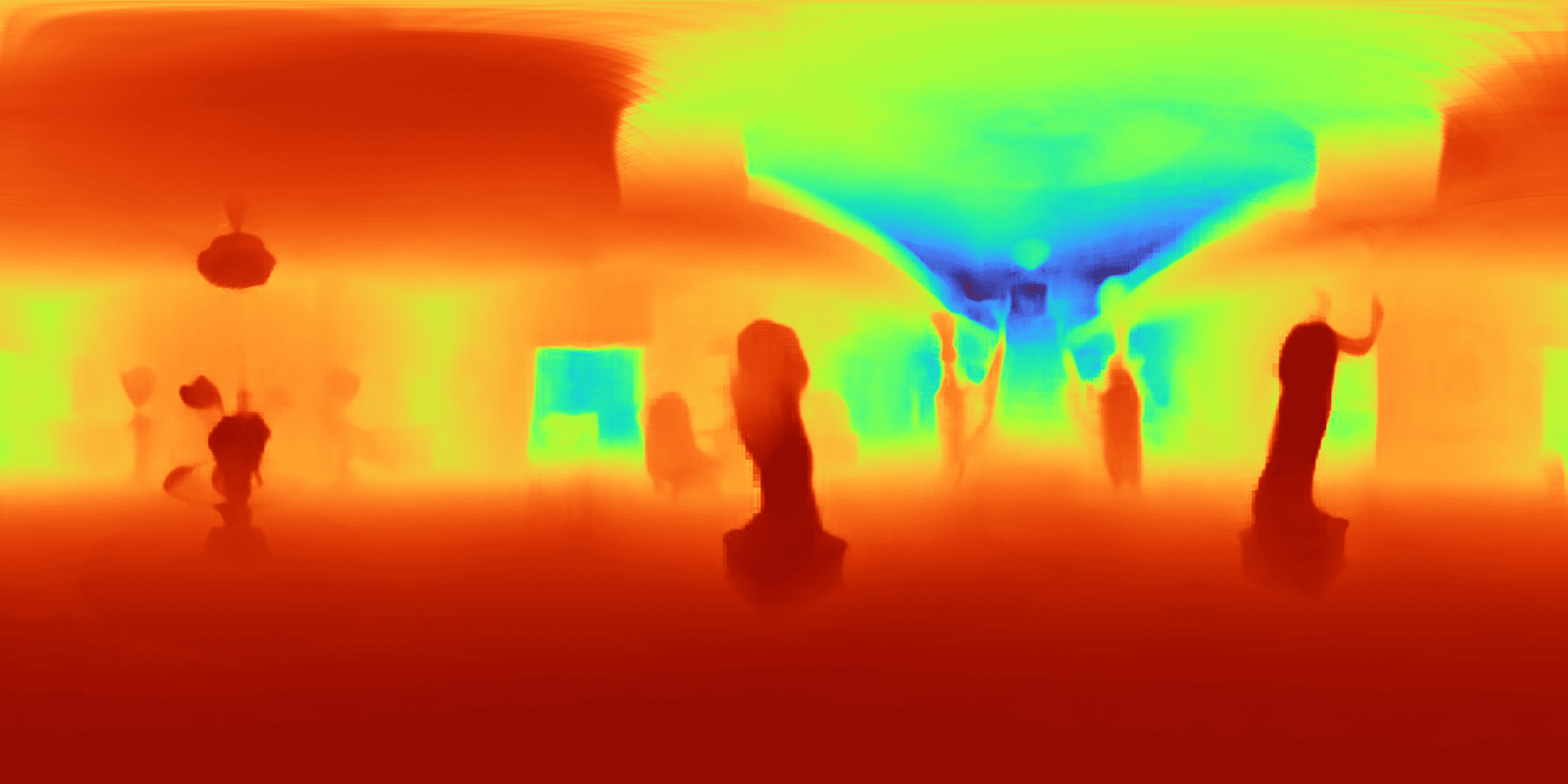} &
        \includegraphics[width=\mywidth]{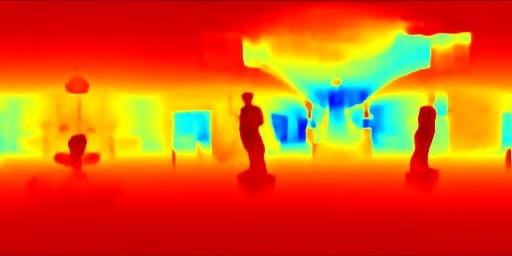} \\

        \includegraphics[width=\mywidth]{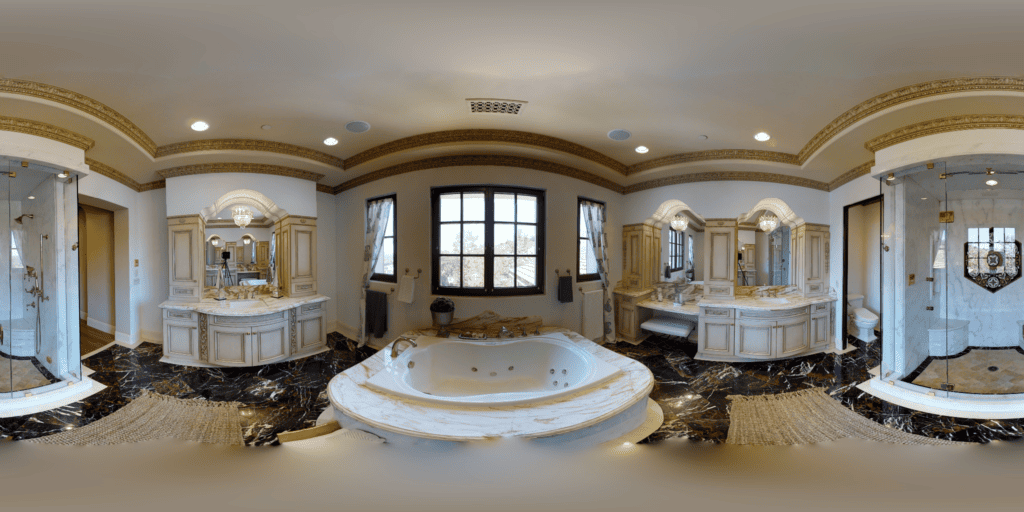} &
        \includegraphics[width=\mywidth]{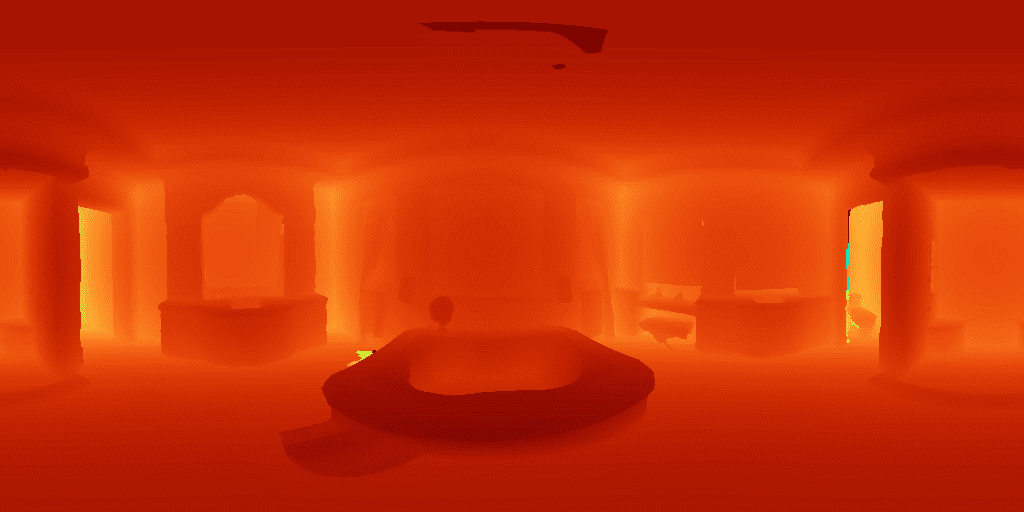} &
                \includegraphics[width=\mywidth]{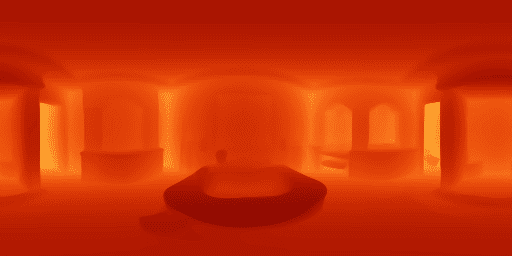} &
        \includegraphics[width=\mywidth]{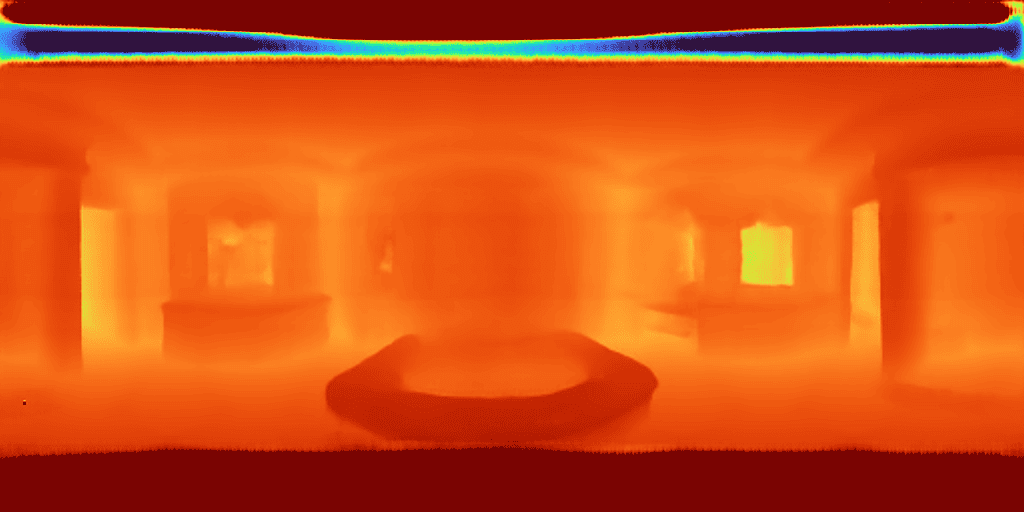} &
        \includegraphics[width=\mywidth]{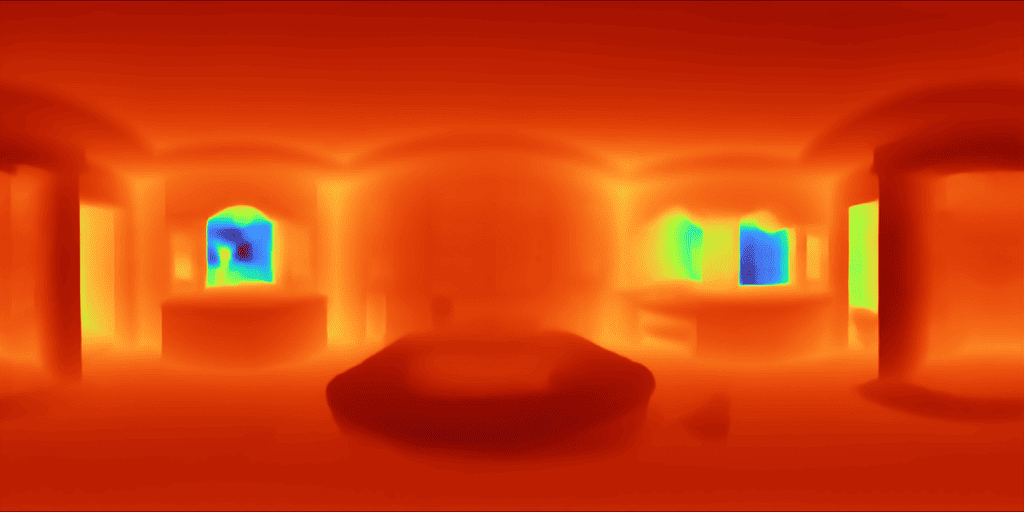} &
        \includegraphics[width=\mywidth]{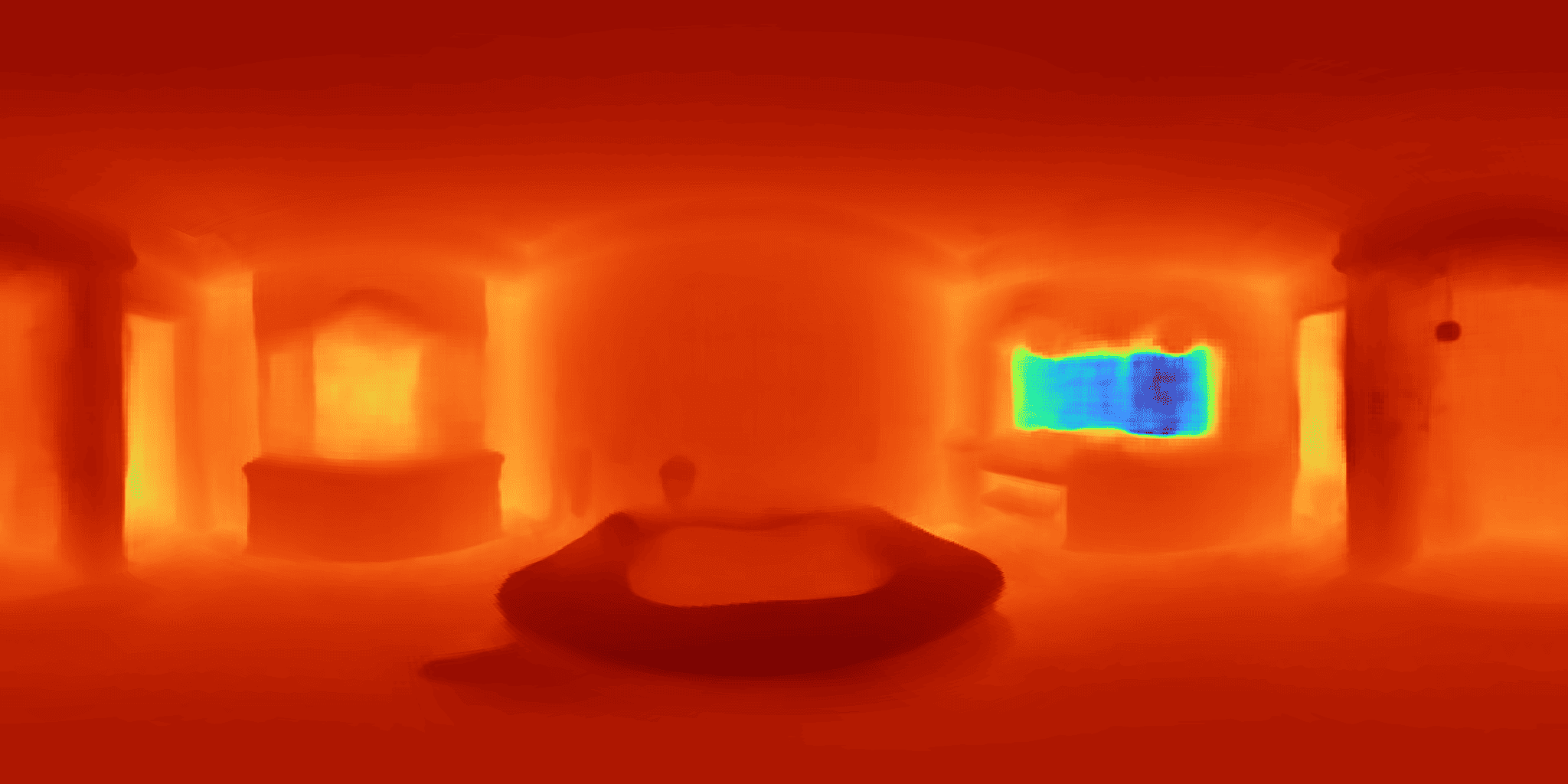} &
        \includegraphics[width=\mywidth]{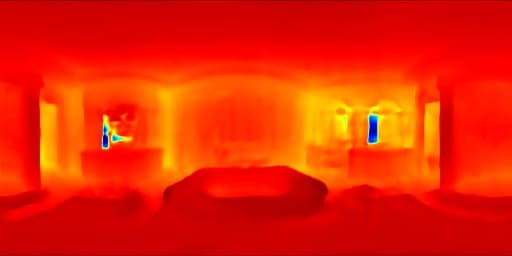} 
 \\
         RGB & GT & 360Recon & Bifuse++ & PanoFormer & FoVA-Depth & 360-MVSNet$^{*}$
    \end{tabular}
    \vspace{-8pt}
    \caption{\textbf{Depth Estimation Comparison.} The predicted depth maps are compared with various methods in different scenes. Our method provides more accurate and detailed depth estimates across these datasets.}
    \label{fig:depth_more}
    \vspace{-0.4cm}
\end{figure*}

In \cref{fig:depth_more}, we present a comparison of depth estimation across different scenes, including a restaurant, garage, study, church, entry hall, and bathroom. Through visual comparison, it is evident that our algorithm provides more detailed and accurate depth estimation in these scenes.

\begin{figure*}[t]
    \centering
    \newcommand{\mywidth}{0.16\textwidth }
    \setlength\tabcolsep{0.05em}
    \newcolumntype{P}[1]{>{\centering\arraybackslash}m{#1}}
    \def\arraystretch{0.30}
    \begin{tabular}{P{\mywidth} P{\mywidth} P{\mywidth} P{\mywidth} P{\mywidth} P{\mywidth}}
        \renewcommand{\arraystretch}{0.05}

        \includegraphics[width=\mywidth]{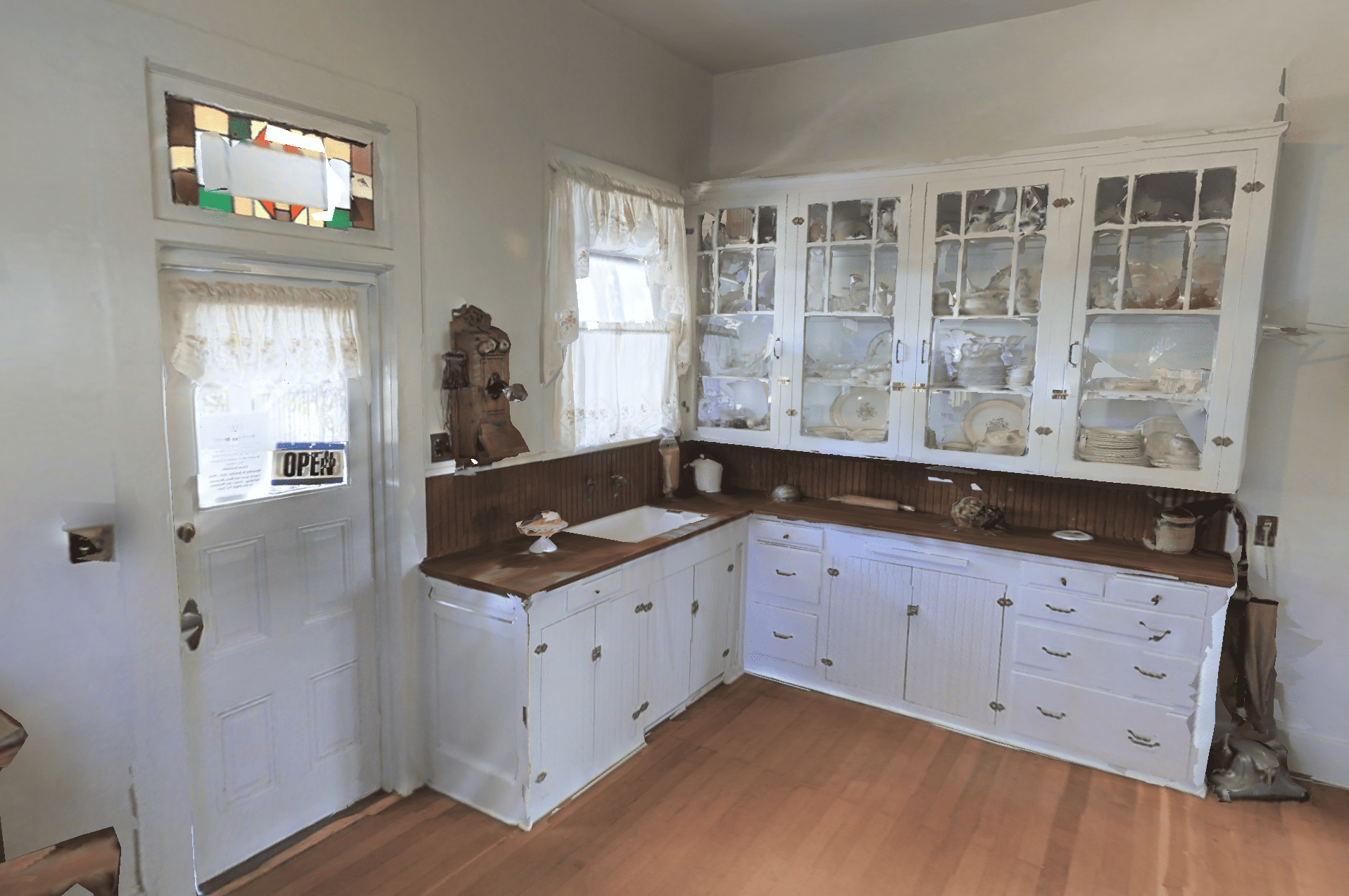}\hspace{-1cm} &
        \includegraphics[width=\mywidth]{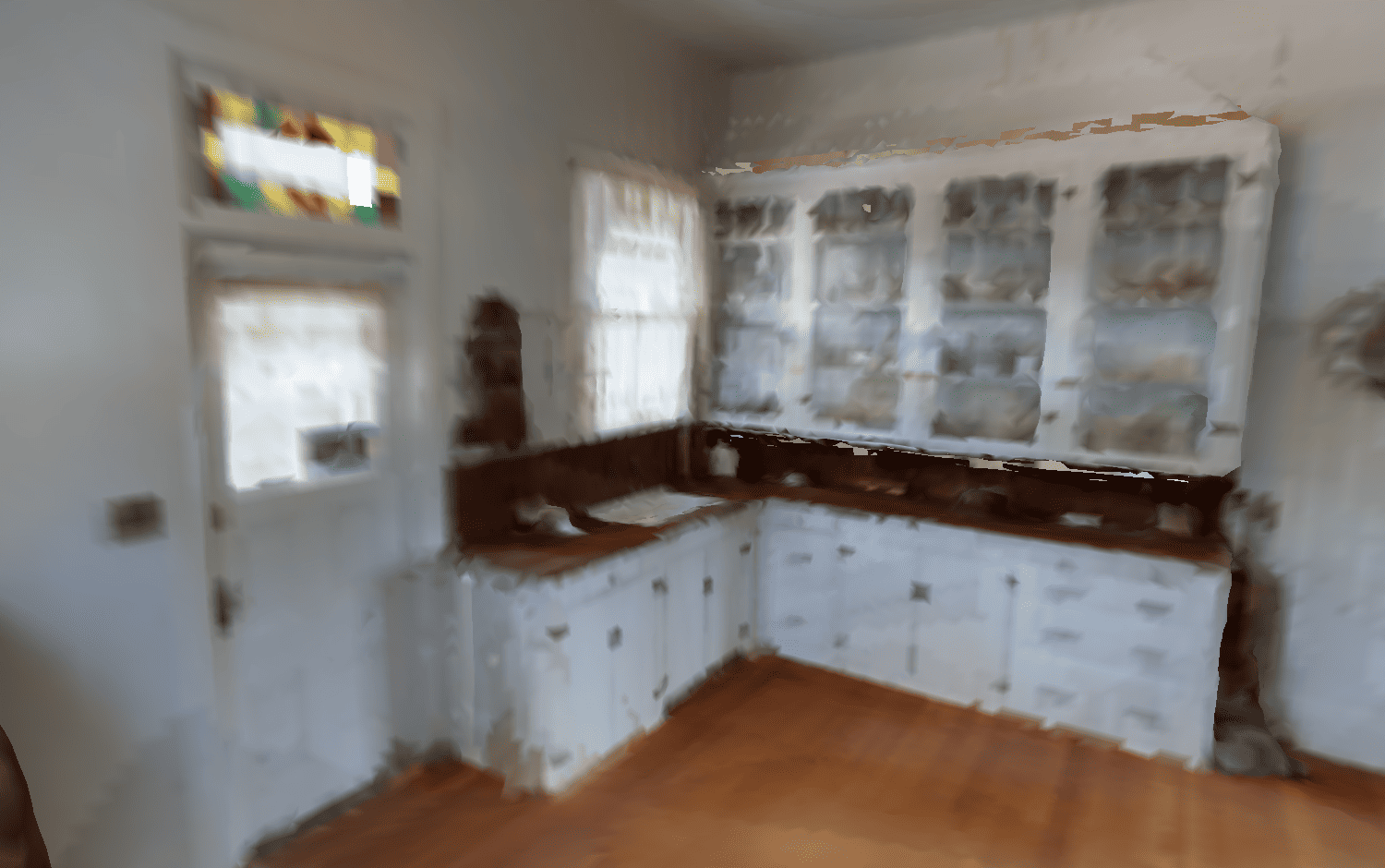} &
        \includegraphics[width=\mywidth]{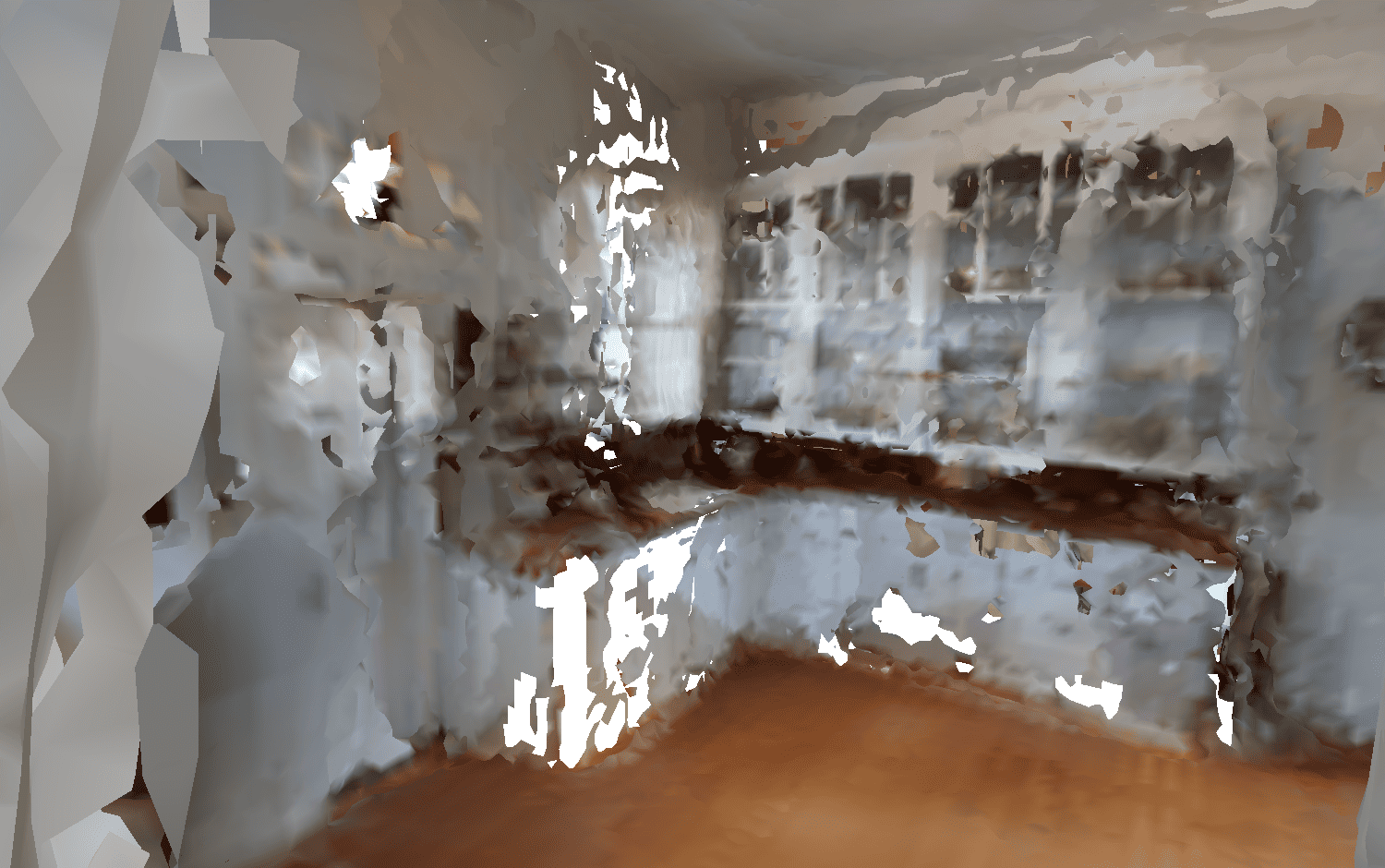}\hspace{-1cm} &
        \includegraphics[width=\mywidth]{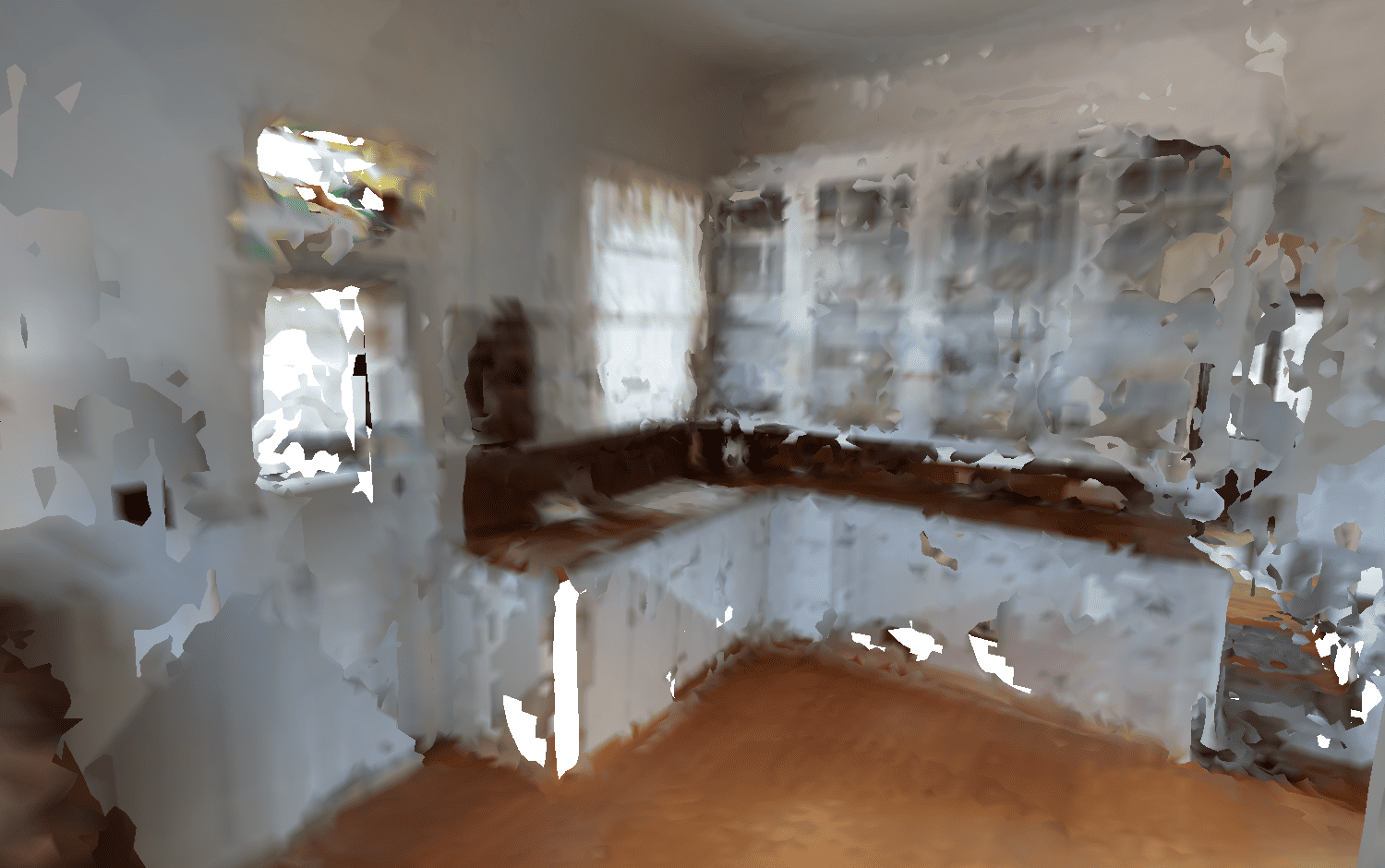}\hspace{-1cm} &
        \includegraphics[width=\mywidth]{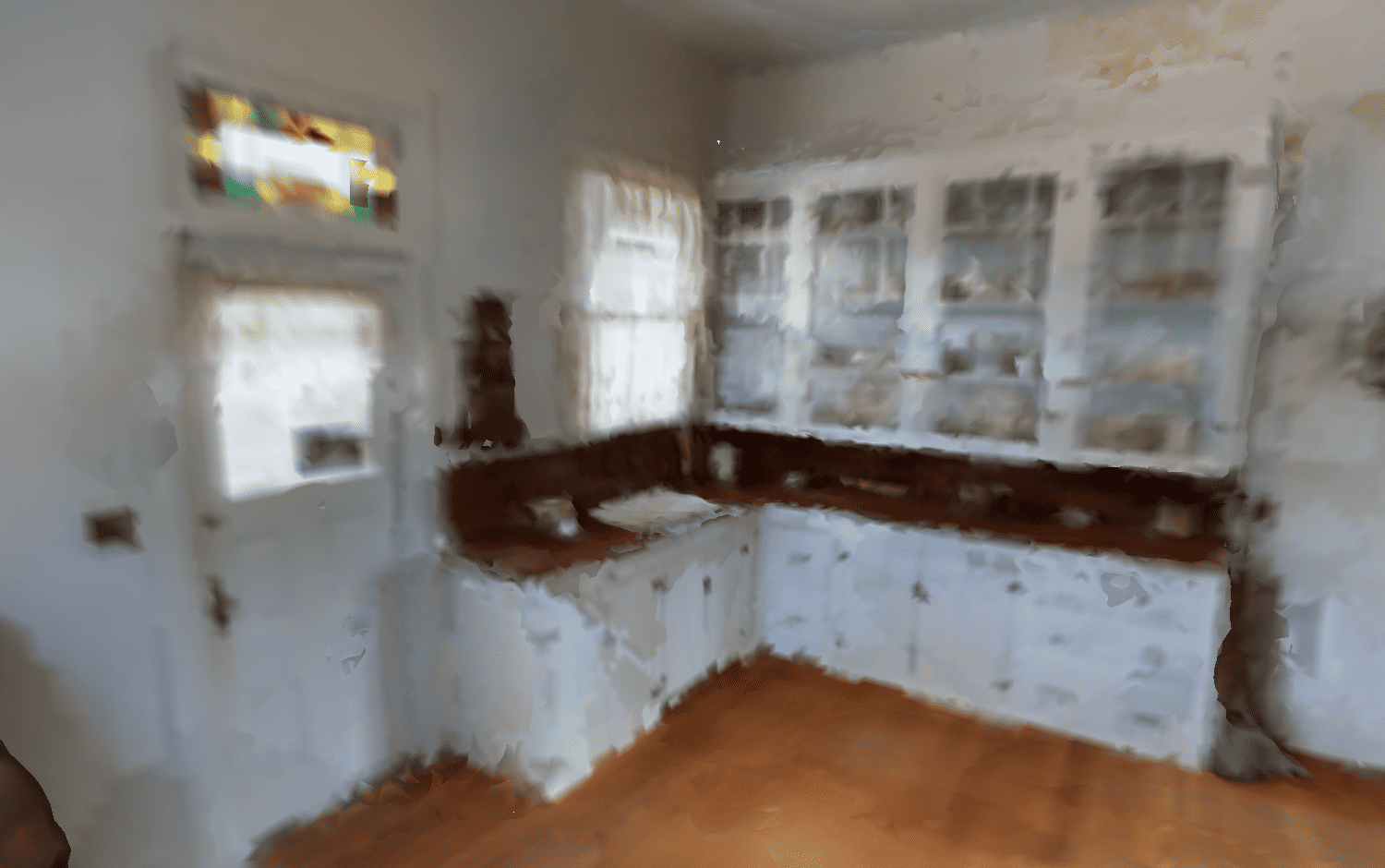}\hspace{-1cm} &
        \includegraphics[width=\mywidth]{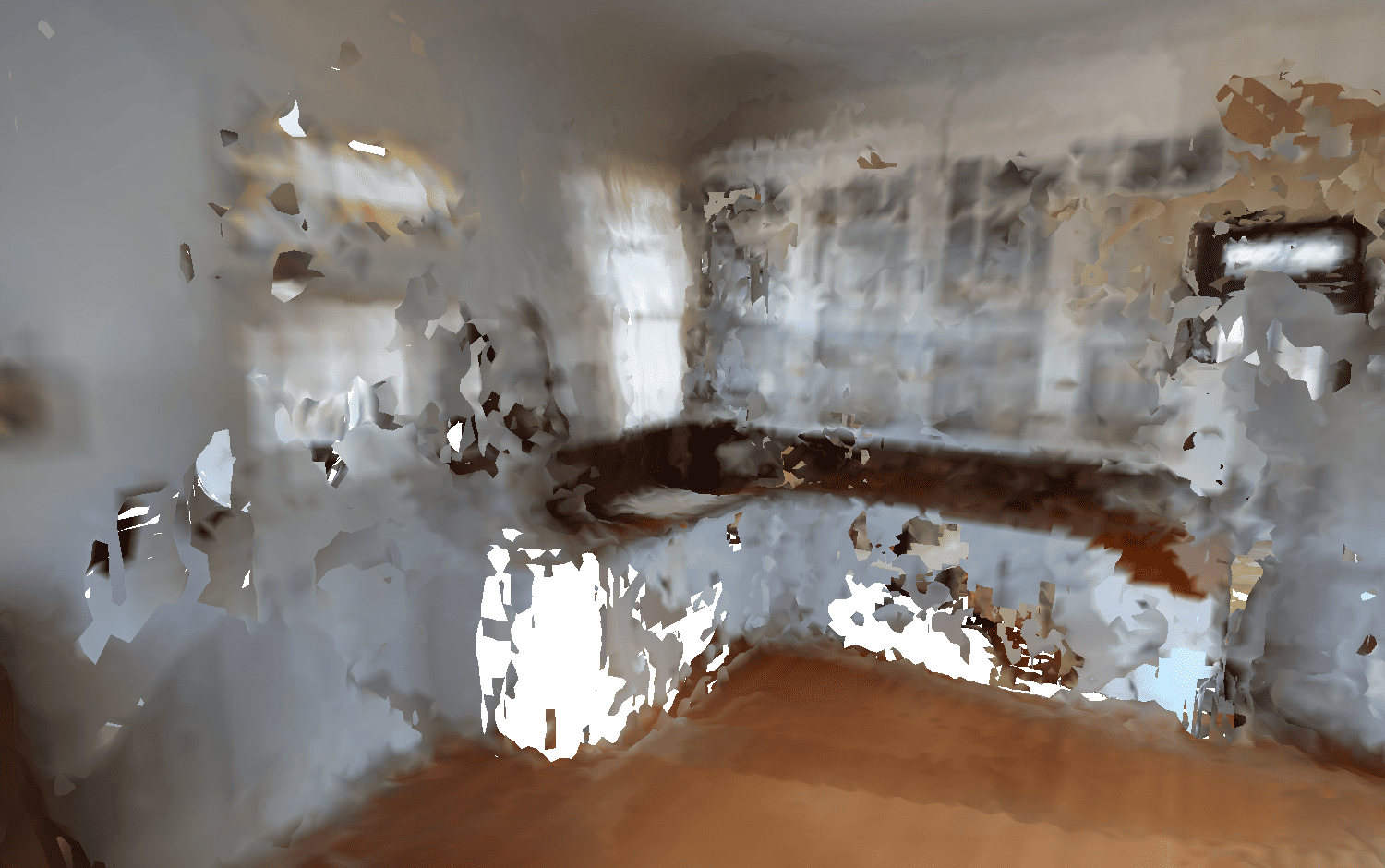}\hspace{-0.8cm} 
         \\

        \includegraphics[width=\mywidth]{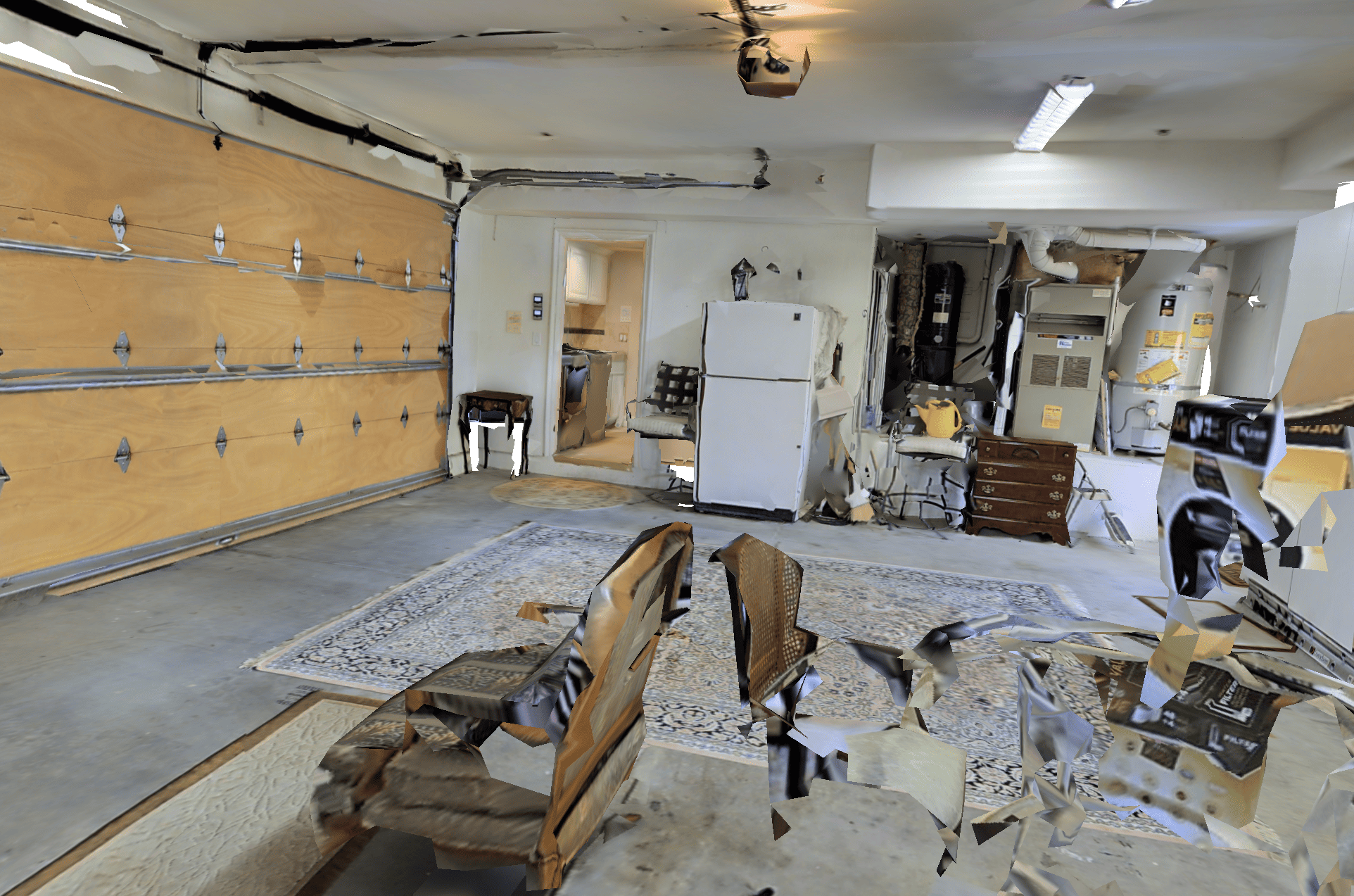} &
        \includegraphics[width=\mywidth]{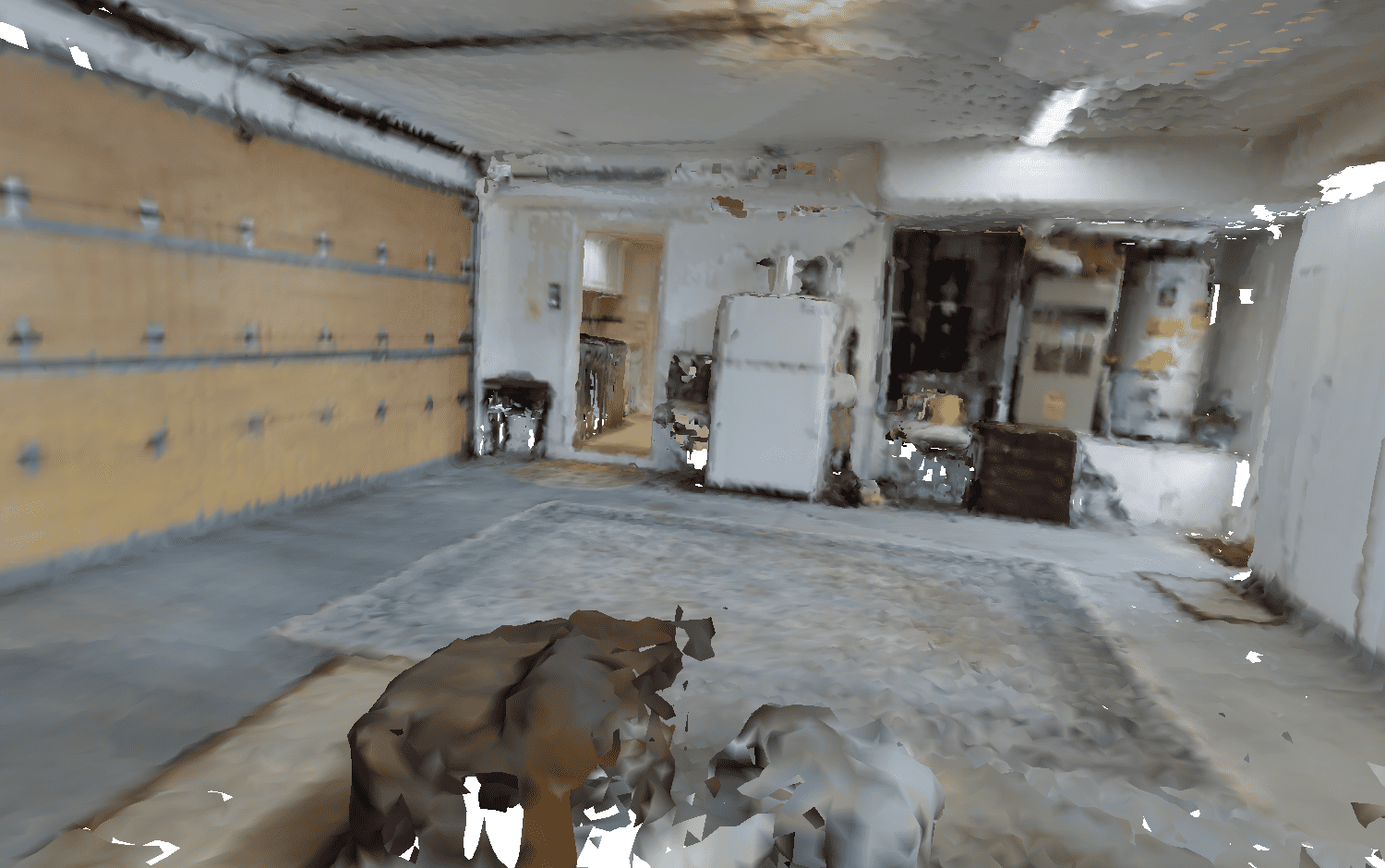} &
        \includegraphics[width=\mywidth]{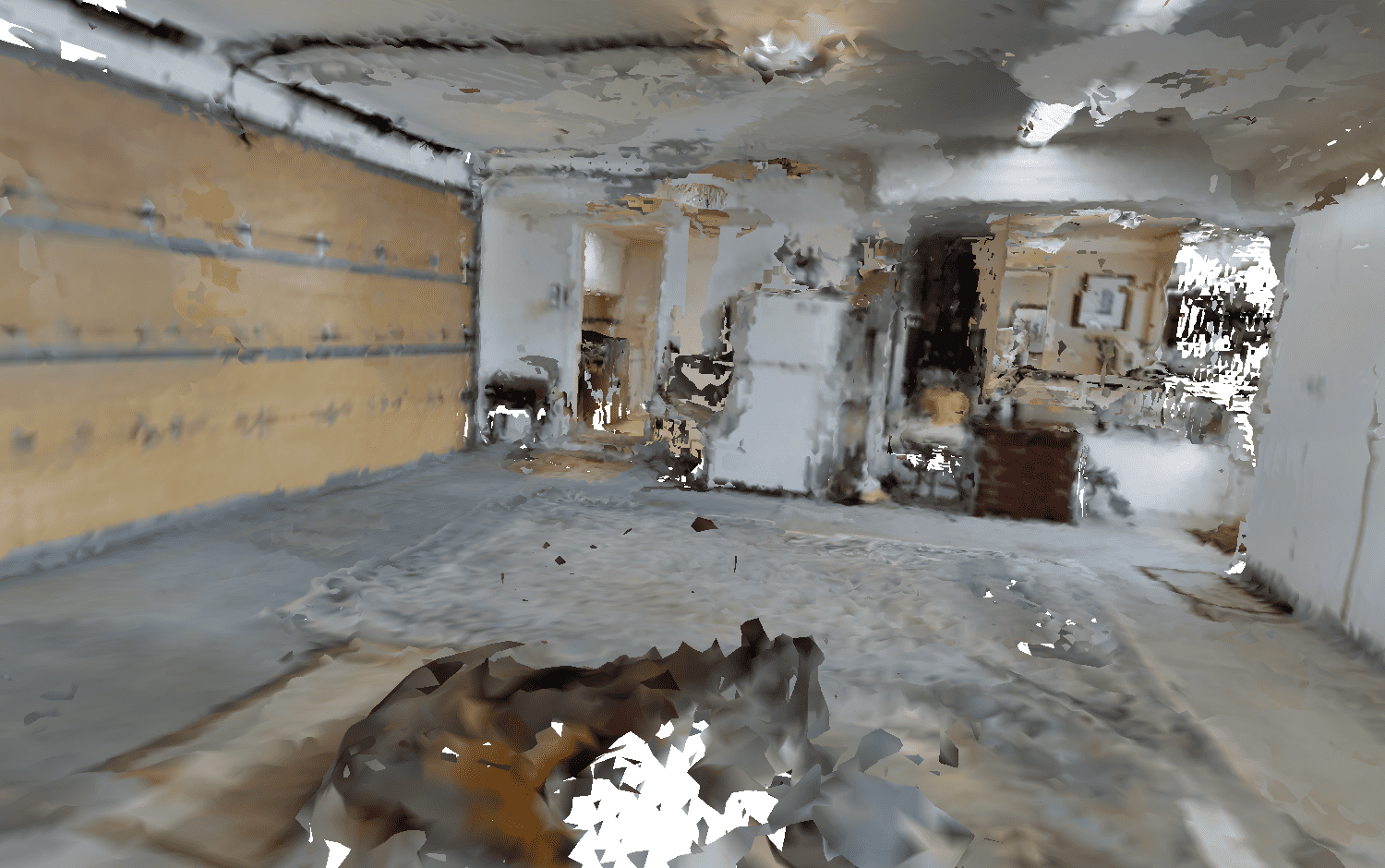} &
        \includegraphics[width=\mywidth]{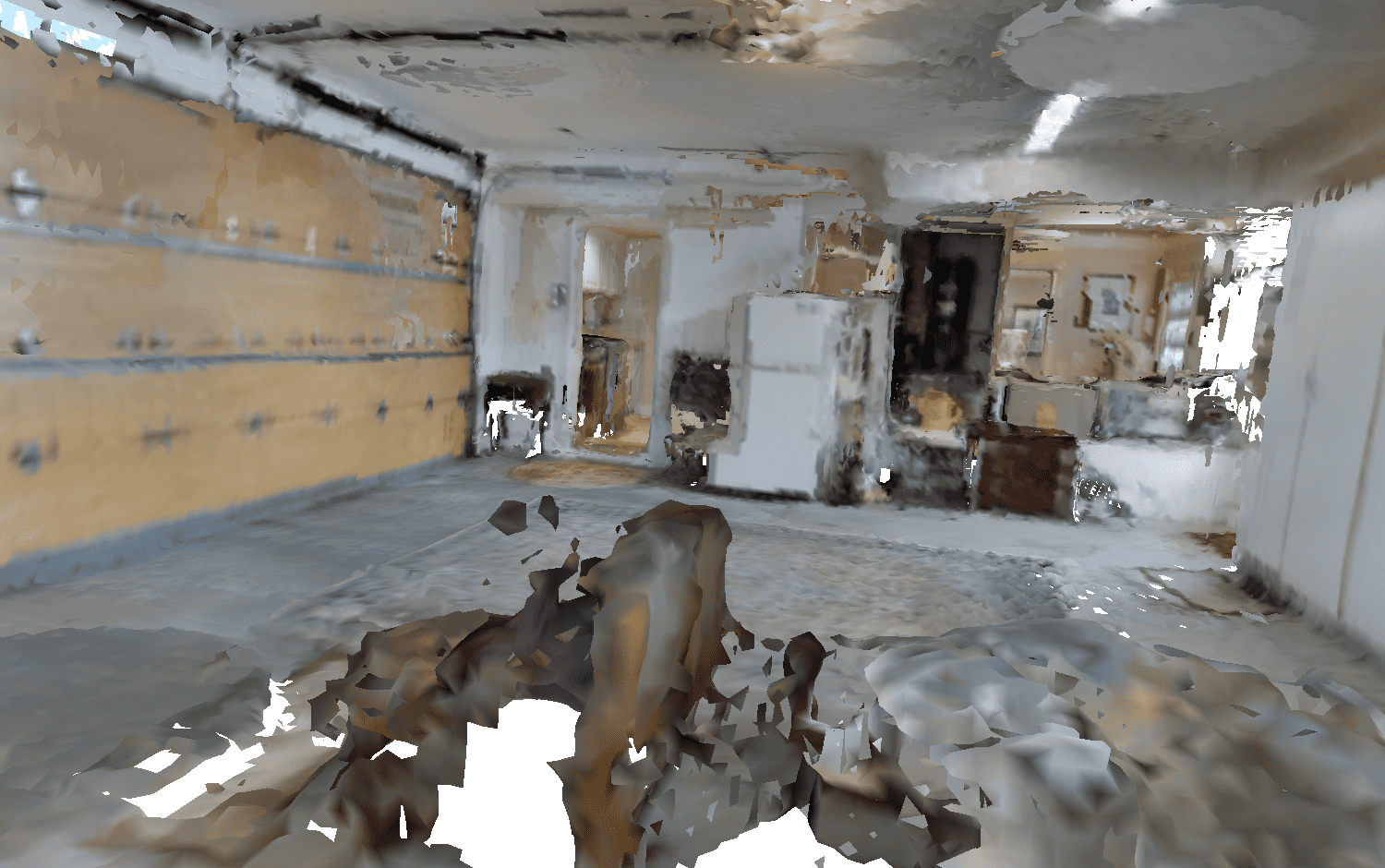} &
        \includegraphics[width=\mywidth]{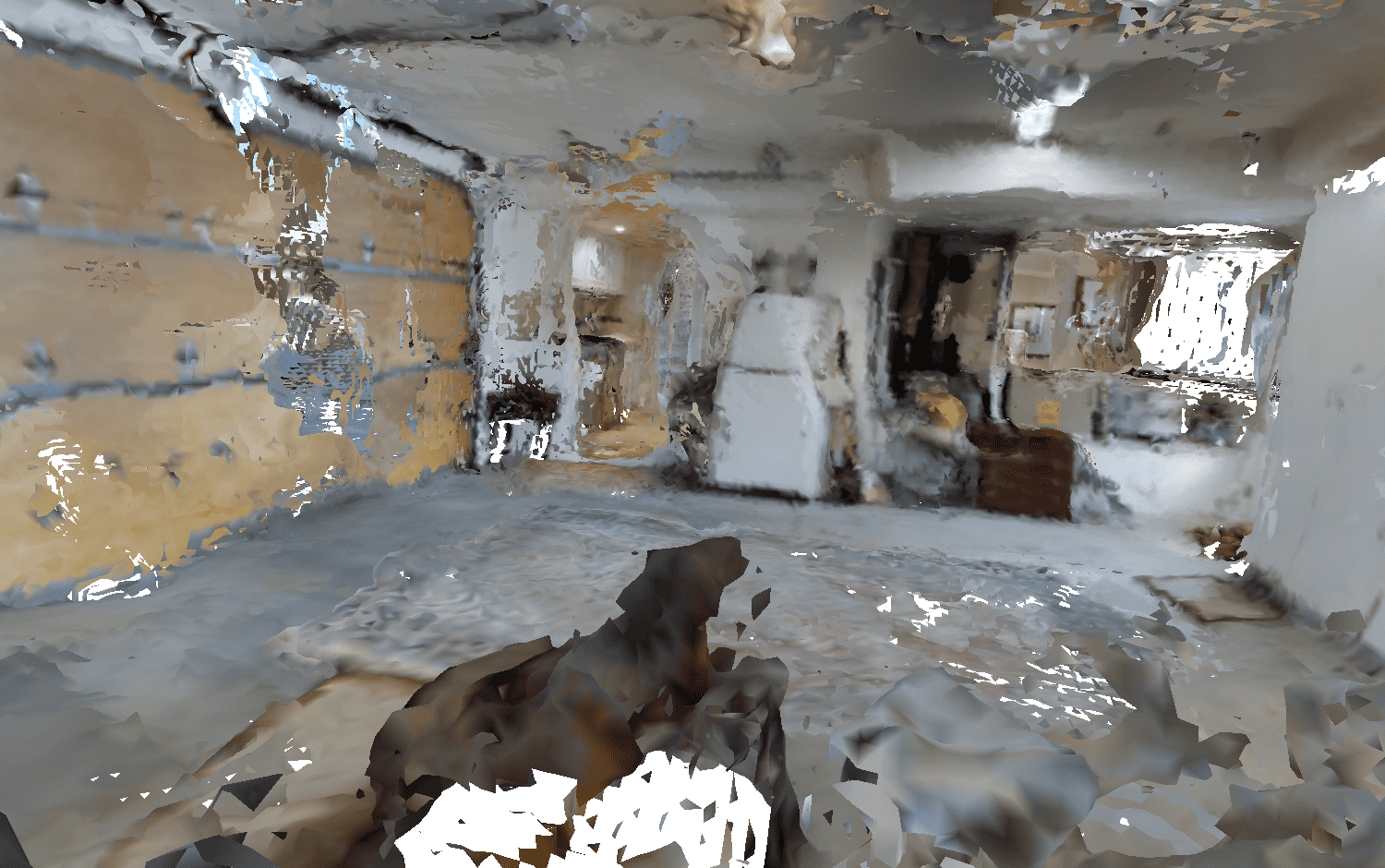} &
        \includegraphics[width=\mywidth]{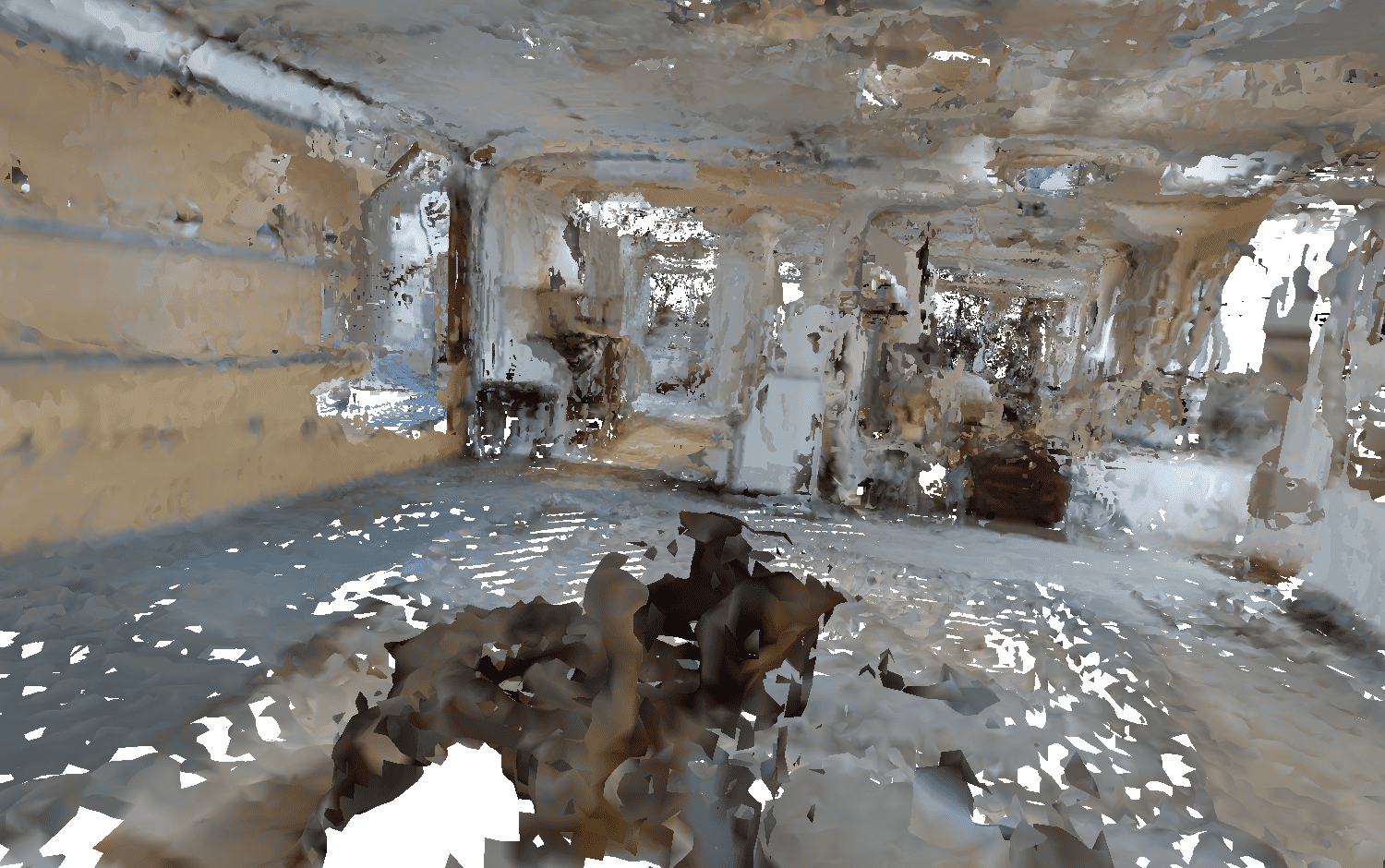} 
 \\

        \includegraphics[width=\mywidth]{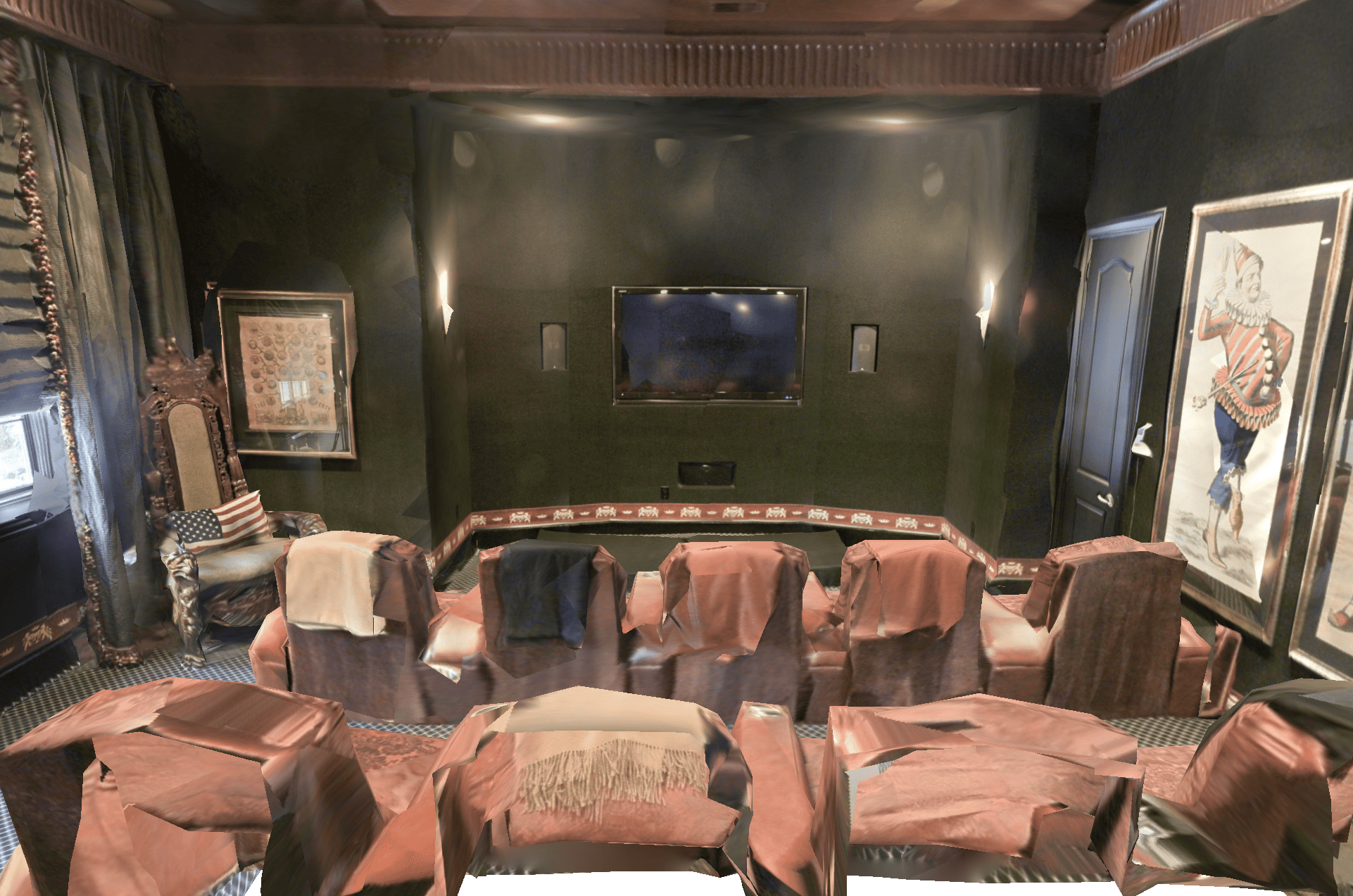} &
                \includegraphics[width=\mywidth]{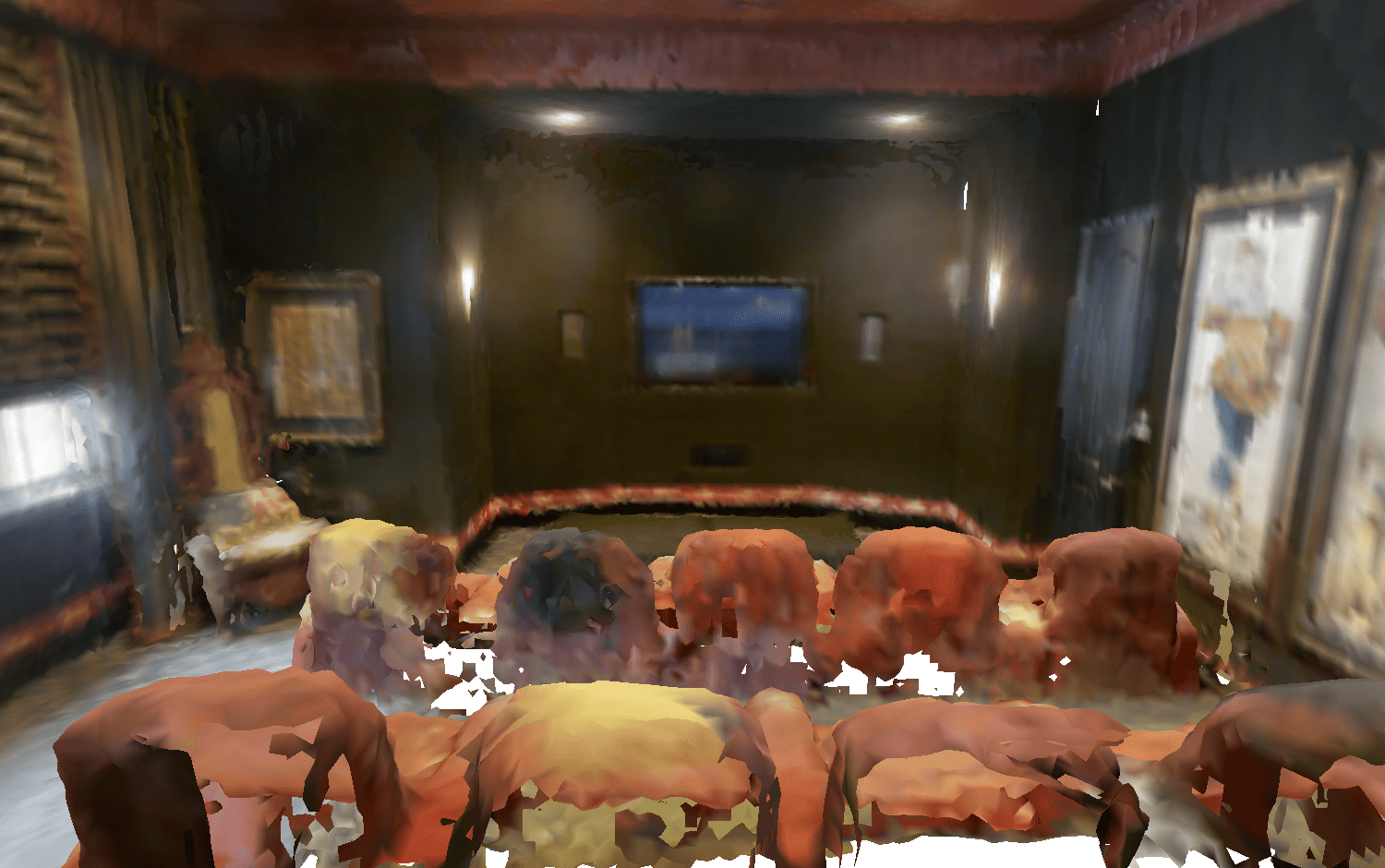}  &
        \includegraphics[width=\mywidth]{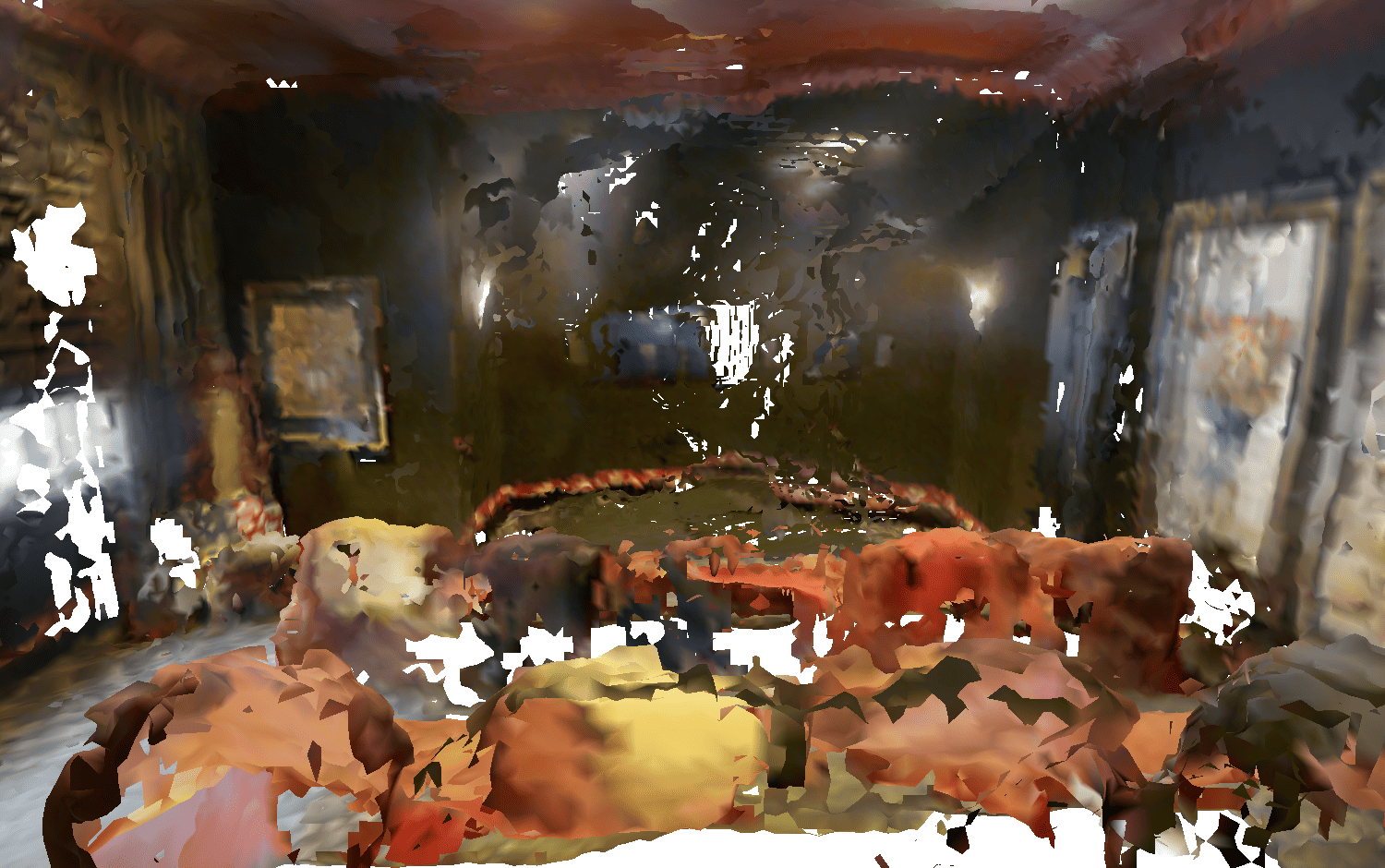} &
        \includegraphics[width=\mywidth]{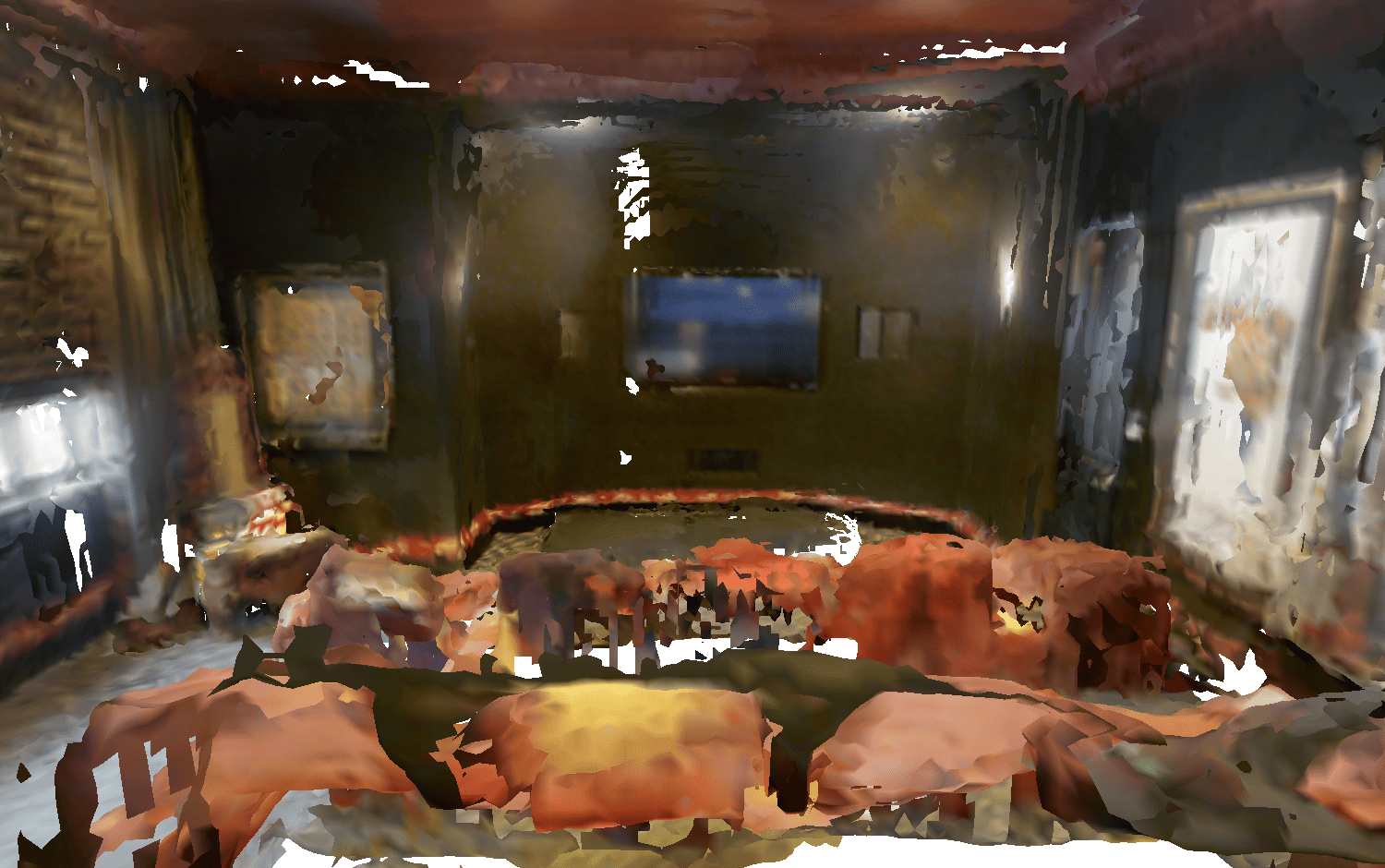} &
        \includegraphics[width=\mywidth]{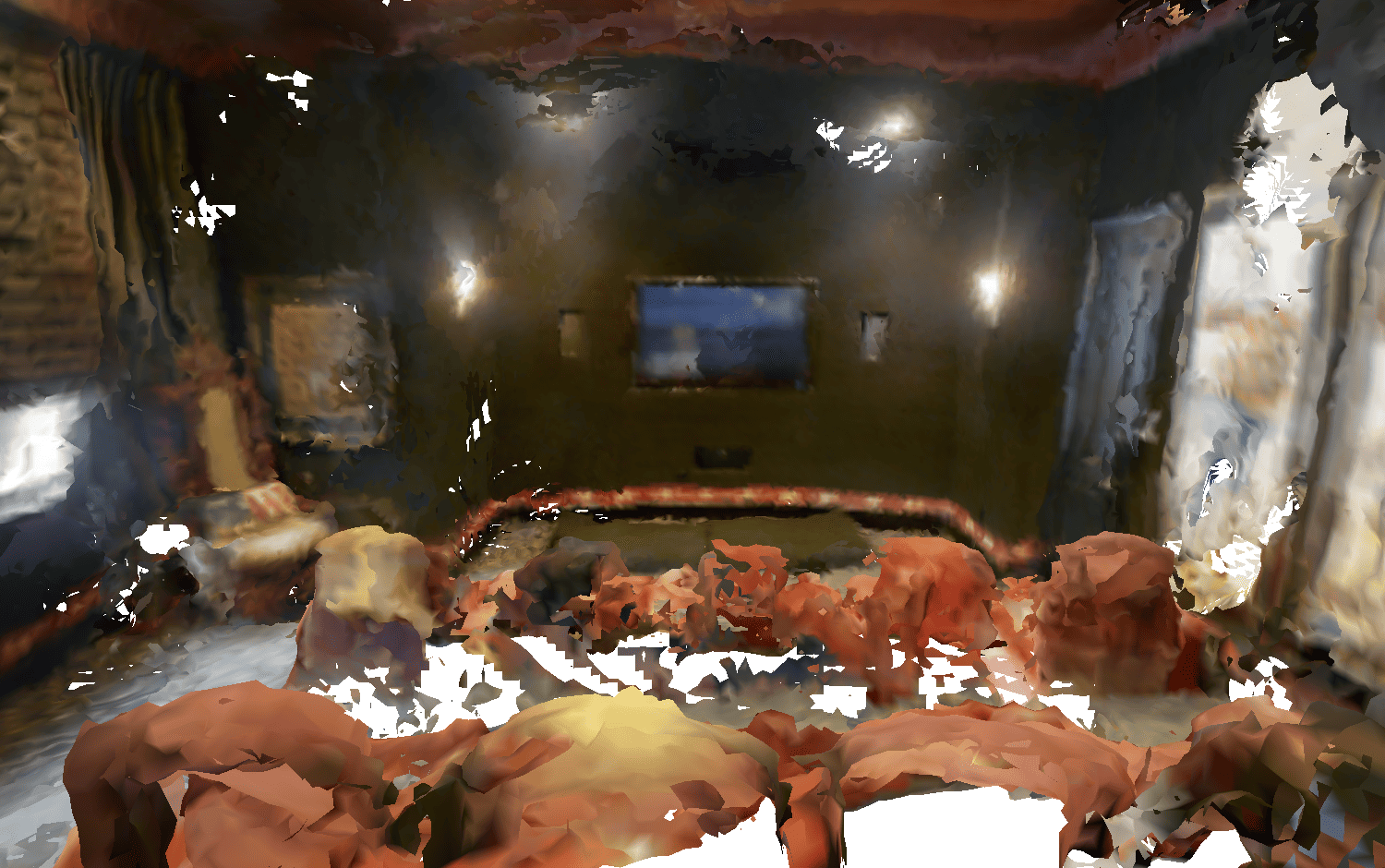} &
        \includegraphics[width=\mywidth]{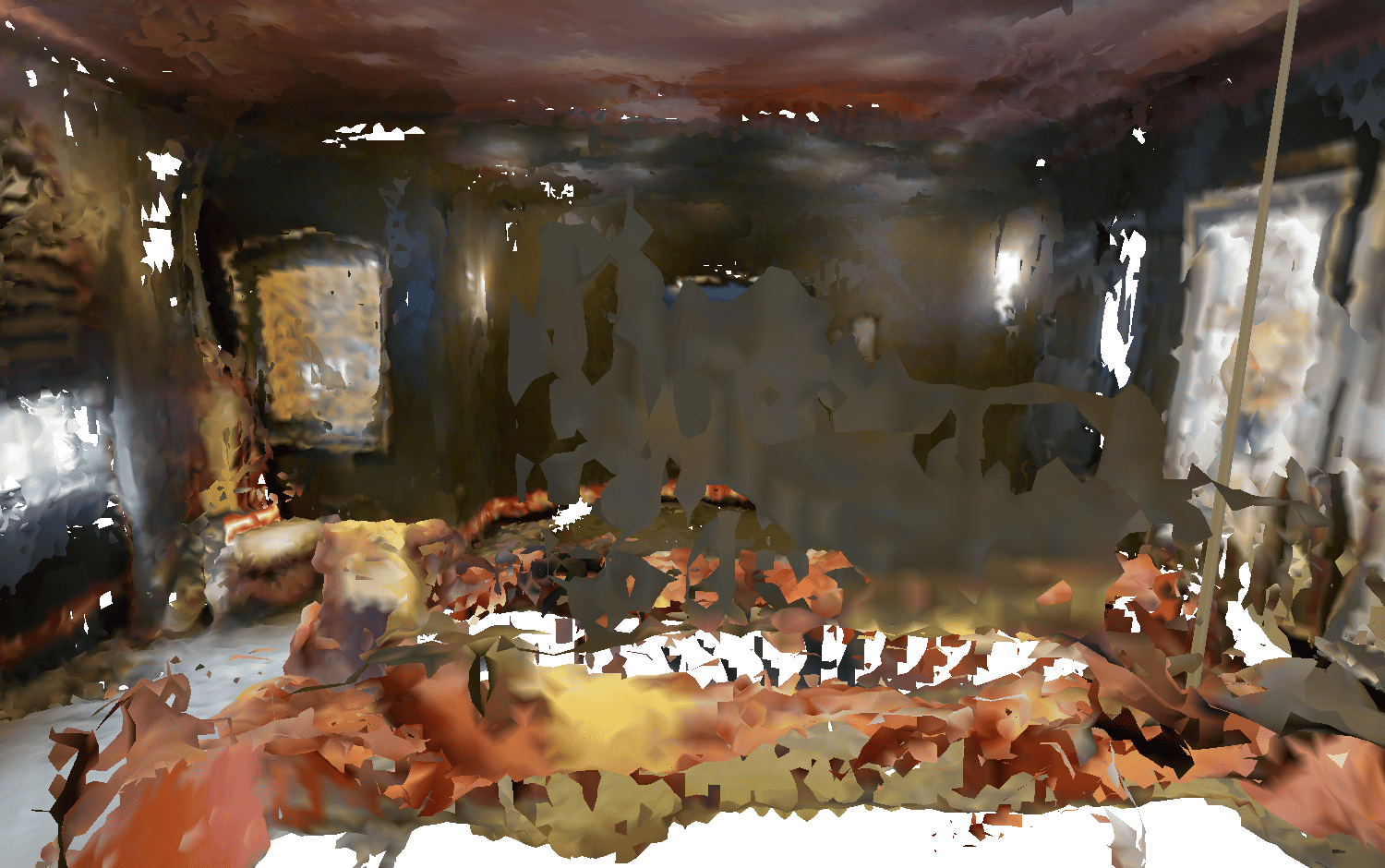} \\

        \includegraphics[width=\mywidth]{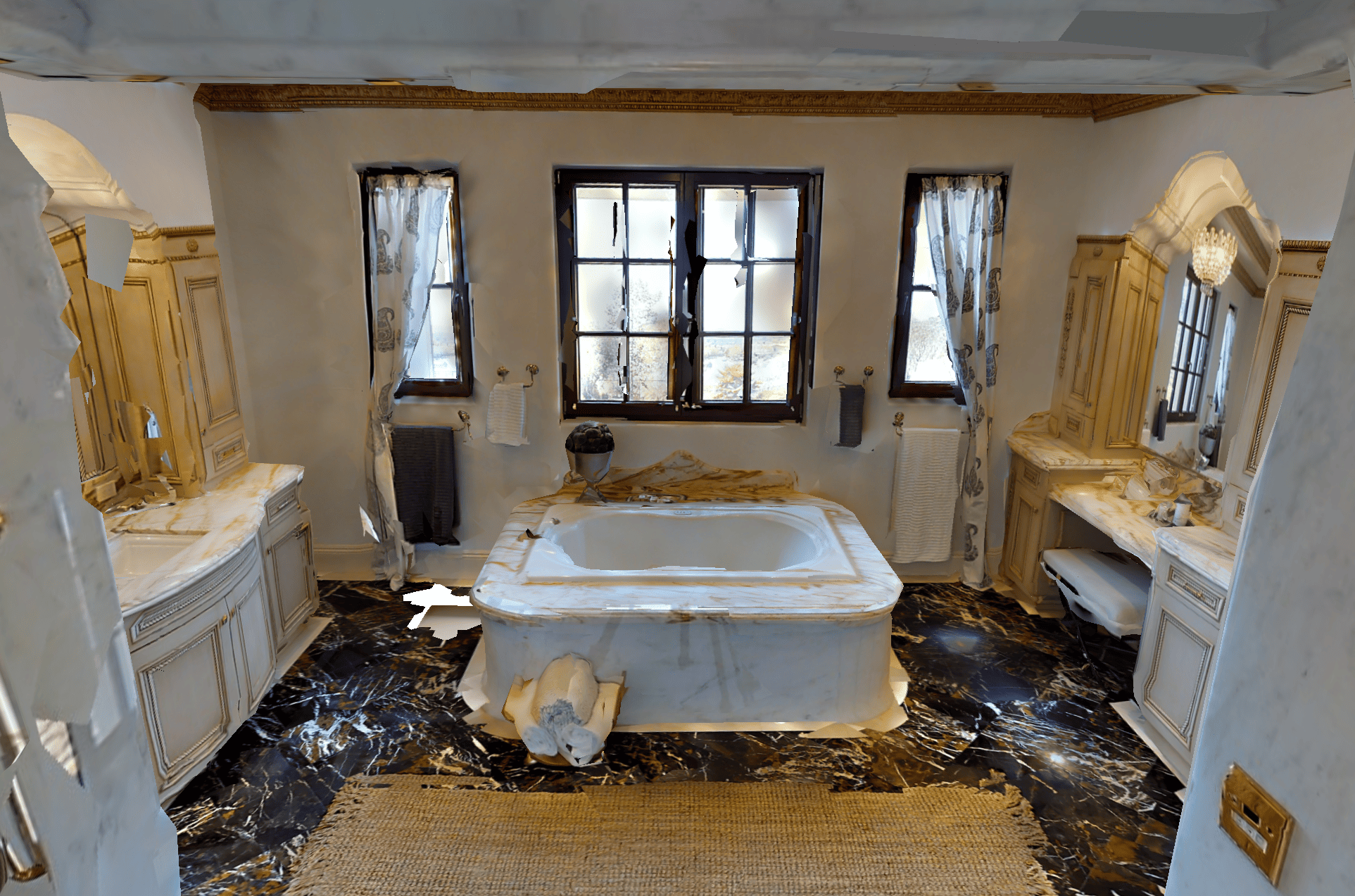} &
                \includegraphics[width=\mywidth]{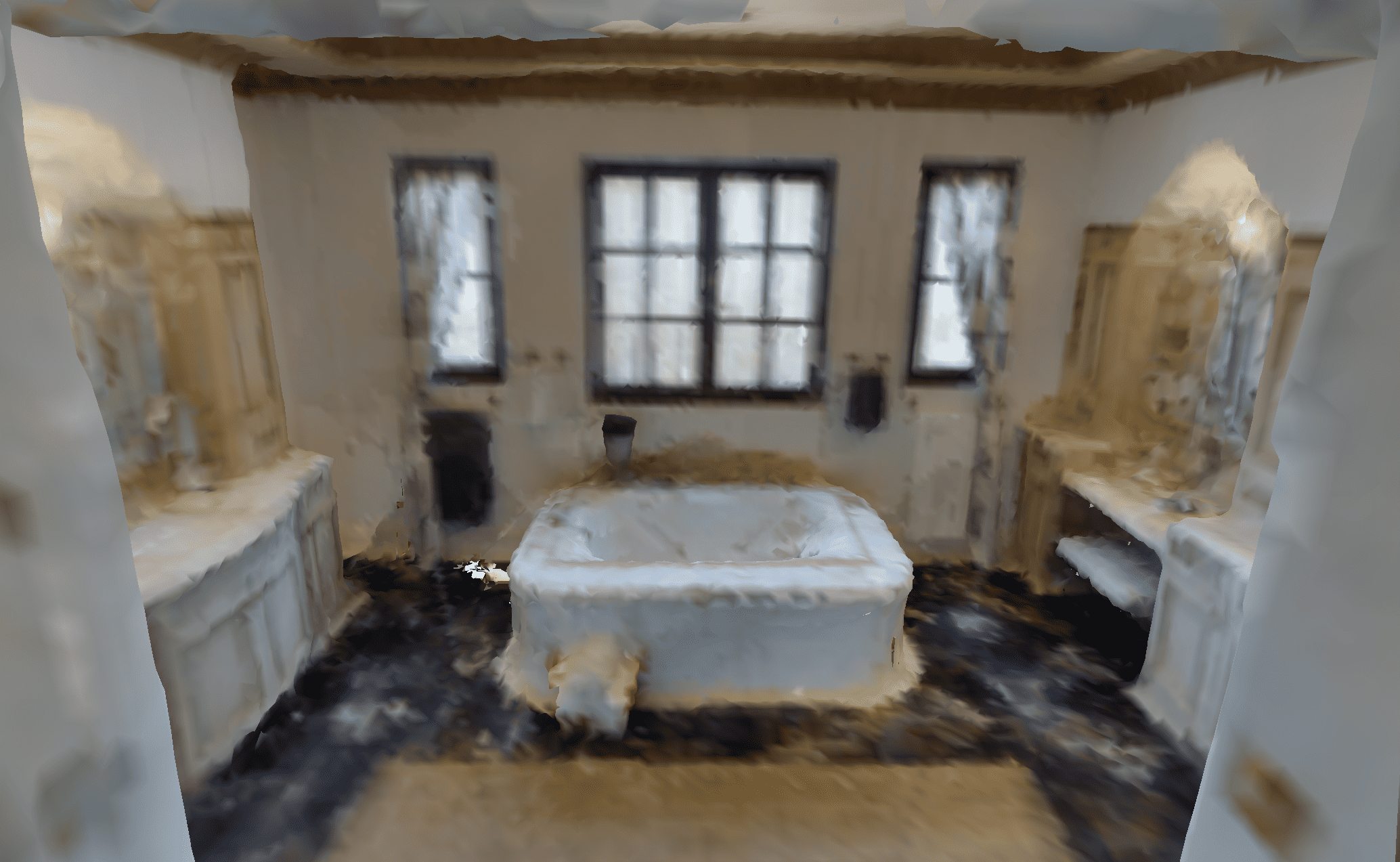} &
        \includegraphics[width=\mywidth]{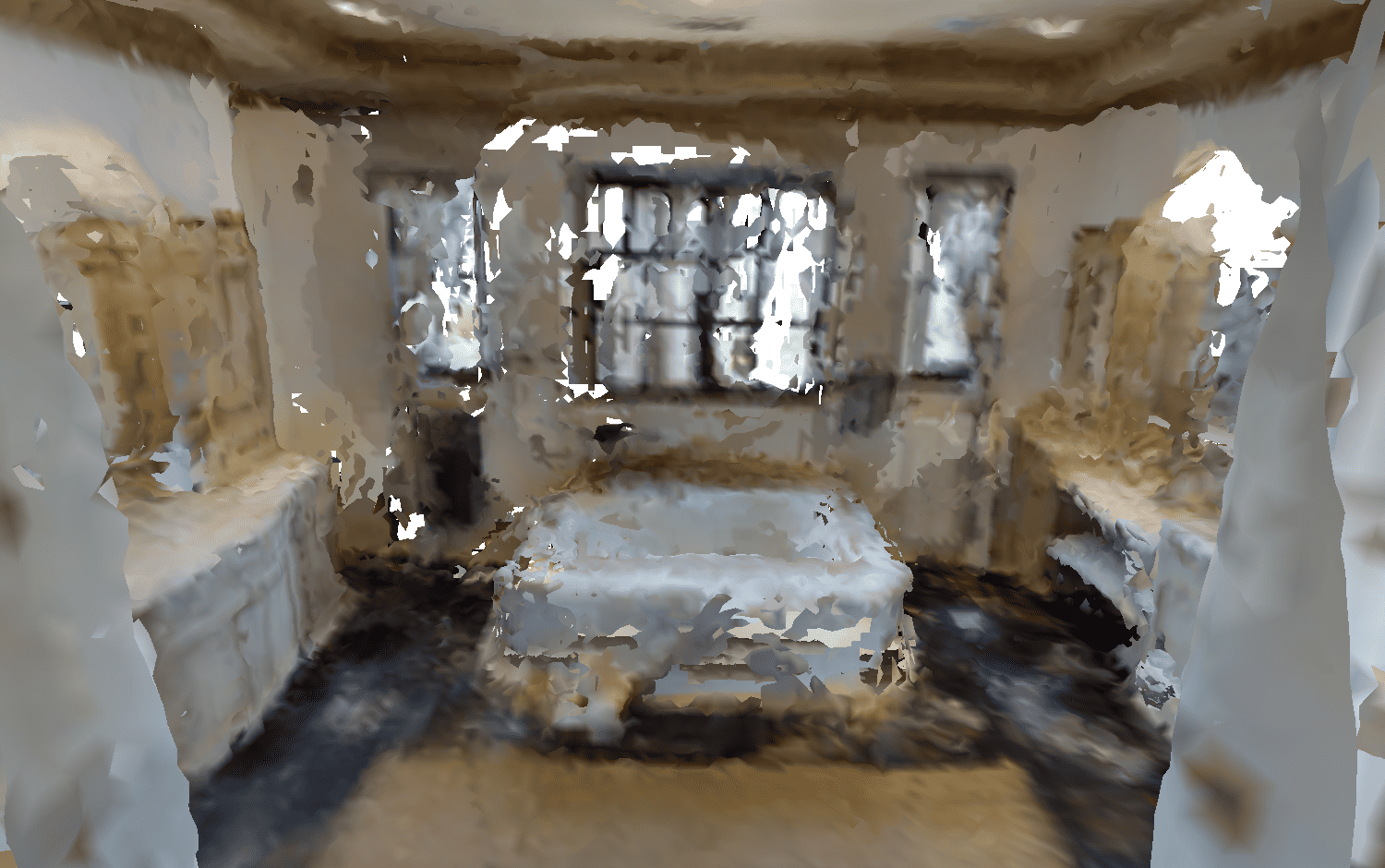} &
        \includegraphics[width=\mywidth]{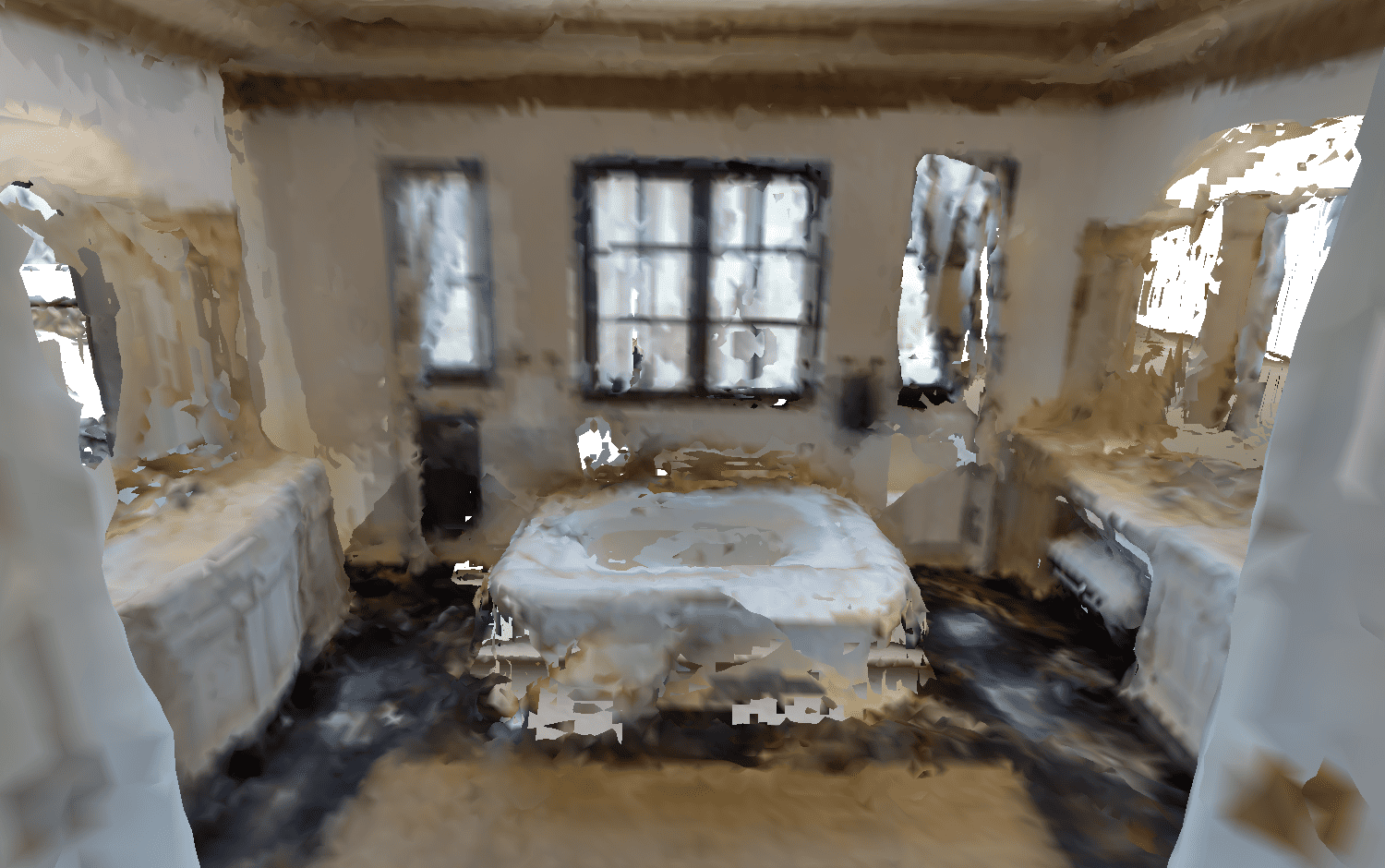} &
        \includegraphics[width=\mywidth]{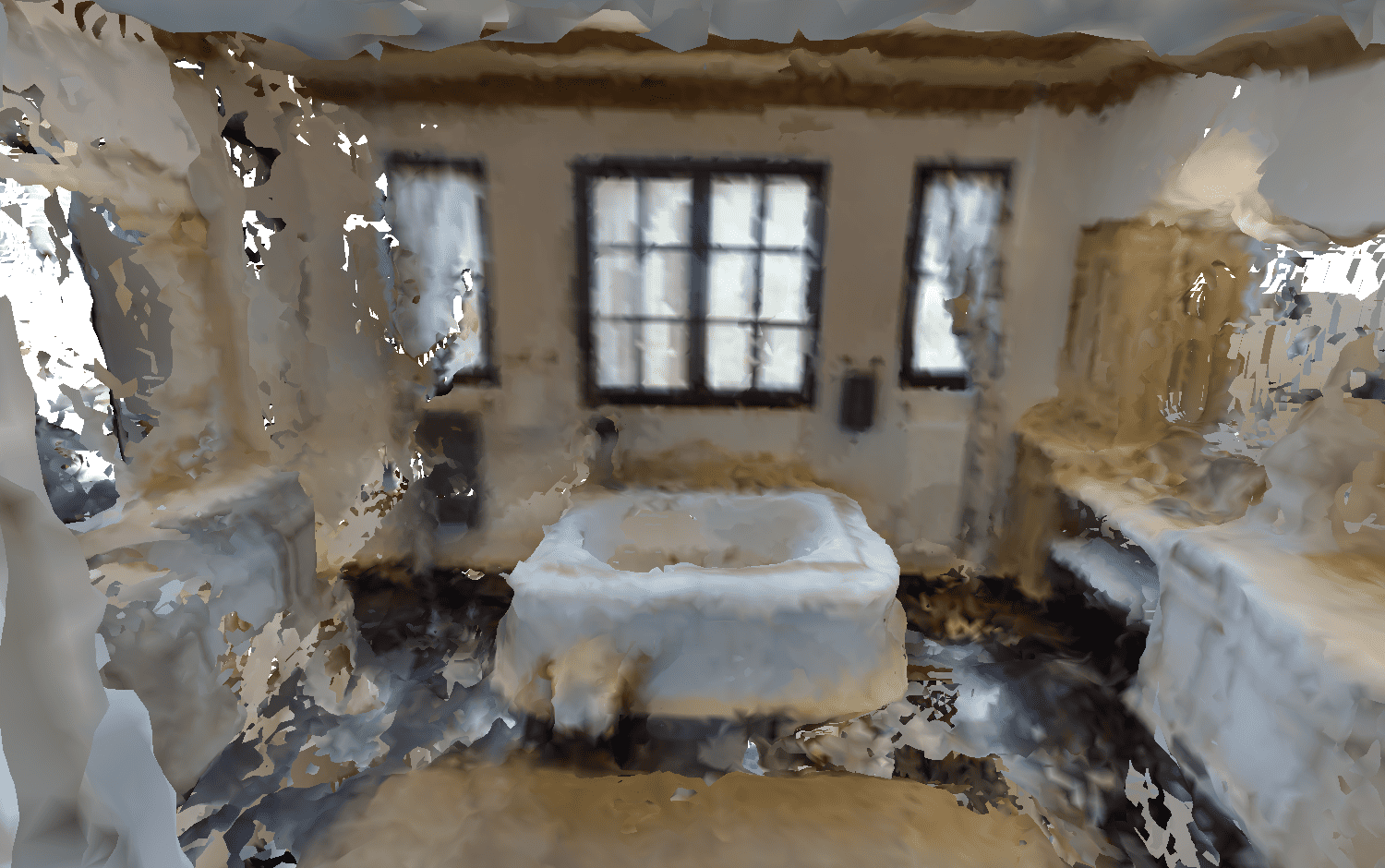} &
        \includegraphics[width=\mywidth]{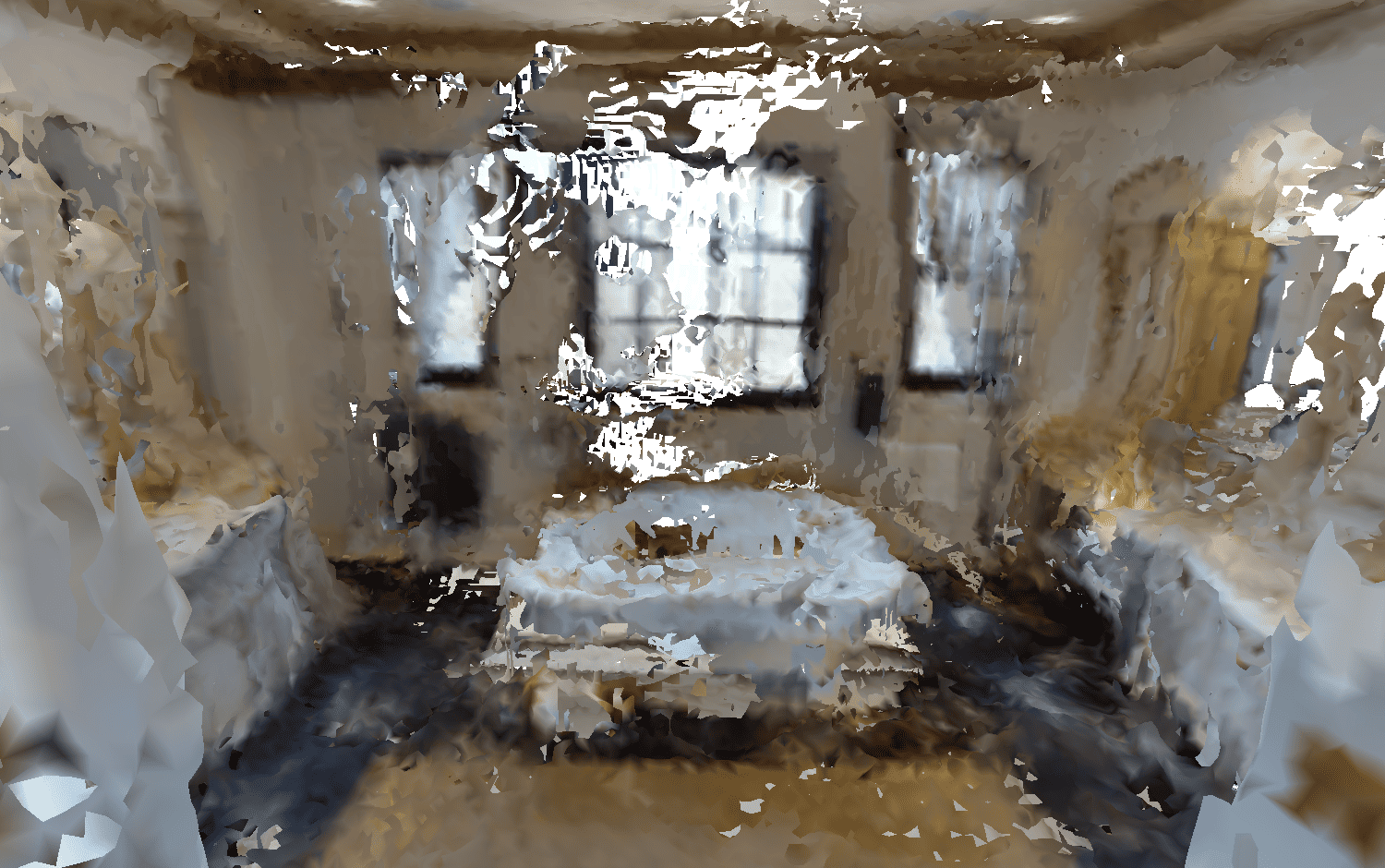} 
 \\

        \includegraphics[width=\mywidth]{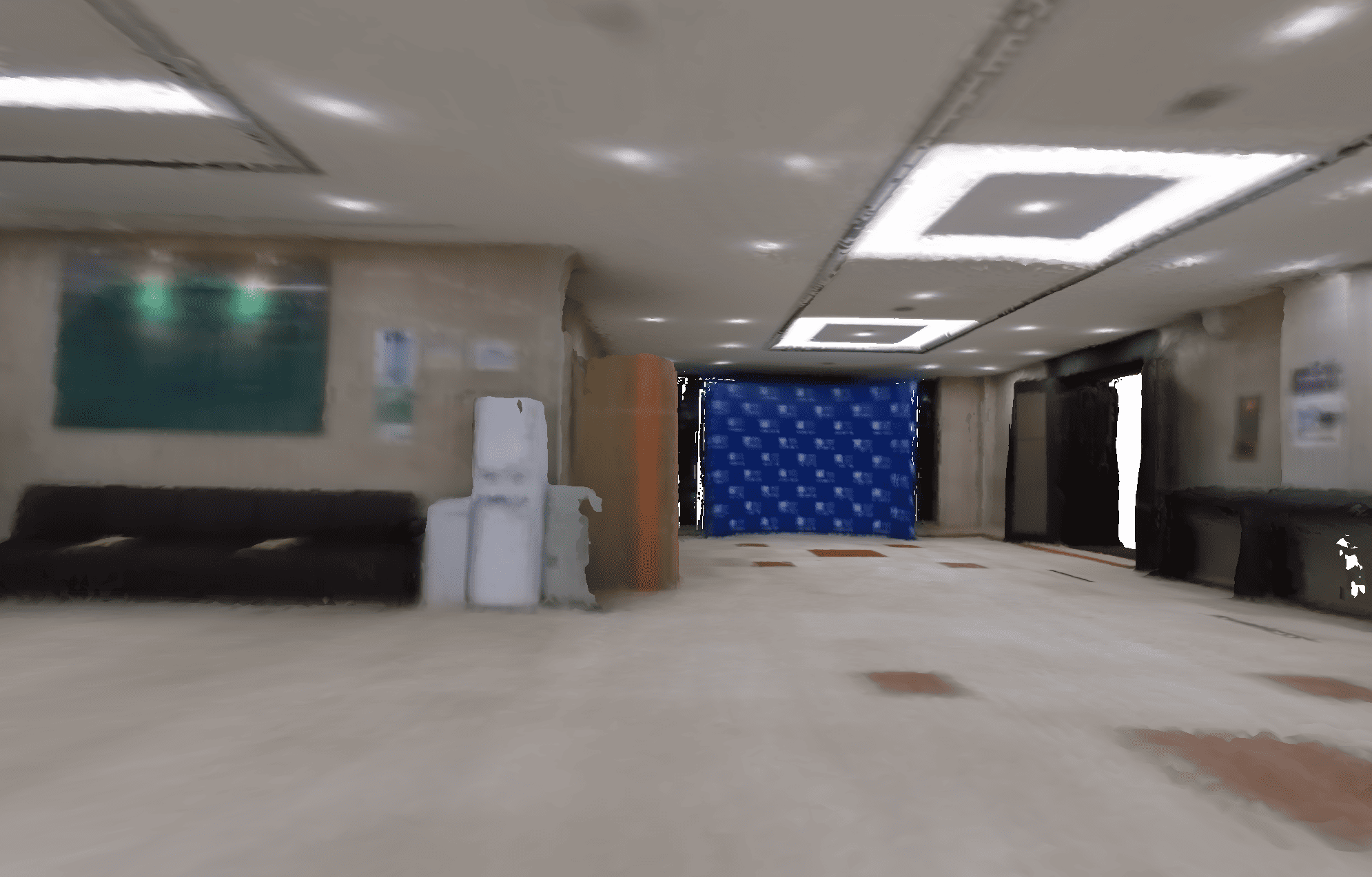} &
                \includegraphics[width=\mywidth]{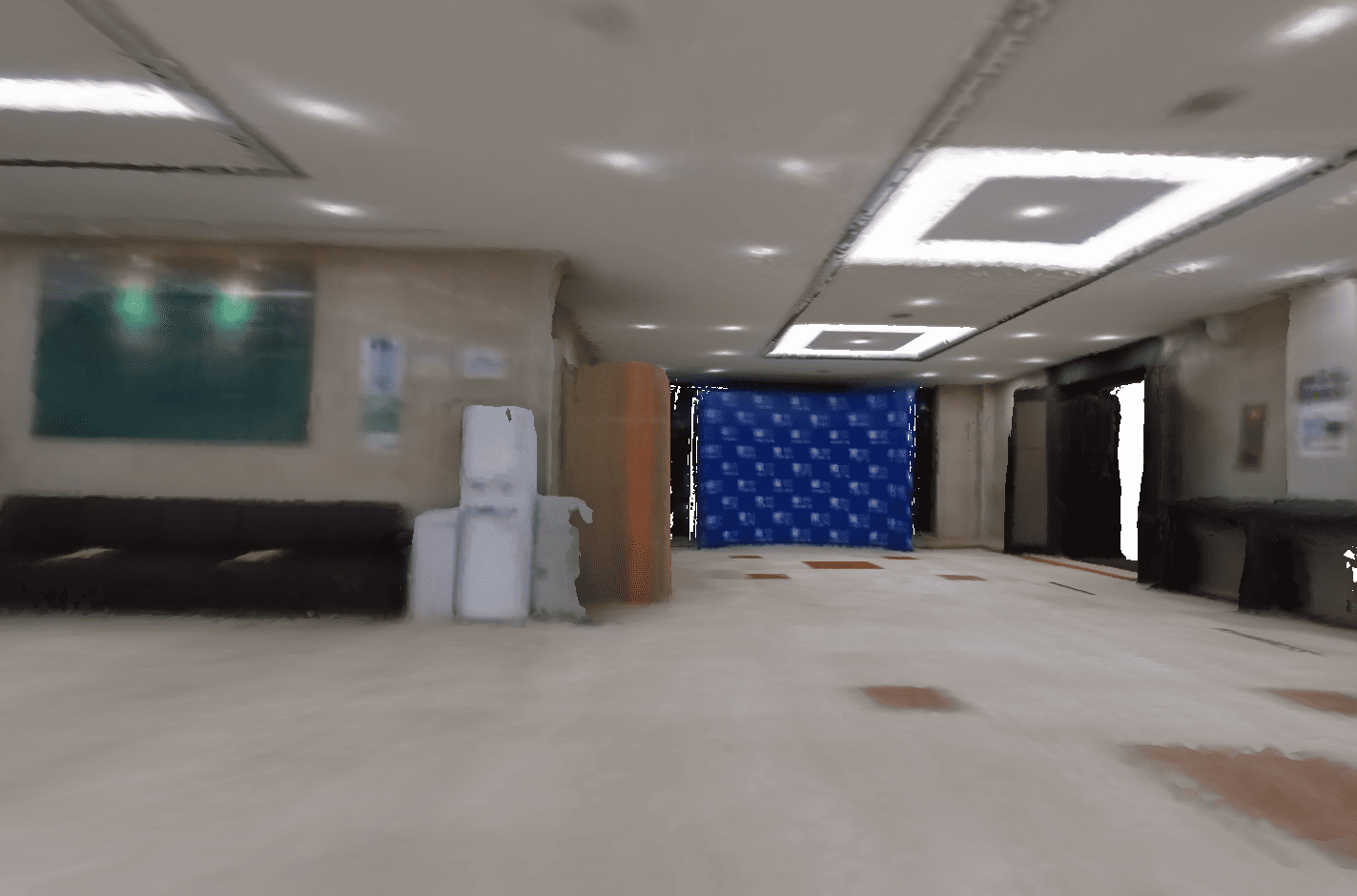} &
        \includegraphics[width=\mywidth]{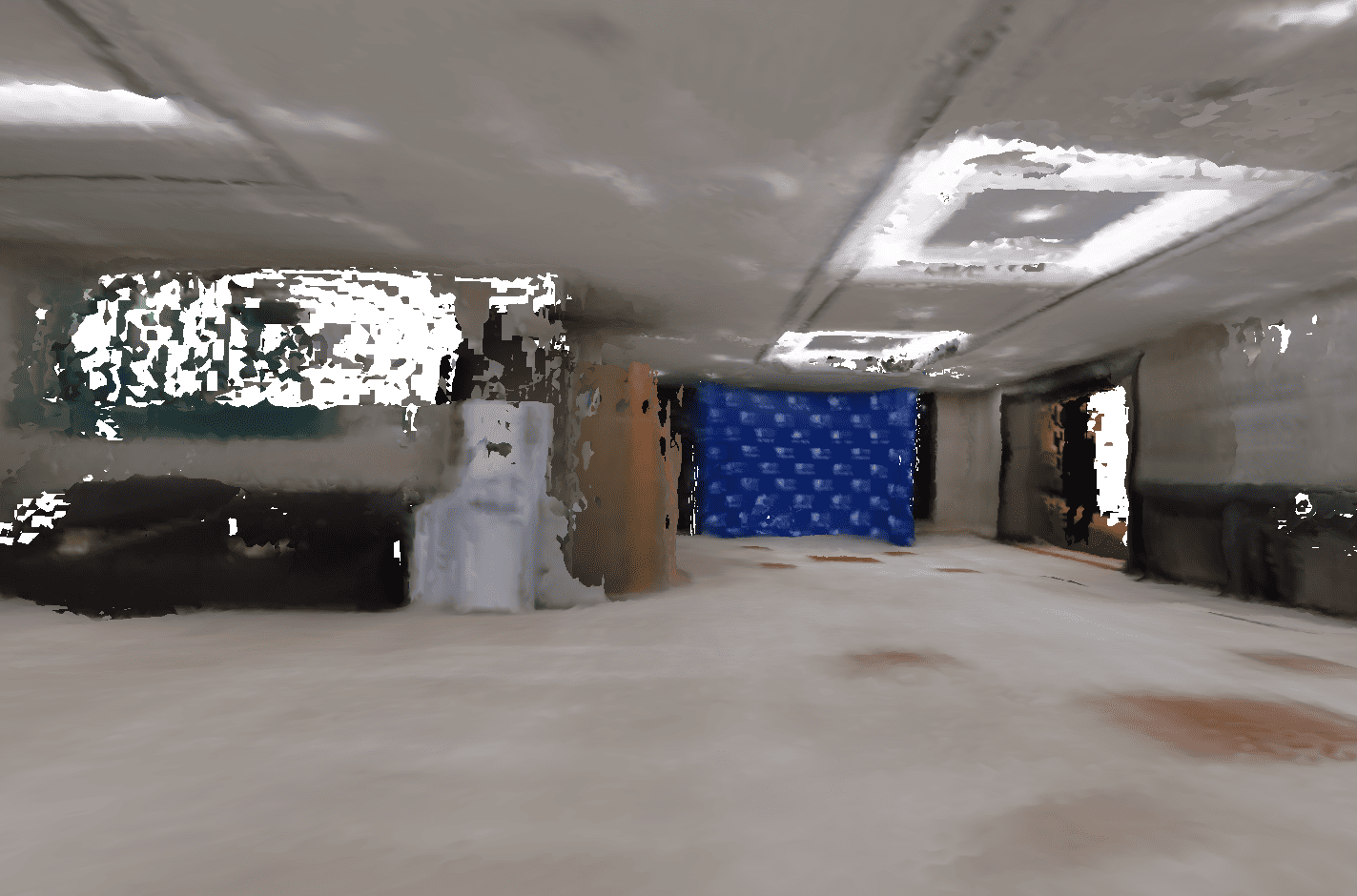} &
        \includegraphics[width=\mywidth]{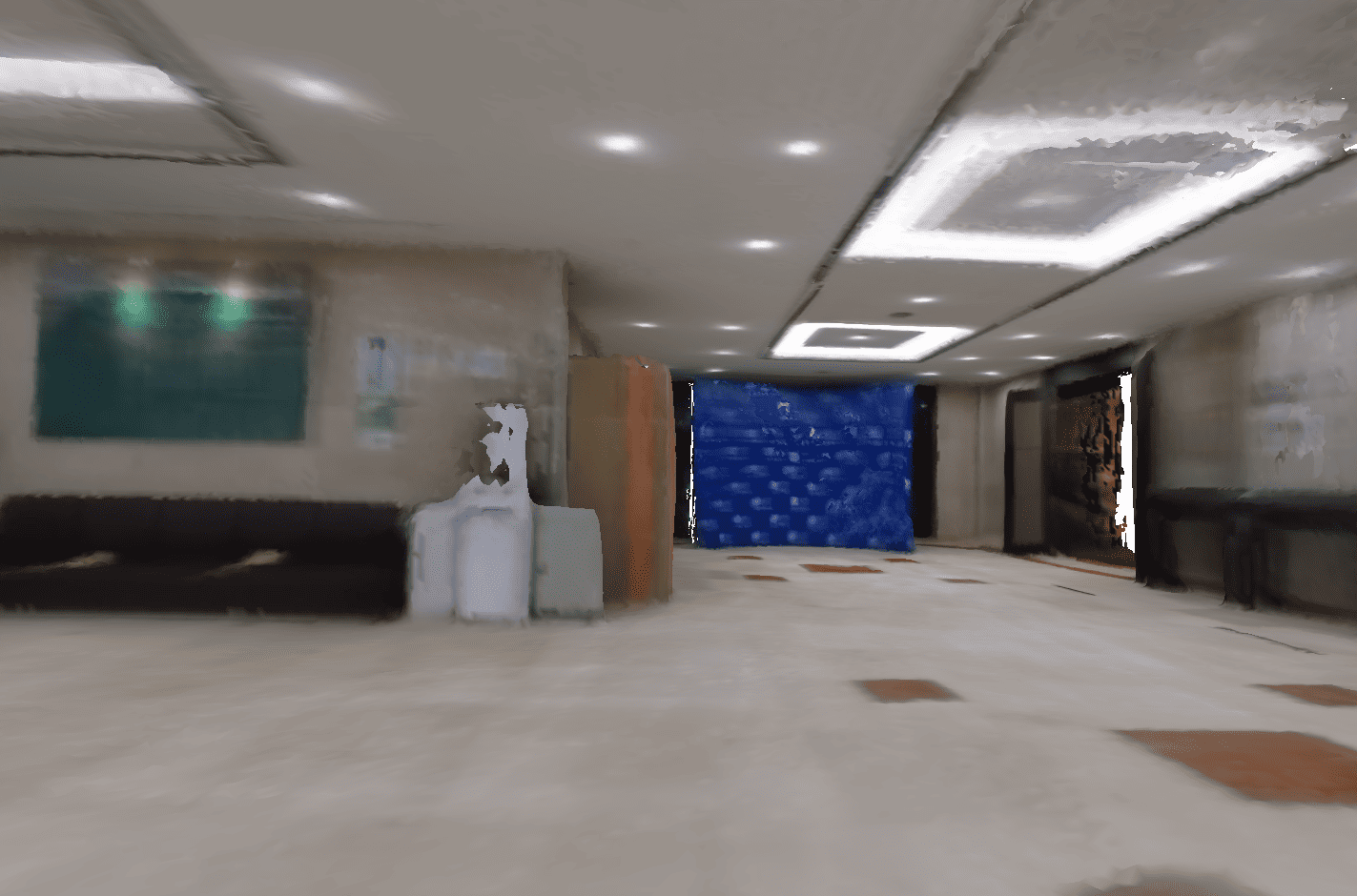} &
        \includegraphics[width=\mywidth]{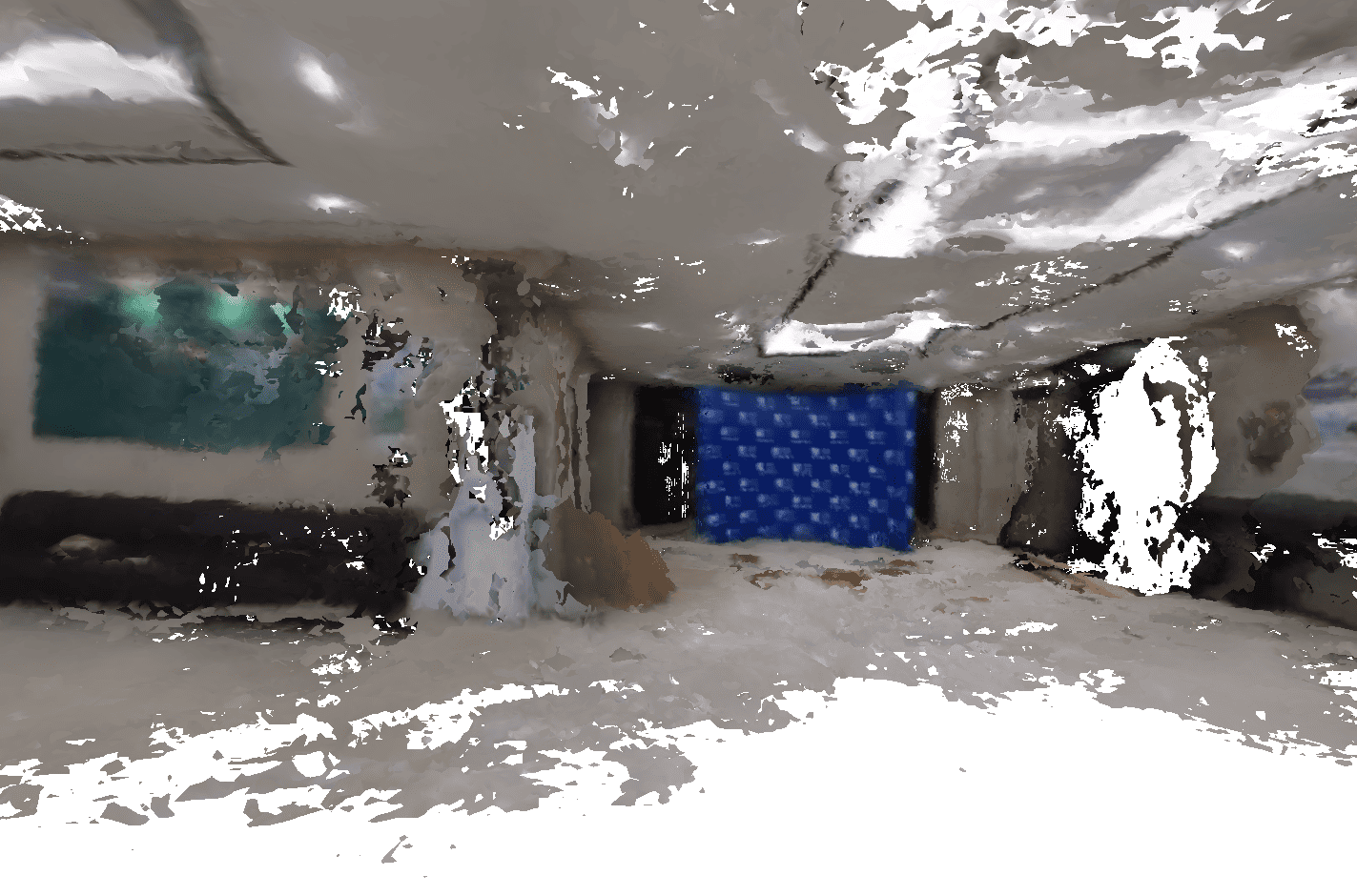} &
        \includegraphics[width=\mywidth]{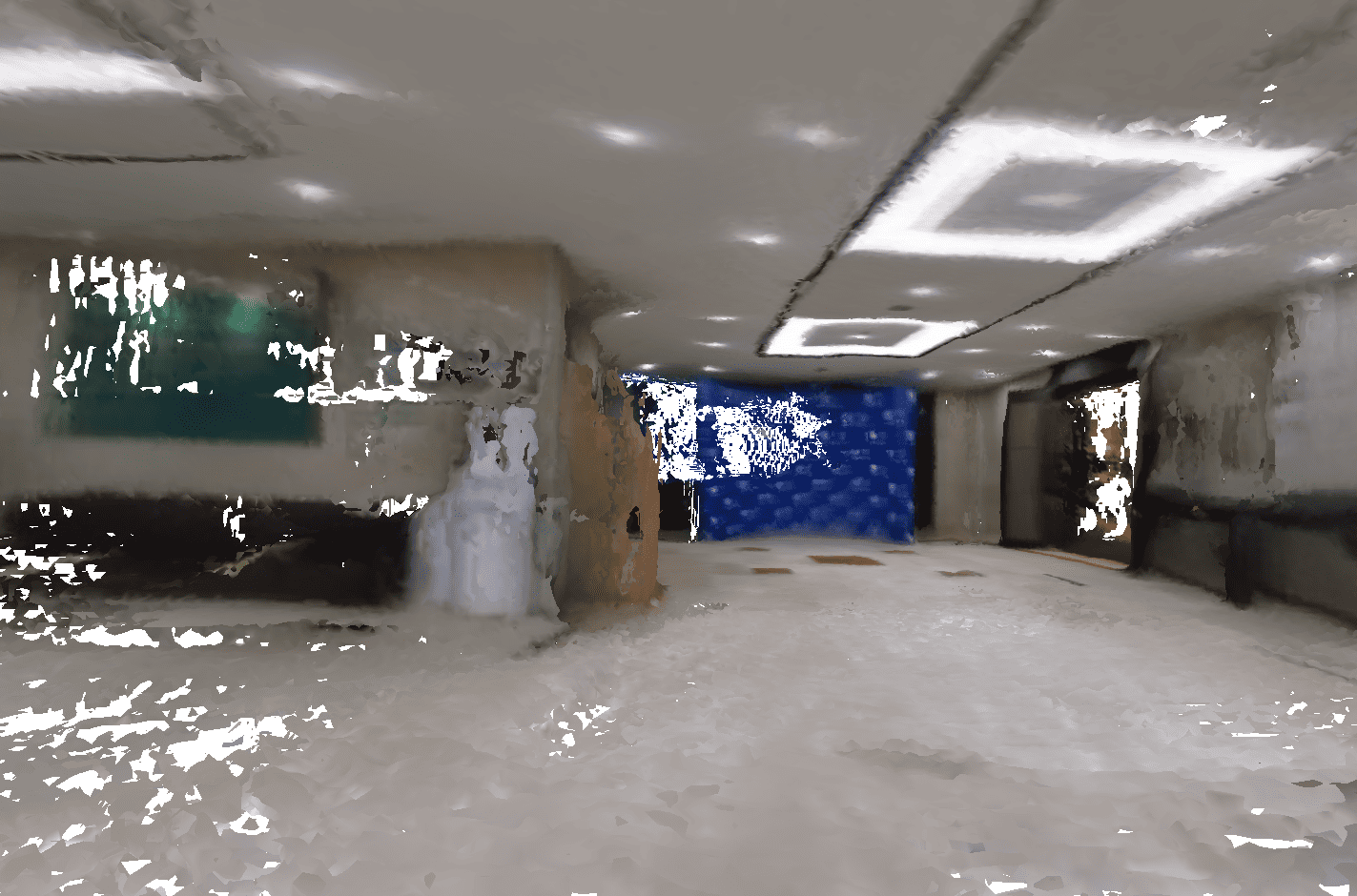} \\

          GT & 360Recon & Bifuse++ & PanoFormer & FoVA-Depth & 360-MVSNet$^{*}$
    \end{tabular}
    \vspace{-8pt}
    \caption{\textbf{3D Reconstruction Comparison.} The 3D reconstruction results are compared with various methods in different scenes. The reconstruction results of our algorithm in these scenarios are more accurate and complete.}
    \label{fig:recon_more}
    \vspace{-0.4cm}
\end{figure*}

 \cref{fig:recon_more} provides a comparison of 3D reconstruction results from different algorithms across various scenes, including a kitchen, garage, home theater, bathroom, and lounge. These scenes are selected from the Matterport3D~\cite{Matterport3D} and OmniScenes~\cite{OmniScenes} datasets. From \cref{fig:recon_more}, it can be qualitatively judged that our algorithm outperforms others in terms of reconstruction quality and completeness in these scenes.

\section{Testing on Self-Collected Data}
\label{collected}

In this section, we further evaluate the generalization and robustness of our algorithm by qualitatively visualizing depth estimation and 3D reconstruction results on our self-collected data.

We employed an Insta360 X2 panoramic camera to capture real-world panoramic image data. The camera generates panoramic images by stitching together outputs from two back-to-back fisheye lenses. We captured approximately one minute of video in a hall environment and extracted panoramic images at a rate of one frame per second. We obtained the camera pose information necessary for the model using SphereSfM~\cite{sphereSfM1,sphereSfM2,sphereSfM3}, which results in discrepancies between the estimated poses and the camera's actual poses during testing.

Qualitative visualizations of depth estimation and 3D reconstruction are presented in \cref{fig:cdepth} and \cref{fig:crecon}. The results demonstrate that the model achieves satisfactory depth estimation and 3D reconstruction performance on our self-collected real-world data, indicating the model's generalization capability and robustness under imprecise camera poses.
\section{Disccussion}
\label{disscusion}
In summary, our supplementary material demonstrates the generalization and robustness of our approach. While maintaining precise depth estimations, our model achieves computational time and GPU memory consumption that are acceptably low compared to existing MVS depth estimation algorithms. This is primarily accomplished by foregoing the use of 3D CNNs for directly processing the 4D cost volume; instead, we employ a simple MLP to reduce the dimensionality of the cost volume. Additionally, we utilize a pre-trained image encoder to enhance image features, allowing the model to leverage detailed texture information alongside multi-view geometric constraints. Furthermore, we adopt spherical convolution methods to mitigate the impact of image distortion on the feature matching process. Through ablation experiments, we have validated the effectiveness of each component of our model.

Currently, our model still has some limitations. For instance, we are unable to reconstruct scene parts that are occluded in all images. This is because the current model only integrates depth and color information into the scene representation using the TSDF algorithm~\cite{TSDF}. Additionally, when reconstructing the 3D scene, the model initializes all voxels in the space without employing methods such as voxel hashing~\cite{hashing}. This leads to decreased reconstruction efficiency as the scene size increases and results in higher GPU memory consumption. Therefore, in future work, we plan to integrate generative algorithms to achieve more comprehensive scene reconstruction and attempt to improve reconstruction efficiency through sparse scene representations.

\end{document}